\documentclass{article}

\usepackage{arxiv}

\usepackage[utf8]{inputenc}
\usepackage[T1]{fontenc}
\usepackage{lmodern}
\usepackage{graphicx}
\usepackage[figurename=Fig.,labelfont=bf,labelsep=period]{caption}
\usepackage{amsmath}
\usepackage{amsfonts}
\usepackage{amssymb}
\usepackage{amsbsy}
\usepackage{newtxtext,newtxmath}
\usepackage[colorlinks=true,citecolor=red,linkcolor=black]{hyperref}
\usepackage{nameref}
\usepackage{bm}
\usepackage{mathrsfs}
\usepackage{amsfonts}
\usepackage{algorithm}
\usepackage[noend]{algpseudocode}
\usepackage{color}
\usepackage{setspace}
\usepackage{import}
\usepackage{url}
\usepackage{multicol}
\usepackage{subfigure}
\usepackage{amsmath,bbm,mathtools,bm}
\usepackage{multicol}
\usepackage{multirow}
\usepackage{xcolor}
\usepackage{comment}

\def\omg{{\Omega}}
\newcommand{\mcU}{\mathcal{U}}

\newcommand{\mcB}{\mathcal{B}}

\newcommand{\mcD}{\mathcal{D}}

\newcommand{\mcN}{\mathcal{N}}
\newcommand{\mcL}{\mathcal{L}}

\def \Cb{\mathbf{C}}

\def \vb{\mathbf{v}}

\def \real{\mathbb{R}}

\newcommand{\verti}[1]{{\left\vert #1
    \right\vert}}  
    
\newcommand{\vertii}[1]{{\left\vert\left\vert #1
    \right\vert\right\vert}}

\title{Embedded Nonlocal Operator Regression (ENOR): Quantifying model error in learning nonlocal operators}

\author{
 Yiming Fan \\
  Department of Mathematics, \\
  Lehigh University, \\Bethlehem, PA 18015, USA.\\
  \texttt{yif319@lehigh.edu} \\
   \And
 Habib N. Najm \\
  Sandia National Laboratories,\\
  Livermore, CA 94550, USA. \\
  \texttt{hnnajm@sandia.gov}
  \And
 Yue Yu$^*$ \\
  Department of Mathematics, \\
  Lehigh University, \\Bethlehem, PA 18015, USA.\\
  \texttt{yuy214@lehigh.edu} \\
  Corresponding author.\\
  \And
 Stewart Silling \\
  Center for Computing Research, \\
  Sandia National Laboratories, \\
  Albuquerque, NM 87123, USA.\\
  \texttt{sasilli@sandia.gov}
  \And
 Marta D'Elia \\
  Stanford University, \\
  Stanford, CA 94305, USA.\\
  \texttt{marti.delia@gmail.com}
}

\begin{document}
\maketitle
\begin{abstract}
Nonlocal, integral operators have become an efficient surrogate for bottom-up homogenization, due to their ability to represent long-range dependence and multiscale effects. However, the nonlocal homogenized model has unavoidable discrepancy from the microscale model. Such errors accumulate and propagate in long-term simulations, making the resultant prediction unreliable. To develop a robust and reliable bottom-up homogenization framework, we propose a new framework, which we coin Embedded Nonlocal Operator Regression (ENOR), to learn a nonlocal homogenized surrogate model and its structural model error. This framework provides discrepancy-adaptive uncertainty quantification for homogenized material response predictions in long-term simulations. The method is built on Nonlocal Operator Regression (NOR), an optimization-based nonlocal kernel learning approach, together with an embedded model error term in the trainable kernel. Then, Bayesian inference is employed to infer the model error term parameters together with the kernel parameters. To make the problem computationally feasible, we use a multilevel delayed acceptance Markov chain Monte Carlo (MLDA-MCMC) method, enabling efficient Bayesian model calibration and model error estimation. 
%to sample the posterior of the on parameters involved in the nonlocal constitutive law, and associated modeling discrepancies relative to higher fidelity computations. 
We apply this technique to predict long-term wave propagation in a heterogeneous one-dimensional bar, and compare its performance with additive noise models. 
Owing to its ability to capture model error, the learned ENOR achieves improved estimation of posterior predictive uncertainty.
%Several experiments are conducted to investigate the uncertainty in this problem. To the best of our knowledge, only a few literatures have been working on the statistical calibration on the homogenization using a nonlocal model. This work is an improvement and extension in this direction, focused on embedded model error representation.
\end{abstract}

% keywords can be removed

\tableofcontents

\section{Introduction}

In many real-world applications, the analyzed physical system is a complex multi-scale process, which often starts from a sequence of mechanical cues at the microscale, and results in the propagation of properties at the macroscle. However, microscale simulations are often infeasible due to high computational costs. To mitigate this challenge, upscaled models are often needed, which can be employed in large scale long-term simulations to inform decision making. For this purpose, various multiscale approaches and homogenized surrogate models were developed \cite{zohdi2017homogenization,bensoussan2011asymptotic,weinan2003multiscale,efendiev2013generalized,junghans2008transport,kubo1966fluctuation,santosa1991dispersive,dobson2010sharp,ortiz1987method,moes1999simplified,hughes2004energy}. Examples include partial differential equation (PDE)-based approaches with effective coefficients \cite{milton2022theory,chapman2021homogenization}, particle-based methods that replace clusters of particles with larger particles \cite{espanol1995statistical,groot1997dissipative,hoogerbrugge1992simulating}, nonlocal models that capture long-range microscale effects via integral operators \cite{beran70,cher06,karal64,rahali15,silling2021propagation,you2021md,you2023towards,you2024nonlocal,zhang2022metanor,tadmor2014critical,du2020multiscale}, and several others.
%In the last few decades, nonlocal models have become popular among various fields of science, such as material science, engineering, image processing, stochastic process and etc \cite{silling2000reformulation,gilboa2009nonlocal,Scalas2000,DElia2017,Meerschaert2012,jafarzadeh2024heterogeneous}. 
Among these approaches, nonlocal models \cite{silling2000reformulation,gilboa2009nonlocal,Scalas2000,DElia2017,Meerschaert2012,silling2024peridynamic,jafarzadeh2024heterogeneous} stand out as relatively recent yet powerful tools \cite{beran70,cher06,karal64,rahali15,smy00,willis85,eringen1972nonlocal,bobaru2016handbook,du2020multiscale} as their integral nature naturally embeds length and time-scales in their definition \cite{silling2000reformulation} allowing them to capture macroscale effects induced by a heterogeneous microscale behavior.

However, despite being appropriate for homogenization tasks, nonlocal models present several challenges. As an example, in heterogeneous materials modeling, the material's response depends on mechanical and microstructural properties, and hence requires a material-specific homogenized nonlocal model. In practice, a nonlocal model is determined by the definition of its ``kernel'' whose functional form and associated parameters are not known \emph{a priori}. Thus, to accurately capture the effects of micro-scale heterogeneities at the macroscale and provide reliable and trustworthy predictions, it is necessary to estimate appropriate nonlocal kernels. The paper \cite{You2021} introduces the nonlocal operator regression (NOR) approach, a data-driven technique for the identification of the nonlocal kernel that best describes a system at the macroscale. 
We refer the reader to \cite{You2021,you2020data,you2021md,xu2021machine,jafarzadeh2024peridynamic,de2023machine,xu2022sub,yu2024nonlocal} for several examples of machine learning-based design of homogenized nonlocal operators, and to \cite{lu2022nonparametric,zhang2022metanor} for the rigorous analysis of its learning theory. 

While NOR provides accurate and reliable deterministic nonlocal homogenization models from data, it introduces unavoidable modeling errors due to the discrepancy between the ``surrogate'' nonlocal model and the ground truth system. In long-term simulations, such a modeling error may propagate and accumulate over time. Hence, it is desired to characterize the discrepancy between the nonlocal model and the true governing equations to estimate the predictions' uncertainty. In our prior work \cite{fan2023bayesian}, we introduced the Bayesian nonlocal operator regression (BNOR) approach, which combines Bayesian inference \cite{Laplace:1814,Casella:1990,Jaynes:2003,Robert:2004,Sivia:2006,Carlin:2011,Kennedy:2001,Higdon:2003,Oliver:2011,Cui:2016b,hakim2018probabilistic,Huan:2018b} and NOR to estimate model uncertainty under the assumption of additive independent Gaussian noise in the data.
Specifically, we modeled the discrepancy between high-fidelity data and the NOR model as additive independent identically distributed (\emph{iid}) noise, and tested the method in the context of stress wave propagation in a randomly heterogeneous material~\cite{fan2023bayesian}. However, the additive \emph{iid} noise assumption is more suited for representing measurement noise rather than a modeling error; this happens because the model discrepancy is expected to exhibit a significant degree of spatial correlation depending on the degree of smoothness in the solution and data. %\MD{[This last sentence is unclear, what are we trying to say? A degree of correlation with what?]}. 
Ignoring this correlation mat lead to over-confidence and low posterior uncertainty in the inferred parameters, with consequent over confidence in posterior predictions of the quantities of interest (QoIs)~\cite{bayarri2007framework}. 

To better represent the discrepancies due to model error, state-of-the-art additive model error techniques augment the predictive model's output with a Gaussian process (GP)\cite{kleijnen2009kriging, Kennedy:2001}. The additive GP provides extensive flexibility in correcting model outputs and fitting available data with a parameterized correlation structure, resulting in meaningful predicted uncertainty. However, adding a GP to model outputs in the context of physical systems may compromise adhering to physical constraints and governing equations; in fact, while a physical model prediction $y=f(x)$ is expected to satisfy relevant physical laws, a GP-augmented prediction $y=f(x)+\epsilon(x)$ may not in principle~\cite{sargsyan2015statistical}.

To address the shortcomings of additive GP constructions in physical models, embedded model error constructions were introduced in~\cite{sargsyan2015statistical,sargsyan2019embedded}; here, specific model components are augmented with statistical terms. Specifically, random variables are embedded in the parameterization of certain model elements based on approximations or modeling assumptions. As a result, this approach enables Bayesian parameter estimation that accounts for model error and identifies modeling assumptions that dominate the discrepancies between the fitted model and the data. To our knowledge, embedded model error applications have relied on random variable embeddings and not GP embeddings.

In this work, we introduce a Bayesian calibration technique based on GP embeddings applied in the context of nonlocal homogenization. In particular, we augment the nonlocal kernel with a GP to represent model error in the kernel. Then, we use Bayesian inference to learn the posterior probability distributions of parameters of the nonlocal constitutive law (the nonlocal kernel function) and the GP simultaneously. To make the solution of the resulting Bayesian inference problem feasible, we employ the Multilevel Delayed Acceptance (MLDA) MCMC ~\cite{lykkegaard2020adaptivemlda,lykkegaard2023multilevel} method, which exploits a hierarchy of models with increasing complexity and cost. 
We illustrate this method using a one-dimensional nonlocal wave equation that describes the propagation of stress waves through an elastic bar with a heterogeneous microstructure.

We summarize our contributions in the following list.
\begin{itemize}
    \item We propose the Embedded Nonlocal Operator Regression (ENOR) approach, which consist of augmenting a nonlocal homogenized model with modeling error with the purpose of capturing discrepancy of the predictions with respect to high fidelity data. In particular, we embed a GP in the nonlocal kernel to characterize the spatial heterogeneity in the microscale. 
    \item The model is inferred from high-fidelity microscale data using MCMC, which learns the parameters of the nonlocal kernel and the correction term jointly. To alleviate the cost associated with costly likelihood estimations in MCMC, we use a multi-level delayed acceptance MCMC formulation.
    \item To illustrate the efficacy of our method, we consider the stress wave propagation problem in a heterogeneous one-dimensional bar. A data-driven nonlocal homogenized surrogate is obtained, together with a GP representing the kernel correction. Compared to the previous work which learns an additive noise~\cite{fan2023bayesian}, the embedded model error correction formulation successfully captures the discrepancy and provides an improved estimation of the posterior predictive uncertainty.
\end{itemize}

\textbf{Paper outline. }
In Section~\ref{sec:background}, we review the nonlocal operator regression approach and the general embedded model error representation formulation. Then, in Section~\ref{sec:BNOREM} we propose the ENOR approach, together with a MLDA-MCMC formulation which uses a low-fidelity model to accelerate the expensive likelihood estimation. In Section~\ref{sec:app}, we examine the posterior distribution and the prediction given by the algorithm. Section~\ref{sec:conclusion} concludes the paper with a summary of our contributions and a discussion of potential follow-up work. 

\section{Background and Related Mathematical Formulation}\label{sec:background}

Consider the numerical solution of a wave propagation problem in the spatial domain $\bar{\Omega}\subset \real^d$ and time domain $[0,T]$, where $d$ is the physical dimension. Given $S$ observations of forcing terms $f^s(x,t)$, we define the corresponding computed high-fidelity displacement fields $\mcD:=\{u^s_{DNS}\}_{s=1}^S$ as the ground-truth dataset. Here we assume that both $f^s$ and $u^s_{DNS}$ are provided at time instants $t^n\in[0,T]$ and grid points $x_i\in\bar{\Omega}$. Without loss of generality, we assume the time and space points to be uniformly distribute, i.e. in one dimension, the spatial grid size $\Delta x$ and time step size $\Delta t$ are constant. We denote the collection of all grid points as $\chi=\{x_i\}_{i=1}^L$. We claim that for the same forcing terms $f^s(x,t)$, the corresponding solution, $u_{NL}^s(x,t)$, $(x,t)\in\bar{\Omega}\times[0,T]$, of a nonlocal model provides an approximation of the ground-truth data, i.e., $u_{NL}^s(x,t)\approx u_{DNS}^s(x,t)$.  The purpose of this work is to develop a Bayesian inference framework, which learns the nonlocal model from $\mcD$ together with its structural error.

Throughout this paper, for any vector $\vb=[v_1,\cdots,v_q]\in\real^q$, we use $\vertii{\vb}_{l^2}:=\sqrt{\sum_{i=1}^qv_i^2}$ to denote its $l^2$ norm.
For a function $u(x,t)$ with $(x,t)\in\bar{\Omega}\times[0,T]$, its discrete $l^2$ norm is defined as
$$\vertii{u}_{l^2(\bar{\Omega}\times[0,T])}:=\sqrt{\Delta t\Delta x\sum_{n=0}^{T/\Delta t} \sum_{x_i\in\chi} (u(x_i,t^n))^2},$$
which can be interpreted as a numerical approximation of the $L^2(\bar{\Omega}\times[0,T])$ norm of $u$. 

\subsection{Nonlocal Operator Regression}\label{sec:background_nor}

In this work, we consider the nonlocal operator regression (NOR) approach first proposed in \cite{you2020data} and later extended in \cite{You2021,you2021md,lu2022nonparametric,zhang2022metanor}. NOR aims to find the best homogenized nonlocal ``surrogate'' model from experimental and/or high-fidelity simulation data pairs. Herein, we focus on the latter scenario. Given a forcing term $f(x,t)$, $(x,t)\in\bar{\Omega}\times[0,T]$ and appropriate boundary and initial conditions, we denote the high-fidelity (HF) model as 
\begin{equation}
    \dfrac{\partial^2u_{HF}}{\partial t^2}(x,t)-\mathcal{L}_{HF}[u_{HF}](x,t)=f(x,t). 
\end{equation}
$\mathcal{L}_{HF}$ is the HF operator and $u_{HF}(x,t)$ is the solution of fine-scale simulations. NOR assumes that nonlocal models are accurate homogenized surrogates of the HF model; a general nonlocal model reads as follows:
\begin{equation}\label{eqn:homonl}
    \dfrac{\partial^2u_{NL}}{\partial t^2}(x,t)-\mathcal{L}_{NL}[u_{NL}](x,t)=f(x,t),  
\end{equation}
where the nonlocal operator $\mathcal{L}_{NL}$ is an integral operator of the form
\begin{equation}\label{eqn:intoperator}
    \mathcal{L}_{NL}:=\mcL_{K_\Cb}[u](x,t)=\int_{\bar{\omg}\cap B_\delta(x)} K_{\Cb}(x,y)(u(y,t)-u(x,t))dy.
\end{equation}
Here, $B_\delta(x):=\{y\in\real^d,\;|x-y|<\delta\}$ denotes the interaction neighborhood of the material point $x$. 
%$\bar{\omg}:=\omg\cup\omg_\delta$ is the domain augmented with nonlocal boundary region 
We further denote 
$\omg_\delta:=\{x\in\bar{\omg}\,|\,\text{dist}(x,\real^d\backslash\bar{\omg})<\delta\}$ is the nonlocal boundary region, $\omg:=\bar{\omg}\backslash\omg_\delta$ as the ``interior'' region inside the domain. $\Cb:=\{C_m\}_{m=0}^M$ is the set of parameters that uniquely determines the kernel $K_{\Cb}$. Here, a radial kernel, $K_{\Cb}(x,y):=K_{\Cb}(|y-x|)$, is employed. This widely adopted setting guarantees the symmetry of the integrand in \eqref{eqn:intoperator} with respect to $x$ and $y$ and induces the conservation of linear momentum and Galilean invariance \cite{bobaru2016handbook,jafarzadeh2024peridynamic,liu2023ino,liu2024harnessing}. In fact, such property allows us to write:
\begin{equation}\label{eqn:symm}
K_{\Cb}(x,y)(u(y,t)-u(x,t))=\dfrac{1}{2}\left[K_{\Cb}(|x-y|)(u(y,t)-u(x,t))-K_{\Cb}(|y-x|)(u(x,t)-u(y,t))\right].
\end{equation}
As done in previous works\cite{fan2023bayesian}, we represent the kernel using a linear combination of Bernstein-polynomials, i.e.,
\begin{equation}\label{model-kernel}
\begin{aligned}
K_\Cb\left(\frac{|z|}{\delta}\right) 
& = \sum_{m=0}^{M}\frac{C_m}{\delta^{d+2}} B_{m,M}
    \bigg (\bigg|\frac{z}{\delta}\bigg|\bigg),\ \ \ 
    B_{m,M}(z) = \binom{M}{m} %\begin{pmatrix} M\\ m\\ \end{pmatrix}
z^m(1-z)^{M-m}\;\; \text{ for } 0\leq z\leq 1.
\end{aligned}
\end{equation}
Note that this construction makes $K_\Cb$ continuous, radial and compactly supported on the ball of radius $\delta$ centered at $x$, also guarantees that \eqref{eqn:homonl} is well-posed \cite{du2017peridynamic}.  %\MD{[Shouldn't we add assumptions on the coefficients?]} \YF{[YF: I'm wondering what's the specific assumption? distribution, range, independencies, etc? For the physical constraints, we have that in Eq.9 and Eq.10.]} \MD{[Conditions for well-posedness.]} \YF{[By construction of the kernel, the nonlocal equation is guaranteed to be well-posed. We used this conclusion directly in\cite{You2021,fan2023bayesian}.]} \MD{[i don't see conditions on the sign]}

To find the nonlocal model that best describes the high-fidelity data, we aim to find the kernel parameters $\Cb$ such that for given loadings $f^s(x,t)$, the corresponding nonlocal solutions $u_{NL,\Cb}^s(x,t)$ in \eqref{eqn:homonl} are as close as possible to the HF solution $u_{DNS}^s(x,t)$. To numerically solve \eqref{eqn:homonl}, we use a mesh-free discretization of the nonlocal operator and a central difference scheme in time, i.e.
\begin{align}
\nonumber(u^s_{NL,\Cb})_{i}^{n+1} &:= 2(u^s_{NL,\Cb})_{i}^n-(u^s_{NL,\Cb})_{i}^{n-1}+\Delta t^2f^s(x_i,t^n)+\Delta t^2 \left(\mcL_{K_\Cb,h}[u^s_{NL,\Cb}]\right)_i^n\\
\nonumber &= 2(u^s_{NL,\Cb})_{i}^n-(u^s_{NL,\Cb})_{i}^{n-1}+\Delta t^2f^s(x_i,t^n)\\
&+\Delta t^2\Delta x\sum_{x_j\in B_\delta(x_i)\cap\chi} K_\Cb(|x_j-x_i|)((u^s_{NL,\Cb})_{j}^n-(u^s_{NL,\Cb})_{i}^n),\label{eqn:numerical_u}
\end{align}
where $(\tilde{u}^s_{NL,\Cb})_{i}^{n}$ denotes the nonlocal solution at $(x_i,t^n)$, and $\mcL_{K_\Cb,h}$ is an approximation of $\mcL_{K_\Cb}$ by the Riemann sum with uniform grid spacing $\Delta x$.
The optimal parameters $\Cb^*$ can be obtained by solving the following optimization problem
\begin{align}
\Cb^*=\underset{\Cb}{\text{argmin}}&\sum_{s=1}^S\dfrac{\vertii{u_{NL,\Cb}^s-u_{DNS}^s}_{l^2(\Omega\times[0,T])}^2}{\vertii{u_{DNS}^s}_{l^2(\Omega\times[0,T])}^2}+\lambda \vertii{\Cb}_{l^2}^2,\label{eqn:opt}\\
\text{s.t. }&  K_\Cb \text{ satisfies physics-based constraints. }\label{eqn:optcond2}
\end{align}
Here, $\lambda$ is a regularization parameter added to guarantee the well-conditioning of the optimization problem. The physics-based constraints depend on the nature of the problem, in the case of stress waves, we impose homogenized properties of plane waves propagating at very low frequency \cite{You2021}, i.e. 
\begin{align}
    &\int_{B_\delta(x)}\verti{y-x}^2K_\Cb(\verti{y-x})dy=2\rho c_0^2, \label{eqn:constraints1}\\
    &\int_{B_\delta(x)}\verti{y-x}^4K_\Cb(\verti{y-x})dy=-8\rho c_0^3R.\label{eqn:constraints2}
\end{align}
The density $\rho$, the effective wave speed $c_0$ and $R$, the second derivative of the wave group velocity with respect to the frequency $\omega$ evaluated at $\omega=0$, are obtained the same way as in \cite{fan2023bayesian}, for both periodic and random materials. Numerically, we impose the physics constraints by approximating \eqref{eqn:constraints1} and \eqref{eqn:constraints2} via a Riemann sum. We note that these constraints can be imposed explicitly so that we can substitute the expression of $C_{M-1}$ and $C_M$, making \eqref{eqn:opt} an unconstrained optimization problem of $\{C_m\}_{m=0}^{M-2}$ \cite{zhang2022metanor}.

We note that in \eqref{eqn:numerical_u}, the numerical approximation $(u^s_{NL,\Cb})^{n+1}$ is calculated based on the numerical approximation of the last time instance $(u^s_{NL,\Cb})_{i}^{n}$. As such, the numerical error accumulates as $n$ increases, and \eqref{eqn:opt} aims to minimize the accumulated error. Alternatively, one can also choose to minimize the step-wise error \cite{lu2022nonparametric,zhang2022metanor}, 
%the step-wise temporal error nonlocal model 
by replacing the approximated solution at the last time instances, $(u^s_{NL,\Cb})^n$ and $(u^s_{NL,\Cb})^{n-1}$, with the corresponding ground-truth data $(u^s_{DNS})^n$ and $(u^s_{DNS})^{n-1}$ in \eqref{eqn:numerical_u}:
\begin{align}
\nonumber(&\tilde{u}^s_{NL,\Cb})_{i}^{n+1}:= 2(u^s_{DNS})_{i}^n-(u^s_{DNS})_{i}^{n-1}+\Delta t^2f^s(x_i,t^n)+\Delta t^2 \left(\mcL_{K_\Cb,h}[u^s_{DNS}]\right)_i^n\\
\nonumber=&2(u^s_{DNS})_{i}^n-(u^s_{DNS})_{i}^{n-1}+\Delta t^2f^s(x_i,t^n)+\Delta t^2\Delta x\sum_{x_j\in B_\delta(x_i)\cap\chi} K_\Cb(|x_j-x_i|)((u^s_{DNS})_{j}^n-(u^s_{DNS})_{i}^n).\label{eqn:numerical_nl}
\end{align}
This formula provides a faster approximated nonlocal solution at the cost of physical stability, which plays a critical role in long-term prediction. Therefore, in this work we aim to minimize the accumulated error by considering \eqref{eqn:numerical_u}.

\subsection{Embedded Model Error}
In this section we introduce the embedded model error procedure developed in \cite{sargsyan2015statistical,sargsyan2019embedded}, which relies on a Bayesian inference framework for the estimation of the model error, with the identification of the contribution of different error sources to predictive uncertainty. This is a natural setting for calibrating a low fidelity (LF/here NL) model against a higher fidelity (HF/here DNS) model accounting for uncertainty. Here, we assume that any discrepancy between the two models' predictions is due to model error, and not to other sources such as measurement error. Thus, to model this discrepancy, we do not pursue traditional approaches that introduce independent identically distributed (\emph{iid}) additive Gaussian noise. Kennedy and O'Hagan~\cite{Kenn:OHag:2001} introduced the use of an additive Gaussian process (GP) to capture the structure of the discrepancy between two models, where the flexibility of the GP allows one to adequately capture the discrepancy in the predictions induced by the model error. 

In this setting, denoting by $q_j$ the data generated from the HF/DNS model $\eta(o_j)$, and by $\zeta(o_j,\Lambda)$ the one generated from the LF/NL model, where $\Lambda$ is a set of model parameters to be estimated, at operating conditions $o_j$ (e.g. spatial or temporal coordinates), for $j=1,2,...,N$ observations, the additive GP construction can be written as  
\begin{equation}
q_j=\zeta(o_j,\Lambda)+\Delta(x_j,\alpha,\xi)    
\end{equation}
where $\Delta(x_j,\alpha,\xi)$ is a GP with parameters $\alpha$ evaluated at the discrete spatial locations $x_j$, and with \emph{iid} standard normal stochastic degrees of freedom $\xi$. This representation and the related approach have found extensive use, \emph{e.g.} \cite{Kenn:OHag:Higg:2002,Higd:2004,Bayarri:2009a}, as they allow for avoiding the overconfidence that comes from ignoring model error in model calibration whether against another model or actual observations. However, the additive GP construction presents some challenges in calibrating physical models \cite{sargsyan2015statistical}, which motivated embedding model error terms into the LF model. In our present notation, this can be written as
\begin{equation}
    q_j\approx h(o_j,\tilde{\alpha},\xi):=\zeta(o_j,\Lambda,\Delta(x_j,\alpha,\xi)),\qquad\qquad  j=1...N
\label{eq:moderr}
\end{equation}
where  $\tilde{\alpha}=(\Lambda, \alpha)$. We note that in the literature the use of model error embedding has relied on a simplified version of the above, where the error term is the random variable $\Delta(\alpha,\xi)$, {\it i.e.}, lacking the spatial dependence, rather than a GP \cite{sargsyan2015statistical,sargsyan2019embedded,Hakim:2018,Huan:2018,Hegde:2024}. This simplification might be necessary because of lack of data or computational constraints and it comes with loss of flexibility in the model error representation. In this work, we retain the full GP formalism, as in \eqref{eq:moderr}. It is also important to point out that one key benefit from model error embedding is that the analyst, knowing where approximations have been made in the computational model at hand, can embed model error terms as diagnostic instruments in different parts of the model. Identifying the structure of the discrepancy between the two models by model error embedding, can highlight the modeling assumptions that are likely the dominant source of predictive discrepancy. Similarly, the specific form of embedding can be employed as a diagnostic instrument to identify, \emph{e.g.} the quality of one submodel correction versus another. 

\subsection{Likelihood Construction}

In order to find the posterior distribution of the model parameters, we use Bayesian inference to estimate $\tilde{\alpha}$. Given data $\mathcal{D}$, we write Bayes' rule, expressing $p(\tilde{\alpha}|\mathcal{D})$, the posterior density of $\tilde{\alpha}$ conditioned on $\mathcal{D}$, as 
\begin{equation}\label{eqn:bayes}
p(\tilde{\alpha}|\mathcal{D})=\dfrac{p(\mathcal{D}|\tilde{\alpha})p(\tilde{\alpha})}{p(\mathcal{D})},
\end{equation}
where $p(\mathcal{D}|\tilde{\alpha})$ is the likelihood, $p(\tilde{\alpha})$ is the prior, and $p(\mathcal{D})$ is the evidence which can be treated as a constant in the parameter estimation context. Implicit in the above is also the conditioning on the NL model being fitted, which we leave out for convenience of notation. 
%\MD{[it's ok to leave it out but if we make this statement we have to explain it]} \YF{[Just a notation for conditioning on data and model. We always have conditioning ofn model. Here, we're not comparing models so we don't need to specify.]} 
A key step in obtaining the posterior distribution is the construction of a justifiable likelihood. In general, this is a significant challenge~\cite{sargsyan2015statistical,sargsyan2019embedded} 
%\MD{[even if it's discussed elsewhere, this is an important part and the reader should know what the challenges are and why we're following a certain approach]} \YF{[This is a general statement. It's always challenging to find an appropriate likelihood, as well as modeling challenge. We modified the phrases in this sentence to avoid confusion. ]} 
where alternate approximations are possible. One convenient approximation, which we choose here, is to use Approximation Bayesian Computation (ABC), a likelihood-free method~\cite{Beaumont:2002,Marjoram:2003,Sisson:2011} that is often necessary to deal with the challenge of computing expensive/intractable likelihoods. Rather than relying on a likelihood to provide a measure of agreement between model predictions and data, ABC methods rely on a measure of distance between summary statistics evaluated from the two data sources. With the summary statistics on the model output $S_h$, and those estimated from the data $S_q$, ABC relies on a kernel density $g(z)$ (a Gaussian), a distance metric $d(S_h,S_q)$, and a tolerance parameter $\epsilon$ to provide a pseudo-likelihood:
$$\mathcal{L}(\tilde{\alpha})=\epsilon^{-1}g(\epsilon^{-1}d(S_h,S_q))=\dfrac{1}{\epsilon\sqrt{2\pi}}\exp\left(-\dfrac{d(S_h,S_q)^2}{2\epsilon^2}\right). $$
Here, for the definition of the distance, we follow \cite{sargsyan2015statistical}. We consider the mean $\mu_j=\mathbb{E}_\xi[h(o_j,\tilde{\alpha},\xi)]$ and standard deviation $\sigma_j=\sigma_\xi[h(o_j,\tilde{\alpha},\xi)]$ statistics from the computational model predictions and subtract them to the data $q_j$ and a scaled absolute difference between the mean prediction and the data $\gamma|\mu_j-q_j|$ respectively, where $\gamma$ is a user-defined parameter. With this, the ABC likelihood reads
\begin{equation}
    \mathcal{L}(\tilde{\alpha})=\dfrac{1}{\epsilon\sqrt{2\pi}}\prod_{j=1}^N\exp(-\dfrac{(\mu_j-q_j)^2+(\sigma_j-\gamma|\mu_j-q_j|)^2}{2\epsilon^2}).
    \label{eq:abc}
\end{equation}
The motivation for this construction is the desire to require the Bayesian-calibrated model to achieve two goals: (1) fit the data in the mean, and (2) provide a degree of predictive uncertainty that is consistent with the spread of the data around the mean prediction. In particular, this second requirement provides protection against overconfidence in predictions, ensuring that predictive uncertainty is representative of the discrepancy from the data resulting from model error, irrespective of data size.

\section{Nonlocal Operator Regression with Embedded Model Error}\label{sec:BNOREM}

In this section we introduce the embedded treatment of nonlocal operator regression, along with implementation details. 

\subsection{Mathematical Formulation}\label{sec:math}

We propose the embedded nonlocal operator regression (ENOR) construction, which aims to quantify the model error in nonlocal operator learning using Bayesian inference. In particular, we incorporate a location-dependent GP, $K_\xi(x,\tilde{\omega})$, to represent embedded model error. Here, this Gaussian random field is defined on $\bar{\omg}\times\omg_p$, where $\bar{\omg}$ is the spatial domain together with the nonlocal boundary region, %\MD{[should it be the closure instead?]}\YF{[This is consistent with previous papers. We don't really need closure here because it's just a notation for the physical space.]} 
and $\omg_p$ is the sample space of a probability space. The dependence on the GP parameters is implicit, and is suppressed here for convenience of notation. To account for the structural error of learning a homogenized kernel $K_{\Cb}$, we modify \eqref{eqn:intoperator} as: %\MD{[why this domain of integration? (I'm probably missing something)]} \YF{[We're integrating in the physical domain and the kernel will restrict the integration being done within the horizon]}
\begin{equation}\label{eqn:enloperator}
   \mcL_{ENL}[u](x,t):=\int_{B_\delta(x)} \left(K_{\Cb}(|y-x|)\left(1+K_\xi\left(\frac{x+y}{2},\tilde{\omega}\right)\right)\right)(u(y,t)-u(x,t))dy.
\end{equation}
Here the GP $K_\xi(x,\tilde{\omega})$ is defined by the following covariance function:  
$$\mathrm{Cov}(K_\xi(x),K_\xi(y))=\sigma_{gp}^2\exp\left(-\dfrac{|x-y|}{l_{gp}}\right),$$
with $\sigma_{gp}$ and $l_{gp}$ being learnable parameters. Then, the nonlocal model \eqref{eqn:homonl} is modified as 
\begin{equation}\label{eqn:enl}
    \dfrac{\partial^2 {u}_{ENL}}{\partial t^2}(x,t)-\mcL_{ENL}[u_{ENL}](x,t)=f(x,t).
\end{equation}
Note that the corrected kernel preserves the symmetry property in \eqref{eqn:symm}
$$K_{\Cb}(|y-x|)\left(1+K_\xi\left(\frac{x+y}{2},\tilde{\omega}\right)\right)=K_{\Cb}(|x-y|)\left(1+K_\xi\left(\frac{y+x}{2},\tilde{\omega}\right)\right),$$
and correspondingly the fundamental momentum preserving and invariance properties.

To represent the GP, we use the Karhunen–Lo\`{e}ve expansion~\cite{Karhunen:1946,spanos2007karhunen} (KLE), {\it i.e.},
\begin{equation}
K_\xi(x,\tilde{\omega})=\sigma_{gp}\sum_{n=1}^\infty \sqrt{\lambda_i}\phi_i(x)\xi_i(\tilde{\omega}),
\label{eq:kle}
\end{equation}
where $(\lambda_i,\phi_i)$ are eigenpairs of the kernel function $\exp\left(-\dfrac{|x-y|}{l_{gp}}\right)$, and $\xi_i$ are independent standard normal random variables. The analytical expression of the eigen-pairs can be found in \cite{lucor2004generalized}, written for $i=1,\ldots$ as
\begin{equation}\label{eq:lamphitau}
\begin{split} 
   &\lambda_i=\dfrac{2/l_{gp}}{(1/l_{gp})^2+w_i^2}, \\
   &\phi_i(x)=\tau \left(\cos(w_ix)+\dfrac{1}{l_{gp}w_i}\sin(w_ix)\right), \\
   &\tau=\left\{\dfrac{1}{2}\left(L(1+(\dfrac{1}{l_{gp}w_i})^2)+\dfrac{\sin(2w_iL)}{2w_i}(1-(\dfrac{1}{l_{gp}w_i})^2)-\dfrac{1}{l_{gp}w_i^2}(\cos(2w_iL)-1)\right)\right\}, \\
\end{split}
\end{equation}
where $L$ is the length of the domain, $\tau$ is a normalizer, and the $w_i$ are obtained by solving the following equation
\begin{equation}\label{eqn:eigen_eq}
   \left(w_i^2-(1/l_{gp})^2\right)\tan(w_iL)-2\dfrac{w_i}{l_{gp}}=0, \quad i=1,2,\ldots\,.
\end{equation}
We truncate the summation up to $R$ terms for computational purpose, where $R$ is chosen such that
$$\sum_{i=1}^R \lambda_i\geq 0.9 \sum_{i=1}^\infty \lambda_i.$$
Substituting \eqref{eq:kle} into \eqref{eqn:enloperator}, we obtain:
\begin{equation}
    \dfrac{\partial^2 {u}_{ENL}}{\partial t^2}(x,t)-\int_{B_{\delta}(x)} K_{\Cb}(|y-x|)(1+\sigma_{gp}\mathbf{\Psi}^T\boldsymbol{\xi})(u_{ENL}(y,t)-u_{ENL}(x,t))dy=f(x,t), \label{eqn:enl_kl}
\end{equation}
where 
$$\mathbf{\Psi}\left(\dfrac{x+y}{2}\right):=\left[\sqrt{\lambda_1}\phi_1\left(\dfrac{x+y}{2}\right),...,\sqrt{\lambda_R}\phi_M\left(\dfrac{x+y}{2}\right)\right]^T, \quad \boldsymbol{\xi}:=[\xi_1,...,\xi_R]^T.$$
With the KLE we get a realization of the GP by generating a sample of $\mathbf{\xi}$. Then, the numerical scheme in \eqref{eqn:numerical_u} can be employed to evaluate the solution. Denoting $(u_{ENL,\Cb}^s)^{n+1}_{i,k}:=u_{ENL,\Cb}^s(x_i,t^{n+1},\boldsymbol{\xi}_k)$ as the numerical solution at $(x_i,t^{n+1})$ for the $s$-th sample and $k$-th GP realization, we have
\begin{align}
\nonumber(u&^s_{ENL,\Cb})_{i,k}^{n+1}=2(u^s_{ENL,\Cb})_{i,k}^n-(u^s_{ENL,\Cb})_{i,k}^{n-1}+\Delta t^2f^s(x_i,t^n)\\
&+\Delta t^2\Delta x\sum_{x_j\in B_\delta(x_i)\cap\chi} K_\Cb(|x_j-x_i|)\left(1+\sigma_{gp}\mathbf{\Psi}\left(\dfrac{x_j+x_i}{2}\right)^T\boldsymbol{\xi}_k\right)((u^s_{ENL,\Cb})_{j,k}^n-(u^s_{ENL,\Cb})_{i,k}^n),\label{eqn:numerical_enl}
\end{align}
where $\boldsymbol{\xi}_k$ is the $k$-th realization of the random vector generating the corresponding GP. Similarly, the step-wise version of the embedded nonlocal model can be written as 
\begin{align}
\nonumber(\tilde{u}&^s_{ENL,\Cb})_{i,k}^{n+1}=2(u^s_{DNS})_{i}^n-(u^s_{DNS})_{i}^{n-1}+\Delta t^2f^s(x_i,t^n)\\
&+\Delta t^2\Delta x\sum_{x_j\in B_\delta(x_i)\cap\chi} K_\Cb(|x_j-x_i|)\left(1+\sigma_{gp}\mathbf{\Psi}\left(\dfrac{x_j+x_i}{2}\right)^T\boldsymbol{\xi}_k\right)((u^s_{DNS})_{j}^n-(u^s_{DNS})_{i}^n).\label{eqn:numerical_enl_step}
\end{align}
To effectively evaluate the pseudo-likelihood, we employ the expression~\eqref{eq:abc}. Note that the generated eigenpairs $(\lambda_i,\phi_i)$ are dependent on $l_{gp}$. Since optimizing with respect to $l_{gp}$ together with the other parameters is computationally infeasible, we treat $l_{gp}$ as a tunable hyperparameter, and perform our ENOR algorithm for a fixed $l_{gp}$ at a time to avoid the repeated cost of \eqref{eq:lamphitau} and \eqref{eqn:eigen_eq}. Therefore the enhanced parameter set $\tilde{\alpha}=(\Cb,\sigma_{gp},l_{gp})$ will be reduced to $\tilde{\alpha}=(\Cb,\sigma_{gp})$. For each observation, we have the following ABC likelihood
\begin{align}
\nonumber&\mcL(\tilde{\alpha}):=p(\mcD|(\Cb,\sigma_{gp}))\\
&=\prod_{s,i,n=1}^{S,L,T/\Delta t}\text{exp}(-\dfrac{1}{2\epsilon^2}(\mu_\Cb^s(x_i,t^n)-u_{DNS}^s(x_i,t^n))^2+(\sigma_\Cb^s(x_i,t^n)-\gamma\verti{\mu_\Cb^s(x_i,t^n)-u_{DNS}^s(x_i,t^n)})^2), \label{eqn:ABC}
\end{align}
where 
\begin{align}\label{eqn:moments}
\mu_\Cb^s(x_i,t^n)&=\dfrac{1}{K}\sum_{k=1}^K(u_{ENL,\Cb}^s)^n_{i,k}, \\
\sigma_\Cb^s(x_i,t^n)&=\sqrt{\dfrac{1}{K-1}\sum_{k=1}^K((u_{ENL,\Cb}^s)^n_{i,k}-\mu_\Cb^s(x_i,t^n))^2}
\end{align}
are the sample mean and standard deviation for $(u_{ENL,\Cb}^s)^{n}_{i}$ and $(u_{ENL,\Cb}^s)^{n}_{i}$, for $K$ samples of $\xi$, and with the current value of $\sigma_{gp}$.

In \cite{fan2023bayesian}, we found that a good prior distribution plays a critical role in achieving a fast convergence of the MCMC algorithm and we used the learnt kernel parameter $\Cb_0$ from a deterministic nonlocal operator regression (DNOR) to construct such a prior. To provide a prior on $\Cb$, we use independent standard normal priors on the kernel parameters $C_m\sim\mcN(C_{0,m},\hat{\sigma}^2)$, $m=1,...M-2$, with $\Cb_0$ being the learnt kernel parameter from DNOR, and the standard deviation $\hat{\sigma}$ as a tunable hyperparameter. In practice, since $\sigma_{gp}>0$, we infer $\ln(\sigma_{gp})$ instead of $\sigma_{gp}$. To get an initial state for $\ln(\sigma_{gp})$, we optimize the following 
\begin{equation}
\ln(\sigma_{gp,0}):=\underset{\ln(\sigma_{gp})}{\text{argmin}}\dfrac{1}{2\epsilon^2}\sum_{s,i,n=1}^{S,L,T/\Delta t}(\mu_\Cb^s(x_i,t^n)-u_{DNS}^s(x_i,t^n))^2+(\sigma_\Cb^s(x_i,t^n)- \gamma|\mu_\Cb^s(x_i,t^n)-u_{DNS}^s(x_i,t^n))|^2)
\end{equation}
and assign a uniform prior on $\ln(\sigma_{gp})$, specifically  $\ln(\sigma_{gp})\sim\mcU[\ln(\sigma_{gp}^{lo}),\ln(\sigma_{gp}^{hi})]$. We treat $\sigma_{gp}^{lo}$ and $\sigma_{gp}^{hi}$, the lower and upper bounds on $\sigma_{gp}$, as tunable hyperparameters. Once the proposal for $\ln(\sigma_{gp})$ exceeds the bounds, the log-posterior will be set to $-\infty$ automatically.
Combining the likelihood in \eqref{eqn:ABC} and the prior (with $\ln(\sigma_{gp})\in[\ln(\sigma_{gp}^{lo}),\ln(\sigma_{gp}^{hi})]$), we can finally define the unnormalized posterior $p(\Cb,\sigma_{gp}|\mcD)\propto p(\mcD|\Cb,\sigma_{gp})p(\Cb,\sigma_{gp}))$ and obtain the negative log-posterior after eliminating the constant terms: 
\begin{align}
\dfrac{1}{2\epsilon^2}\sum_{s,i,n=1}^{S,L,T/\Delta t}(\mu_\Cb^s(x_i,t^n)-u_{DNS}^s(x_i,t^n))^2+(\sigma_\Cb^s(x_i,t^n)- \gamma|\mu_\Cb^s(x_i,t^n)-u_{DNS}^s(x_i,t^n))|^2)+\dfrac{\vertii{\Cb-\Cb_0}^2_{l_2}}{2\hat{\sigma}^2}.
\label{eqn:pos}
\end{align}

\subsection{Implementation Details}

\begin{figure}[htp]
    \centering
    \includegraphics[width=.75\columnwidth]{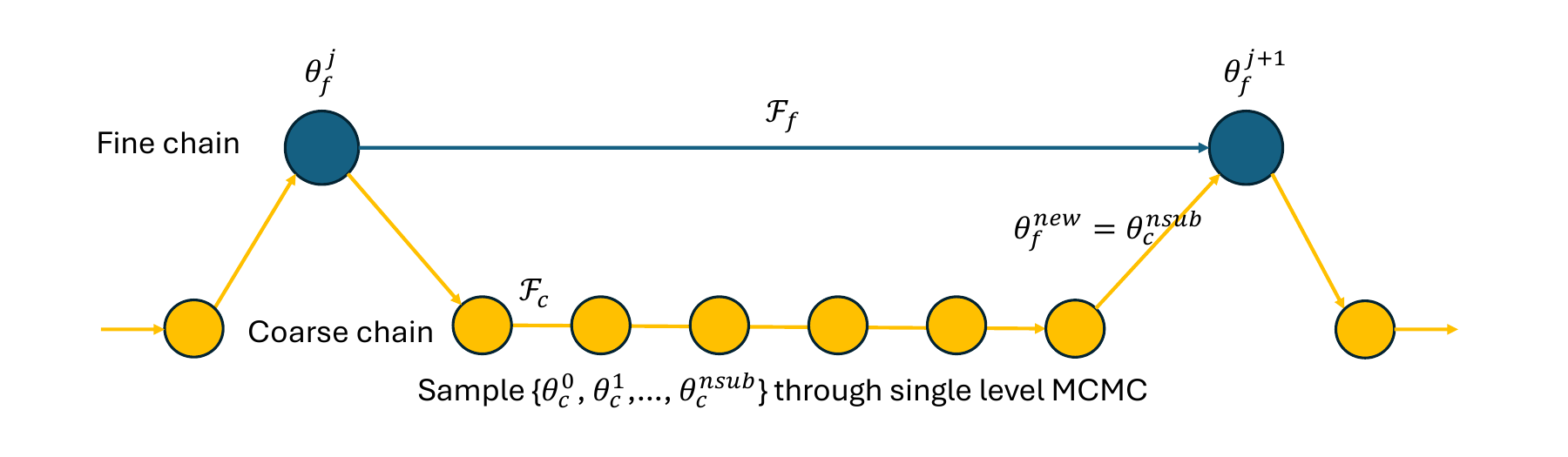}
    \caption{Schematic of generating a proposal $\theta'$ for a two-level MLDA algorithm.}
    \label{fig:mlda}
\end{figure}
As can be seen from \eqref{eqn:pos}, the evaluation of the posterior requires the estimation of the first and second order moments of the output using $K$ GP samples. For each sample, the nonlocal meshfree method \eqref{eqn:numerical_enl} is applied for each spatial point, time step, and observation, making this numerical evaluation expensive in MCMC. In order to improve the efficiency of the MCMC procedure, we employ the Multilevel Delayed Acceptance (MLDA) MCMC technique \cite{lykkegaard2020adaptivemlda,lykkegaard2023multilevel}, which exploits a hierarchy of models of increasing complexity to efficiently generate samples from an unnormalized target distribution. For the purpose of illustration, we summarize the key factors of two-level Delayed Acceptance (TLDA) MCMC here, while the method could be extended to a model hierarchy with arbitrarily many levels by recursion. For the vanilla Metropolis-Hastings (MH) algorithm which is a typical single level MCMC, consider sampling a trace $\{\theta^1,...,\theta^{n_{single}}\}$ from a target distribution $\pi_t(\cdot)$, using a proposal distribution $\pi_p(\cdot|\cdot)$, and an initial state $\theta^0$. MH accepts a new proposal $\theta^{new}$ given $\theta^j$ for $j=0,1,...n_{single}-1$ with probability 
$$\min \left\{1,\dfrac{\pi_t(\theta^{new})\pi_p(\theta^j|\theta^{new})}{\pi_t(\theta^{j})\pi_p(\theta^{new}|\theta^j)}\right\},$$ 
otherwise it rejects $\theta^{new}$ and sets $\theta^{j+1}=\theta^{j}$. In the context of Bayesian inference, the MH target distribution is the posterior distribution, i.e. $\pi_t(\theta|\mcD)=p(\theta|\mcD)$. 
Unlike the vanilla MCMC which has only one model, in TLDA, a cheaper model is employed to reduce the computational cost. Fig.\ref{fig:mlda} illustrates the work flow for TLDA, where the coarse subchains are sampled in order to provide proposals for the fine model. Denote by $\mathcal{F}_f$ the fine forward model and by $\mathcal{F}_c$ the coarse forward model, with $\pi_f$ and $\pi_c$ being their target distributions respectively. Starting from $\theta_f^j$, in the coarse level, one can generate the subchain $\{ \theta_c^1,\theta_c^2,...,\theta_c^{n_{sub}}\}$ of length $n_{sub}$ using MH or any single-level MCMC method. After the subchain is finished, we take $\theta_c^{n_{sub}}$ as the proposal for the fine chain (i.e. $\theta_f^{new}=\theta_c^{n_{sub}}$) and accept this proposal with probability 
\begin{equation}\label{eqn:tlda_prob}
\min\left\{1,\dfrac{\pi_f(\theta_f^{new})\pi_c(\theta_f^j)}{\pi_f(\theta_f^j)\pi_c(\theta_f^{new})}\right\},
\end{equation}
otherwise reject $\theta_f^{new}$ and set $\theta_f^{j+1}=\theta_f^j$~\cite{lykkegaard2023multilevel}. 

When the approximation provided by the coarse model is poor, many samples will be rejected by the fine model resulting in a very low acceptance rate. As outlined in \cite{lykkegaard2023multilevel}, an enhanced Adaptive Error Model (AEM) based on \cite{cui2011bayesian} is useful to account for and correct the discrepancy between the fine and coarse models. We use the two-level AEM~\cite{cui2011bayesian,lykkegaard2023multilevel} in the present use of TLDA. 
For parameters $\tilde{\alpha}$ and operating conditions $o$, the bias $\mathcal{B}(o)$ between the two models can be written as 
\begin{equation}\label{eq:aem}
    \mcB(o)=\mathcal{F}_f(o,\tilde{\alpha},\xi)-\mathcal{F}_c(o,\tilde{\alpha},\xi)
\end{equation}
When the parameter set $\tilde{\alpha}$ is sampled from the prior distribution, then 
$$\mcB(o)\sim\mathcal{F}_f(o,\tilde{\alpha},\xi)-\mathcal{F}_c(o,\tilde{\alpha},\xi).$$ 

Denoting the fine model solution as $u_{ENL,\Cb}^s(x_i,t^n,\boldsymbol{\xi}):=u_{ENL}^s(\Cb,\ln(\sigma_{gp}),\boldsymbol{\xi})^{n}_{i}$, the coarse model (the step-wise nonlocal model) solution as $\tilde{u}_{ENL,\Cb}^s(x_i,t^n,\boldsymbol{\xi}):=\tilde{u}_{ENL}^s(\Cb,\ln(\sigma_{gp}),\boldsymbol{\xi})^{n}_{i}$, and the trainable parameter set as $\tilde{\beta}:=(\Cb,\ln(\sigma_{gp}),\boldsymbol{\xi})$, we have the following sample mean and standard deviation for the correction term 
\begin{align}
\begin{split}\label{eqn:moment_c_to_f}
\hat{\mathbb{E}}[B^{s,n}_{i}]=\mu^s(x_i,t^n)-\tilde{\mu}^s(x_i,t^n), \quad 
(\hat{\sigma}[B^{s,n}_{i}])^2=(\sigma^{s}(x_i,t^n))^2-(\tilde{\sigma}^{s}(x_i,t^n))^2,
\end{split}
\end{align}
where 
\begin{align}\label{eqn:moments_aem}
\begin{split}
\mu^s(x_i,t^n)&=\hat{\mathbb{E}}_{\tilde{\beta}}[u^s_{ENL}(\tilde{\beta})_i^n]=\dfrac{1}{N_0}\sum_{n_0=1}^{N_0}u_{ENL}^s(\tilde{\beta}_{n_0})^{n}_{i},\\
\sigma^s(x_i,t^n)&=\hat{\mathbb{V}}^{1/2}_{\tilde{\beta}}[u^s_{ENL}(\tilde{\beta})_i^n]=\sqrt{\dfrac{1}{N_0-1}\sum_{n_0=1}^{N_0}(u_{ENL}^s(\tilde{\beta}_{n_0})^{n}_{i}-\mu^s(x_i,t^n))^2},\\
\tilde{\mu}^s(x_i,t^n)&=\hat{\mathbb{E}}_{\tilde{\beta}}[\tilde{u}^s_{ENL}(\tilde{\beta})_i^n]=\dfrac{1}{N_0}\sum_{n_0=1}^{N_0}\tilde{u}_{ENL}^s(\tilde{\beta}_{n_0})^{n}_{i},\\
\tilde{\sigma}^s(x_i,t^n)&=\hat{\mathbb{V}}^{1/2}_{\tilde{\beta}}[\tilde{u}^s_{ENL}(\tilde{\beta})_i^n]=\sqrt{\dfrac{1}{N_0-1}\sum_{n_0=1}^{N_0}(\tilde{u}_{ENL}^s(\tilde{\beta}_{n_0})^{n}_{i}-\tilde{\mu}^s(x_i,t^n))^2}.
\end{split}
\end{align}
Here, $N_0$ is the number of samples of $\tilde{\beta}$ which will be used for computing the sample mean and standard deviation. 
We highlight that the enhanced model parameter set $\tilde{\alpha}=(\Cb,\ln(\sigma_{gp}))$ should be sampled simultaneously with $\boldsymbol{\xi}$. 
In contrast, $\mu$ and $\sigma$ are calculated by averaging over multiple samples of $\boldsymbol{\xi}$ only, for each fixed enhanced parameter set $\tilde{\alpha}=(\Cb,\ln(\sigma_{gp}))$ in \eqref{eqn:moments}.
In other words, the sample moments computed in \eqref{eqn:moments_aem} are to be used for correction for any parameter $\tilde{\alpha}$ inside the prior distribution instead of a fixed parameter.

By using the AEM technique, \eqref{eqn:pos} can be well approximated by the coarse model following
\begin{align}
\begin{split}
&\dfrac{1}{2\epsilon^2}\sum_{s,i,n=1}^{S,L,T/\Delta t}(\tilde{\mu}_\Cb^s(x_i,t^n)+\hat{\mathbb{E}}(\mcB^{s,n}_{i})-u_{DNS}^s(x_i,t^n))^2\\
&+(\sqrt{(\tilde{\sigma}_\Cb^s(x_i,t^n))^2+\hat{\mathbb{V}}(B^{s,n}_{i})}-\gamma|\tilde{\mu}_\Cb^s(x_i,t^n)+\hat{\mathbb{E}}(\mcB^{s,n}_{i})-u_{DNS}^s(x_i,t^n))^2|)+\dfrac{\vertii{\Cb-\Cb_0}^2_{l_2}}{2\hat{\sigma}^2}, 
\label{eqn:pos_aem}
\end{split}
\end{align}
where $\hat{\mathbb{E}}(\mcB^{s,n}_{i})$ and $\hat{\mathbb{V}}(\mcB^{s,n}_{i})$ are the sample mean and variance for the correction term $\mcB$ computed using \eqref{eqn:moments_aem}. 
Technically, this step is also a reference for tuning the upper and lower bounds for the prior distribution of $\ln(\sigma_{gp})$. Such an interval should be selected in such a way that at least at $\Cb=\Cb_0$ the log-negative posterior in \eqref{eqn:pos_aem} which is approximated by the coarse model and the AEM could roughly reproduce the log-negative posterior in \eqref{eqn:pos} which is generated using the fine model. We summarize our method in Algorithm \ref{algorithm:1}. 

\begin{algorithm}
\caption{A Two-Phase Learning Algorithm}
{{\begin{algorithmic}[1]
\State{{\bf Find a good initial state}

\textbf{1a)} Learn the estimated kernel parameter, $\Cb_0={C_{0,1},...,C_{0,M-2}}$, by minimizing the time-accumulated error via 
\begin{equation}
\Cb_0:=\underset{\Cb}{\text{argmin}}\sum_{s=1}^S\dfrac{\vertii{u_{NL,\Cb}^s-u_{DNS}^s}_{l^2(\Omega\times[0,T])}^2}{\vertii{u_{DNS}^s}_{l^2(\Omega\times[0,T])}^2}+\lambda\vertii{\Cb}_{l^2}^2.
\end{equation}

\textbf{1b)} With fixed $\Cb=\Cb_0$, find $\ln(\sigma_{gp,0})$ by minimizing 
\begin{equation}
\ln(\sigma_{gp,0}):=\underset{\ln(\sigma_{gp})}{\text{argmin}}\sum_{s,i,n=1}^{S,L,T/\Delta t}((\mu_\Cb^s(x_i,t^n)-u_{DNS}^s(x_i,t^n))^2+(\sigma_\Cb^s(x_i,t^n)- \gamma|\mu_\Cb^s(x_i,t^n)-u_{DNS}^s(x_i,t^n)|)^2).
\end{equation} 
}

\State{{\bf Perform MLDA}

\textbf{2a)}
Using prior $C_m\sim N(C_{0,m},\hat{\sigma}^2)$, $m=0,...,M-2$ and $\ln(\sigma_{gp})\sim \mcU[\ln(\sigma_{gp}^{lo}),\ln(\sigma_{gp}^{hi})]$, compute the sample mean and standard deviation of the correction term $\mcB$ following
\begin{align*}
\begin{split}
\hat{\mathbb{E}}[\mcB^{s,n}_{i}]=\mu^s(x_i,t^n)-\tilde{\mu}^s(x_i,t^n), \quad 
\hat{\mathbb{V}}[\mcB^{s,n}_{i}]=(\sigma^{s}(x_i,t_n))^2-(\tilde{\sigma}^{s}(x_i,t^n))^2
\end{split}
\end{align*}
and tune $[\ln(\sigma_{gp}^{lo}),\ln(\sigma_{gp}^{hi})]$ by recursively doing this step until the loss given by \eqref{eqn:pos_aem} could roughly match the loss given by \eqref{eqn:pos} at $\Cb=\Cb_0$ inside the interval.

\textbf{2b)}
Perform MLDA by evaluating the following negative log-posterior in the coarse level
\begin{align}
\begin{split}
\dfrac{1}{2\epsilon^2}\sum_{s,i,n=1}^{S,L,T/\Delta t}&(\tilde{\mu}_\Cb^s(x_i,t^n)+\hat{\mathbb{E}}(\mcB^{s,n}_{i})-u_{DNS}^s(x_i,t^n))^2+(\sqrt{(\sigma_\Cb^s(x_i,t^n))^2+\hat{\mathbb{V}}[\mcB^{s,n}_{i}]}\\
&-\gamma|\mu_\Cb^s(x_i,t^n)+\hat{\mathbb{E}}(\mcB^{s,n}_{i})-u_{DNS}^s(x_i,t^n))^2|)+\dfrac{\vertii{\Cb-\Cb_0}^2_{l_2}}{2\hat{\sigma}^2}, 
\end{split}
\end{align}
and the following negative log-posterior in the fine level
\begin{align}
\dfrac{1}{2\epsilon^2}\sum_{s,i,n=1}^{S,L,T/\Delta t}(\mu_\Cb^s(x_i,t^n)-u_{DNS}^s(x_i,t^n))^2+(\sigma_\Cb^s(x_i,t^n)- \gamma|\mu_\Cb^s(x_i,t^n)-u_{DNS}^s(x_i,t^n))|^2)+\dfrac{\vertii{\Cb-\Cb_0}^2_{l_2}}{2\hat{\sigma}^2}.
\end{align}
}

\State{{\bf Postprocessing}

Perform convergence check for the parallel chains, extract effective samples from the MCMC chain, analyze the uncertainty of the corresponding solution and other quantities of interests.
}
\end{algorithmic}}}
\label{algorithm:1}
\end{algorithm}

\section{Application: Homogenization for a Heterogeneous Elastic Bar}\label{sec:app}

\begin{figure}[htp]
    \centering
    \includegraphics[width=.65\columnwidth]{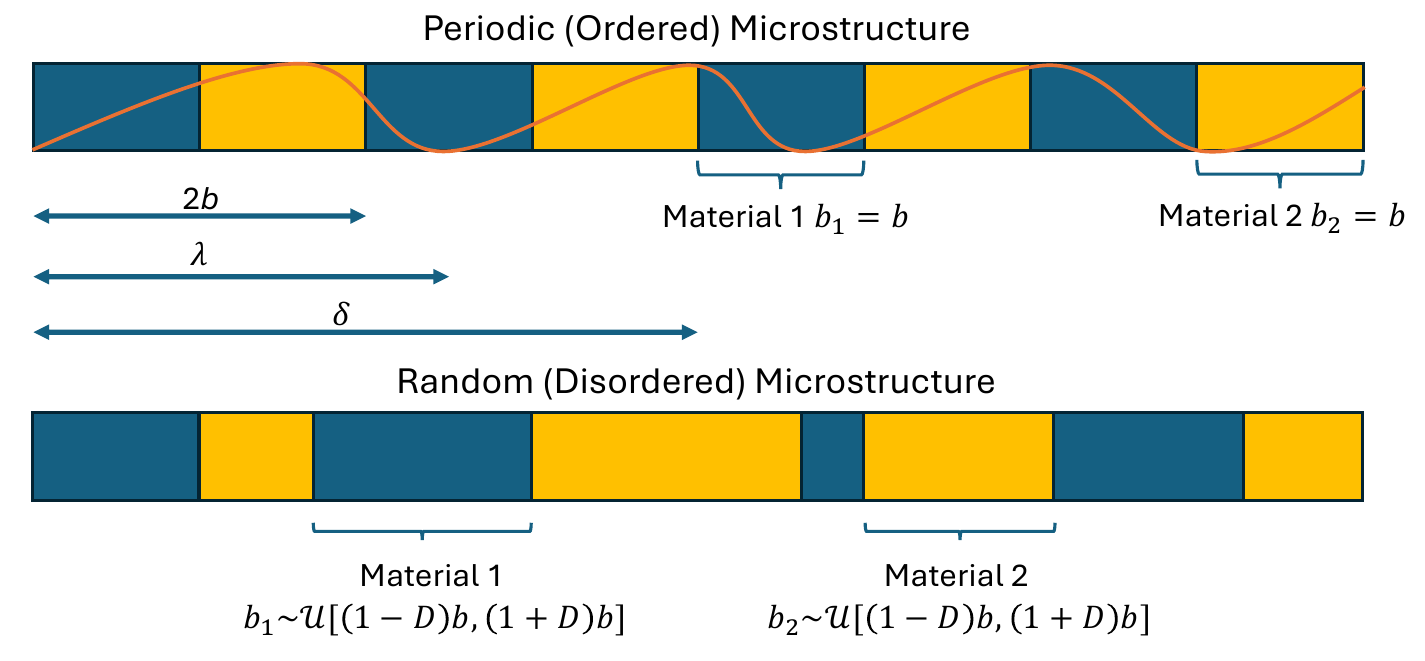}
    \caption{One-dimensional bar composite of material 1 and material 2. (Top) periodic microstructure with fixed layer size$=b$. (Bottom) Random microstructure with layer size satisfying random distribution $\sim\mcU[(1-D)b,(1+D)b]$. }
    \label{fig:bar}
\end{figure}

%\hnn{Here and everywhere below, we should put units on dimensional quantities such as time, length, and frequency, when we report their values.} \YF{[We're following the material settings in \cite{silling2021propagation}, where no units are assigned for the quantities. ]}\YF{[We're using non-dimensionalized settings. Sentence added at the start of sec 4.1.]}

In this section, we examine the efficacy of the proposed ENOR approach on inferring the nonlocal homogenized surrogate for modeling the propagation
of stress waves through a  one-dimensional bar \cite{you2021md,fan2023bayesian}. In particular, we consider a composite bar either made of periodic layers or randomly generated layers (see Fig.\ref{fig:bar}), and we assume that the amplitude of the waves is sufficiently small, so that a linear elasticity model is a valid way to describe the wave motion within the layers and at their interfaces.  With this assumption of linear elastodynamics, the propagation of waves through the bar can be described as:
\begin{equation}\label{eqn:wave}
\dfrac{\partial^2 u}{\partial t^2}(x,t)-\dfrac{1}{\rho}\dfrac{\partial}{\partial x} \left(E(x)\dfrac{\partial u}{\partial x}(x,t)\right)=\dfrac{1}{\rho}f(x,t),
\end{equation}
where $\rho$ is the mass density which is assumed to be constant throughout the body, $u$ is the displacement, and $f$ is the external load density. $E(x)$ is the elastic modulus which varies spatially according to the microstructure, {\it i.e.}, we have $E=E_1$ in the blue regions of Fig.\ref{fig:bar}, and $E=E_2$ in the yellow regions. $s(x,t):=E(x)\frac{\partial u}{\partial x}(x,t)$ is the stress. On the interfaces of two materials, the following jump conditions hold: $[u(x,t)]=0$, $\left[E(x)\dfrac{\partial u}{\partial x}(x,t)\right]=0$.

Ideally, one can solve \eqref{eqn:wave} using numerical solvers on fine discretizations of the computational domain to explicitly represent all interfaces. However, in real-world applications such as projectile impact modeling \cite{silling2021propagation,abrate2003wave}, one is interested in modeling the decay of the wave over distances that are several thousand times larger than the layer size. In these circumstances, a fine numerical solver is prohibitively expensive and a homogenized surrogate model is desired to provide scalable predictions.

In this example, we first generate short-term high-fidelity simulation data by solving \eqref{eqn:wave} using characteristic line method. This method, which we denote as the direct numerical simulation (DNS) technique, assumes that the waves running in the opposite direction converge on the node, and update the material velocity explicitly from the jumping condition which is a consequence of the momentum conservation. Due to this property, this DNS solver is free of truncation error and approximation error as in the classical PDE solver, which allow us to simulate the exact velocity of wave propagation through arbitrarily many microstructural interfaces. We refer the reader to \cite{silling2021propagation,fan2023bayesian} for more details of this method for further details. Then, our goal is to construct a nonlocal homogenized surrogate model from this high-fidelity simulation dataset $\mcD$.

\subsection{Example 1: Periodic Microstructure}

Throughout the section, a non-dimensionalized setting is employed for the physical quantities for the simplicity of numerical experiments, following the setup in \cite{silling2021propagation}. First, we consider a periodic heterogeneous bar, where the layer size for the two materials is a constant $b=0.2$. The bar length is $L=20$ and the physical domain is set to be $[-L/2,L/2]$. Components 1 and 2 have the same density $\rho=1$ and Young's moduli are set as $E_1=1$ and $E_2=0.25$, respectively. For the purpose of training and validation, three types of datasets/settings are considered:
%Note that comparing with \cite{fan2023bayesian}, in type 1 and type 2 data, a shorter bar is employed in this work considering the computational resource while preserving the essence of the training data.
%\subsubsection{Data Generation and Settings}

\medskip\noindent
\textit{Setting 1: Oscillating source (20 loading instances).} We set $L=20$. The bar starts from rest such that $v(x,0)=u(x,0)=0$, and an oscillating loading is applied with $f(x,t) \!= e^{-\left(\frac{2x}{5kb}\right)^2} \!e^{-\left(\frac{t-t_0}{t_p}\right)^2}\!\cos^2\left(\frac{2\pi x}{kb}\right)$ with $k=1,2,\ldots,20$. Here we take $t_0=t_p=0.8$.

\medskip\noindent
\textit{Setting 2: Plane wave with ramp  (11 loading instances).} We also set the domain parameter as $L=20$. The bar starts from rest ($u(x,0)=0$) and is subject to zero loading ($f(x,t)=0$). For the velocity on the left end of the bar, we prescribe  
\[v(-L/2,t)=
\left\{
\begin{aligned}
&\sin(\omega t)\sin^2\left(\frac{\pi t}{30}\right),&t\leq 15\\
&\sin(\omega t),&t>15
\end{aligned}
\right.
\] 
for $\omega=0.35,0.7,\cdots,3.85$.

\medskip\noindent
\textit{Setting 3: Wave packet  (3 loading instances).} We consider a longer bar with $L=266.6$, with the bar starting from rest ($u(x,0)=0$), and is subject to zero loading ($f(x,t)=0$). The velocity on the left end of the bar is prescribed as
$v(-L/2,t)=\sin(\omega t)\exp{(-(t/5-3)^2)}$ with $\omega=2$, $3.9$, and $5$.

For all data types, the parameters for the nonlocal solver and the optimization algorithm are set to $\Delta x= 0.05$, $\Delta t = 0.02$, $\delta =1.2$, and $M = 24$. For training purposes, we generate data of types 1 and 2 till $T = 2$. Then, to investigate the performance of our surrogate model in long-term prediction tasks we simulate till $T=100$ for setting 3.

%\YF{Results to show (comment this out after finished): }
%\begin{enumerate}
%    \item Ablation study: single level MCMC trace plot. (The convergence is slow, cannot be used for further analysis.)
%    \item Convergence check ($\hat{R}$, trace plot, pdf). \YF{(1 $l_{gp}$ for instance)}
%    \item Combined trace (acceptance rate, ess, trace plot, pdf). \YF{(1 $l_{gp}$ for instance)}
%    \item kernel, vg, dispersion and comparison between $l_{gp}$
%    \item PFP(PP) of ENOR and comparison between $l_{gp}$
%    \item validation on wave packet and comparison between $l_{gp}$
%    \item comparison with BNOR
%    \item sensitivity analysis.
%\end{enumerate}

\subsubsection{Results from MCMC Experiments}

\begin{figure}[htp]
\centering
\subfigure[First parameter]{\includegraphics[width=.30\columnwidth]{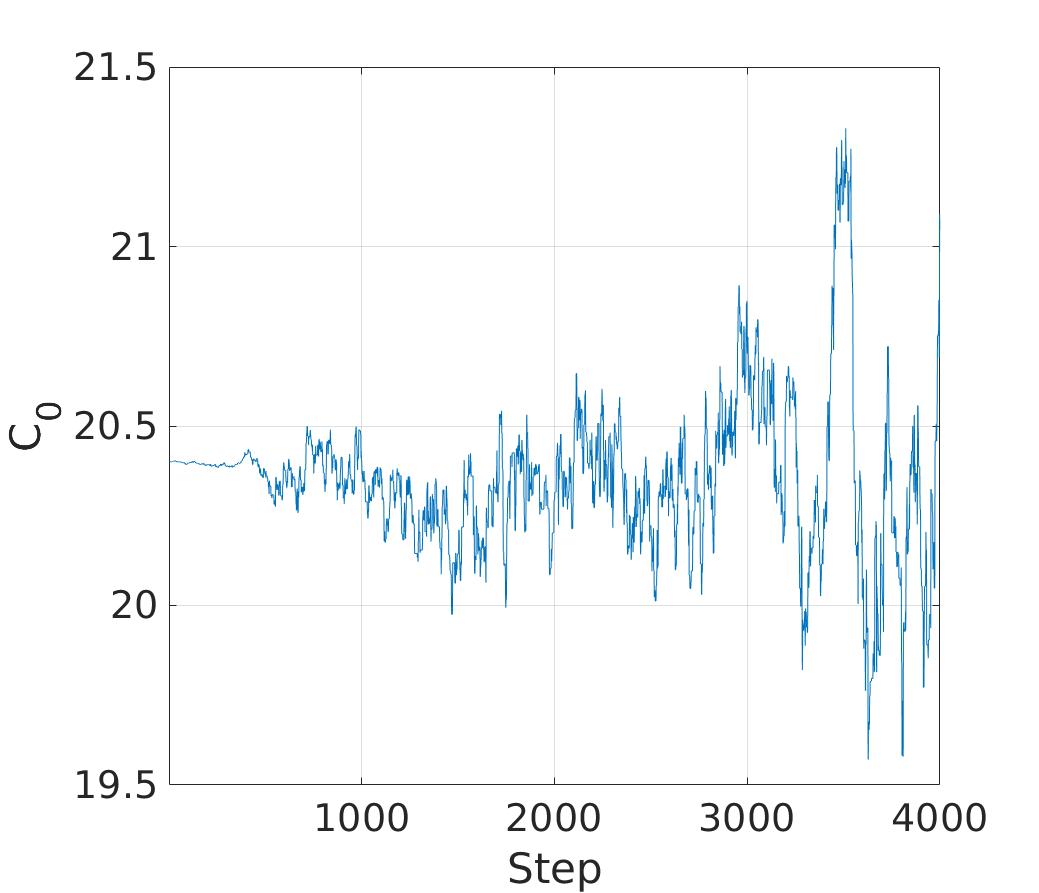}}
\subfigure[Second parameter]{\includegraphics[width=.30\columnwidth]{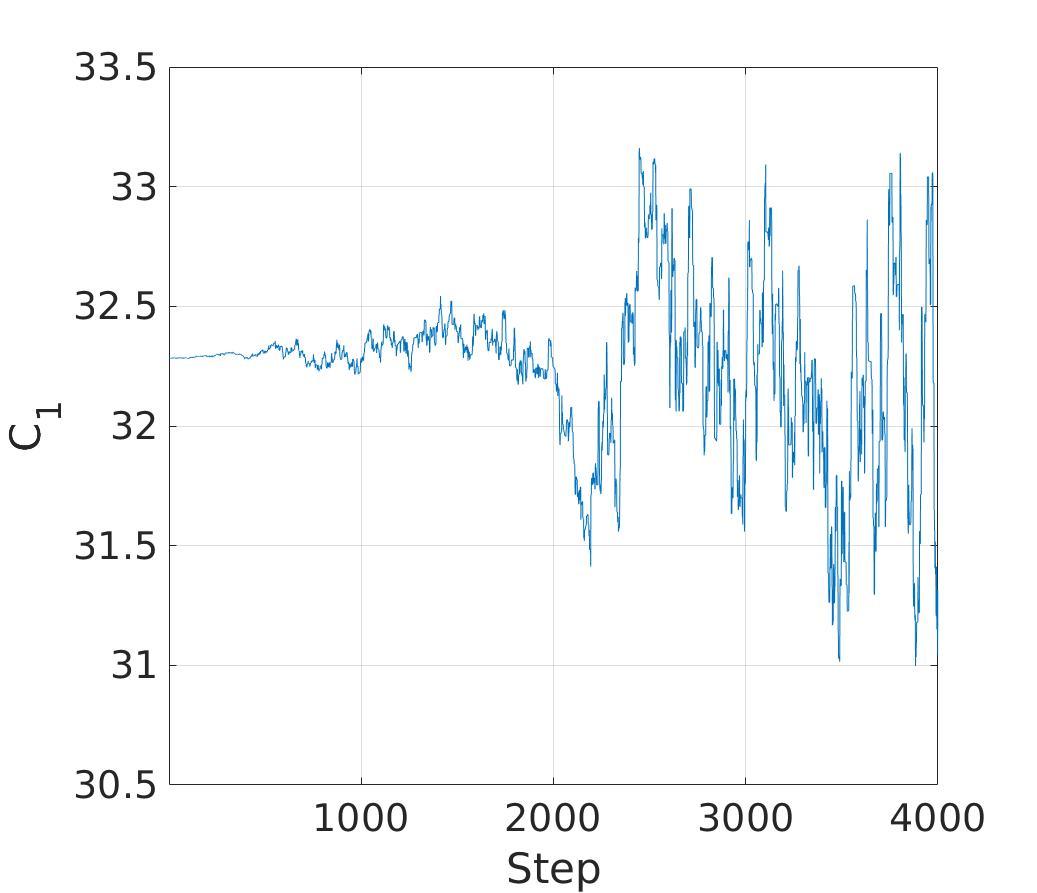}}
\subfigure[$\ln(\sigma_{gp})$]{\includegraphics[width=.30\columnwidth]{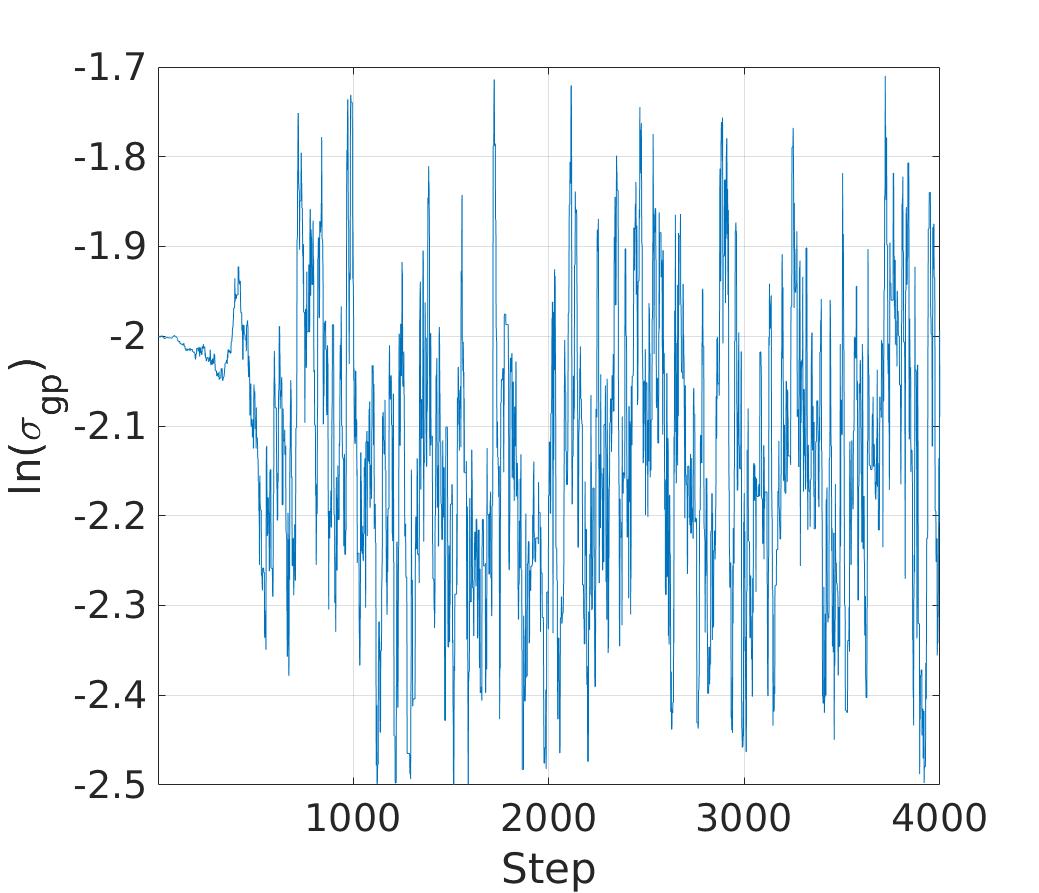}}
\caption{Trace plot for the MCMC using single-level DEMetropolisZ sampler.}
\label{fig:single_level}
\end{figure}

We use PyMC \cite{abril2023pymc} for all the MCMC computations below. We begin by verifying the convergence behavior of our algorithm. In particular, we test a single-level MCMC with the differential evolution Metropolis (DEMetropolis) algorithm \cite{ter2008differential} on the fine level with 4,000 draws and a burn-in stage of 300.  DEMetropolis, or DEMetropolis(Z) is a variation of Metropolis-Hastings algorithm that uses randomly selected draws from the past to make more educated jumps. Results in Fig.~\ref{fig:single_level} indicate that the single-level MCMC suffers from an extremely long burn-in stage and poor mixing. We treat these single-level results as a baseline which we use to examine the relative performance of the multilevel method below.

\begin{figure}[htp]
\centering
\subfigure[First parameter]{\includegraphics[width=.30\columnwidth]{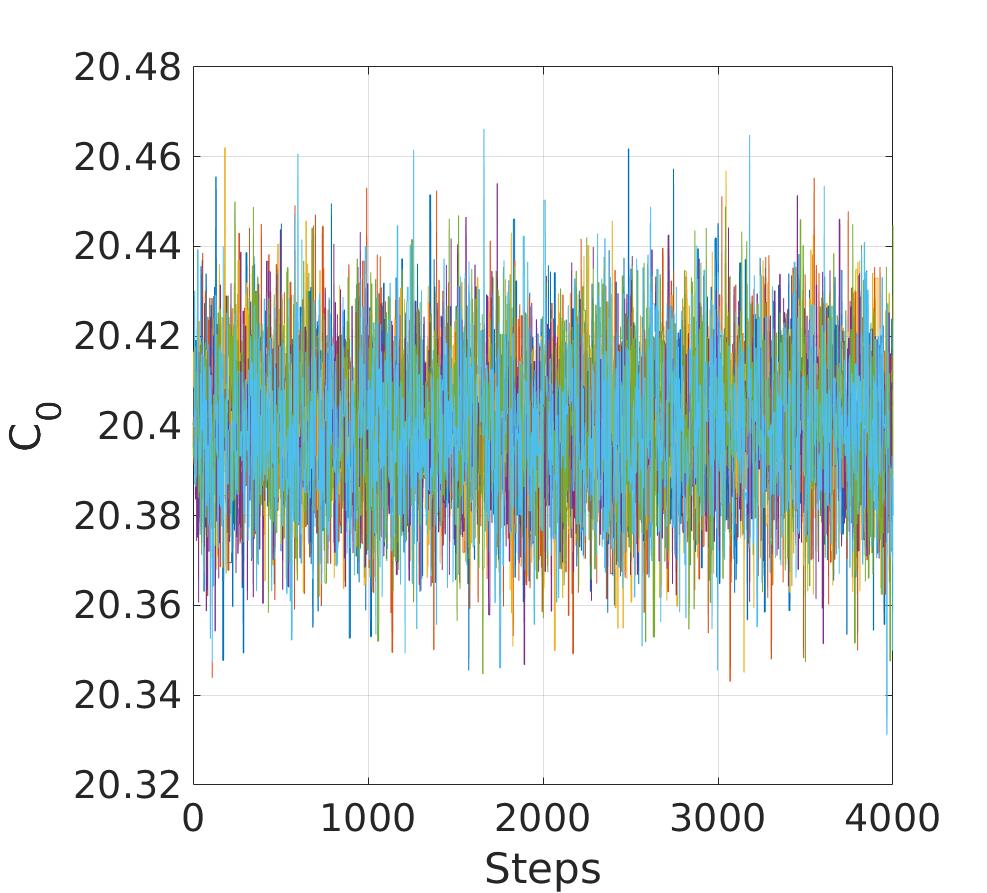}}
\subfigure[Second parameter]{\includegraphics[width=.30\columnwidth]{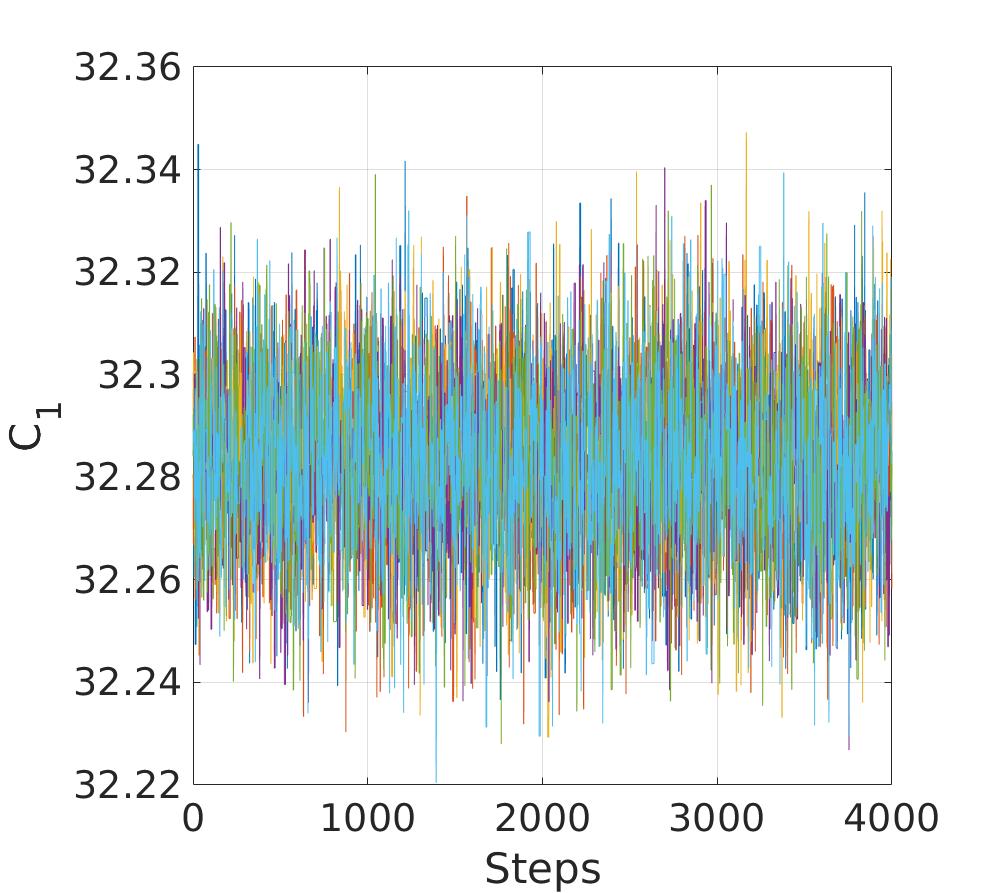}}
\subfigure[$\ln(\sigma_{gp})$]{\includegraphics[width=.30\columnwidth]{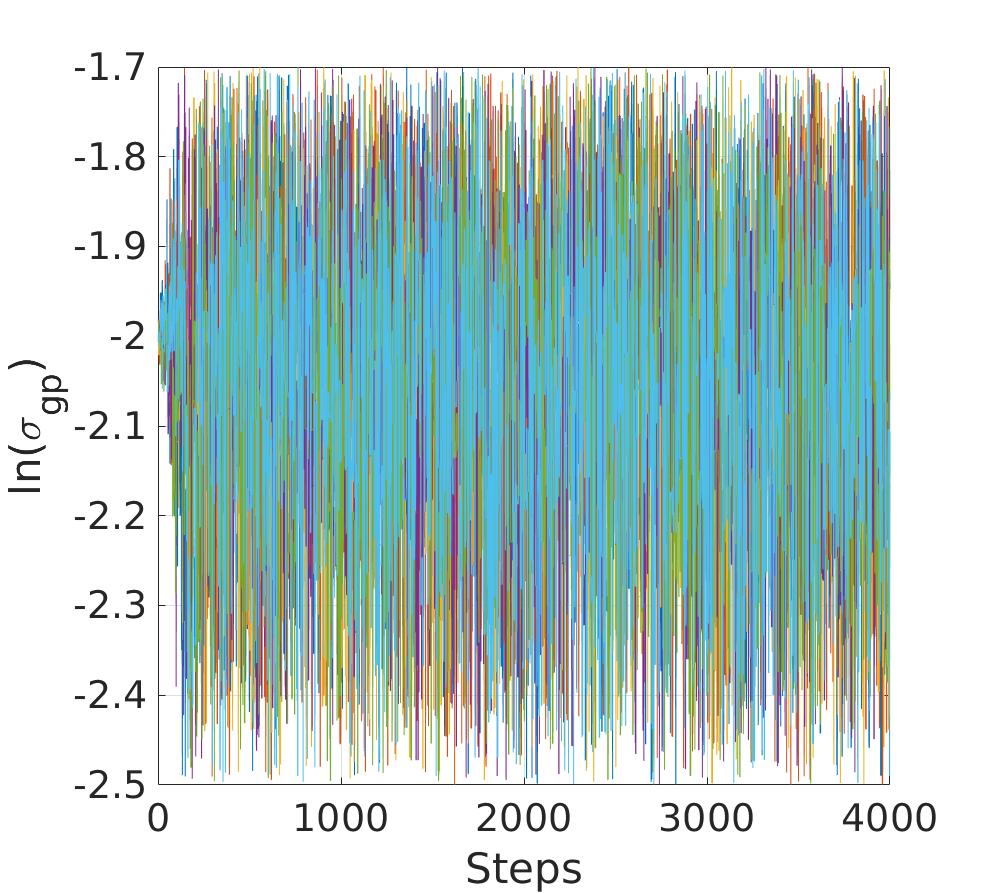}}
\subfigure[Joint distribution of the first and second parameters in $\Cb$]{\includegraphics[width=.48\columnwidth]{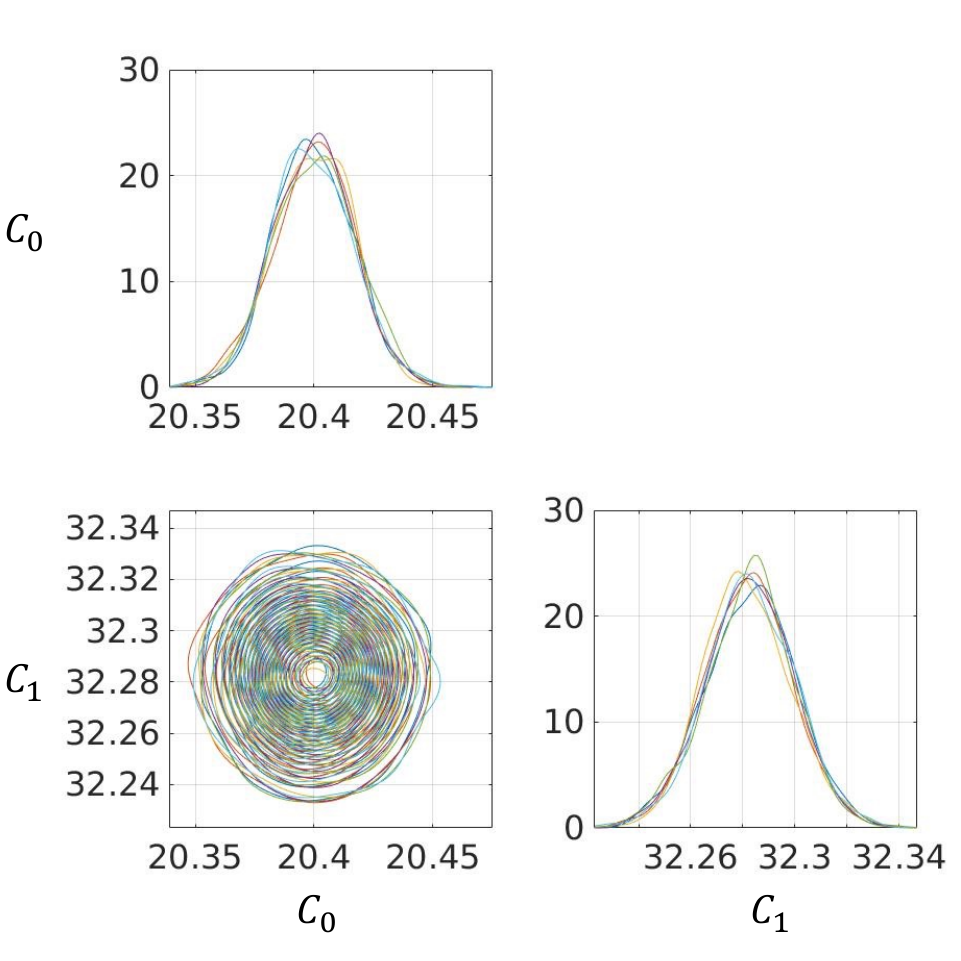}}
\subfigure[Joint distribution of the third parameter in $\Cb$ and $\ln(\sigma_{gp})$]{\includegraphics[width=.48\columnwidth]{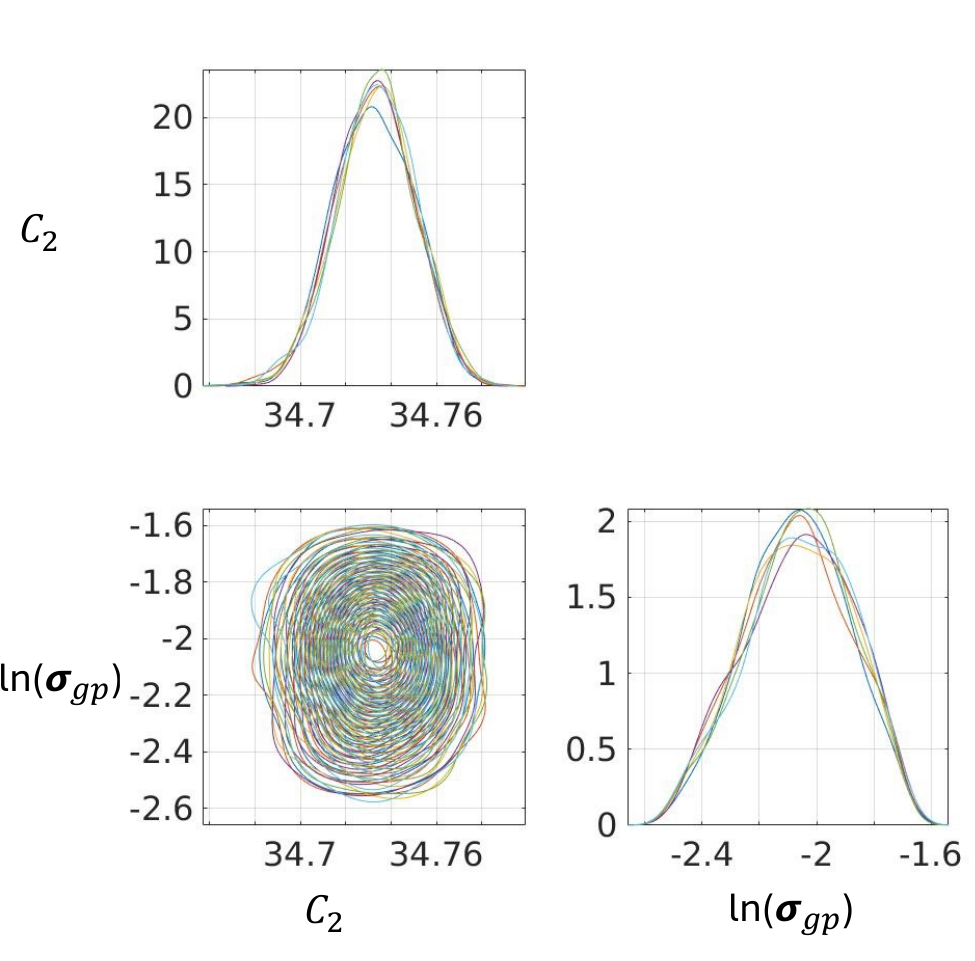}}
\caption{Convergence check: Trace plots and PDFs for the traces. For each trace, we have an acceptance rate $\approx$ 0.42, and an effective sample size $\approx$ 1,000 (out of 4,000 draws).}
\label{fig:conv_trace}
\end{figure}

We first illustrate the utility of the multilevel algorithm using a correlation length $l_{gp}=L/2$.  
We run 6 independent chains, each with 4,000 draws, a burn-in stage of 300 and a subchain length on the coarse level of 100. Each chain is initialized using the scheme proposed in Section \ref{sec:BNOREM}. We examine the quality of the chains both visually and quantitatively. Fig.\ref{fig:conv_trace} illustrates that all the three independent chains attain essentially the same stationary state. Further, using the improved $\hat{R}$ statistic~\cite{vehtari2021rank} relying on ArviZ in Python, we find that the $\hat{R}$ values for all 24 parameters are very close to 1, with the highest being $1.0019$, again highlighting the convergence of the chains. Note that, for an ergodic process, this statistic decays to 1 in the limit of infinite chain length~\cite{vehtari2021rank}.

\begin{figure}[htp]
\centering
\subfigure[First parameter]{\includegraphics[width=.30\columnwidth]{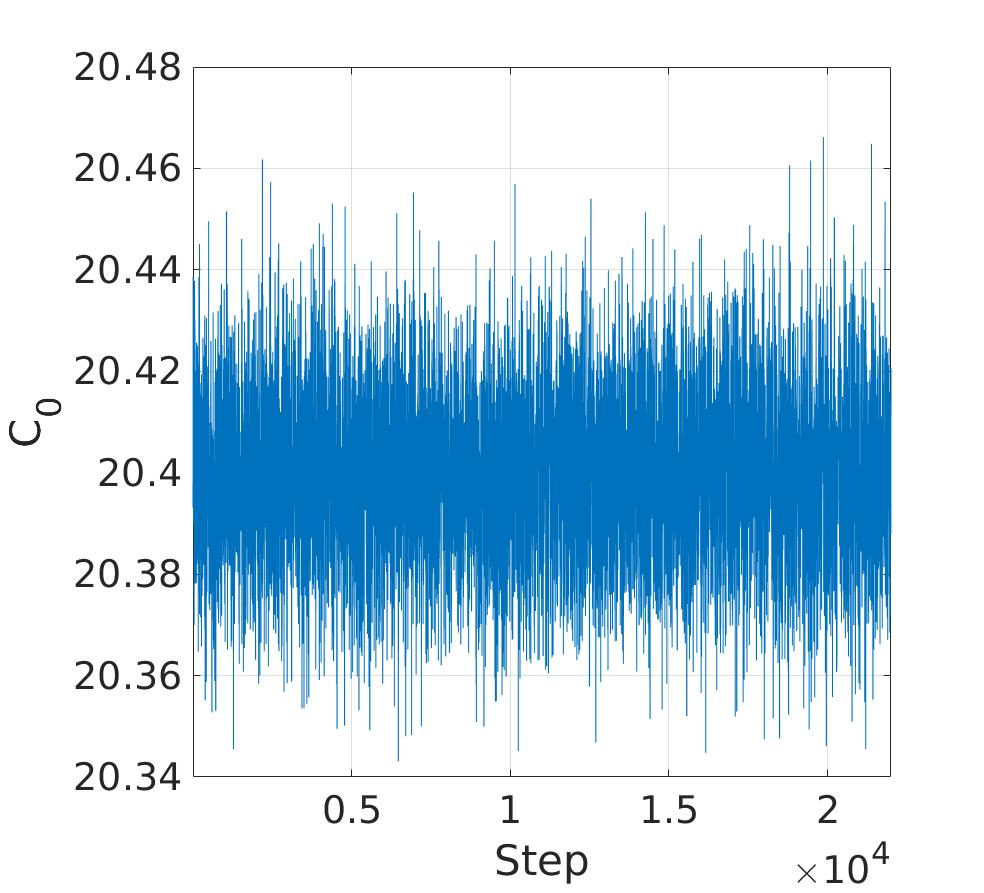}}
\subfigure[Second parameter]{\includegraphics[width=.30\columnwidth]{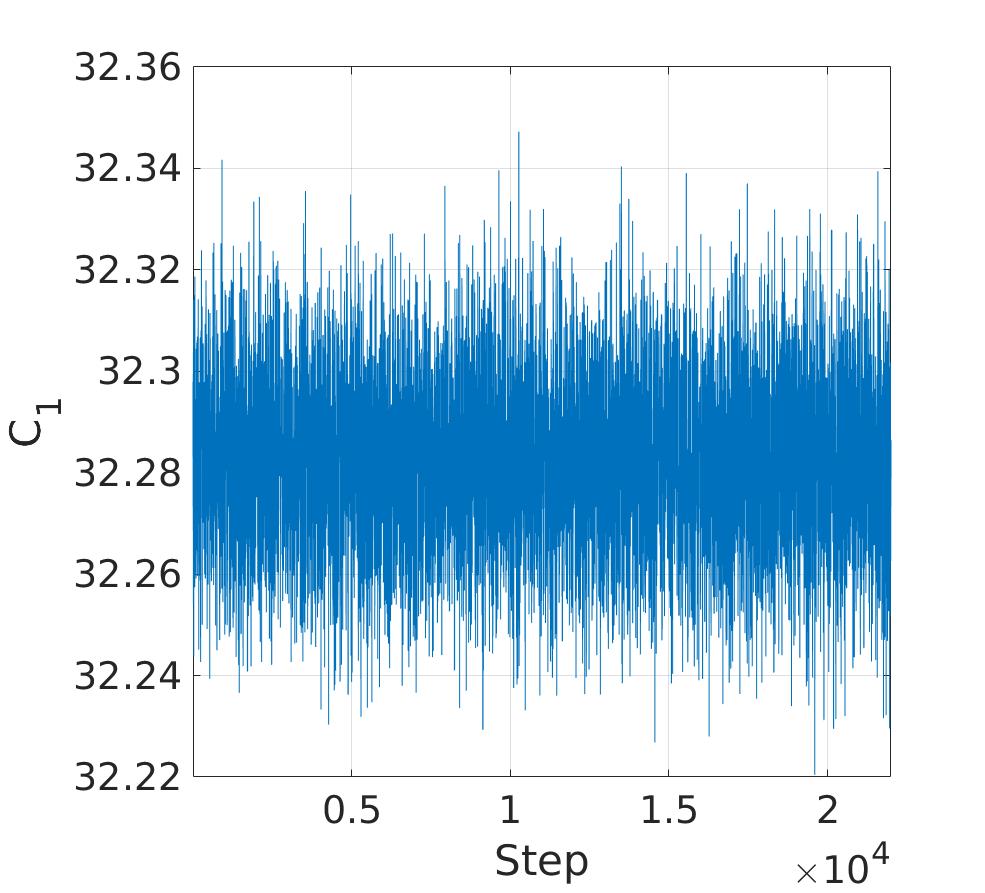}}
\subfigure[$\ln(\sigma_{gp})$]{\includegraphics[width=.30\columnwidth]{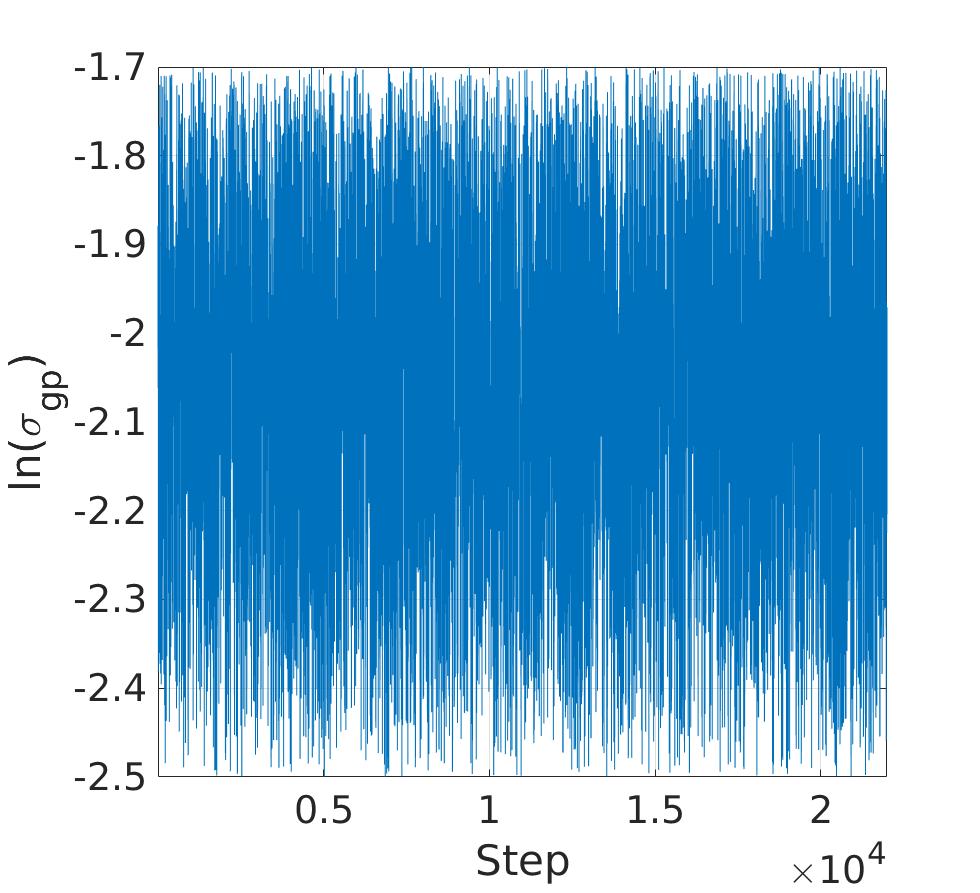}}
\subfigure[Joint distribution of the first and second parameters in $\Cb$]{\includegraphics[width=.48\columnwidth]{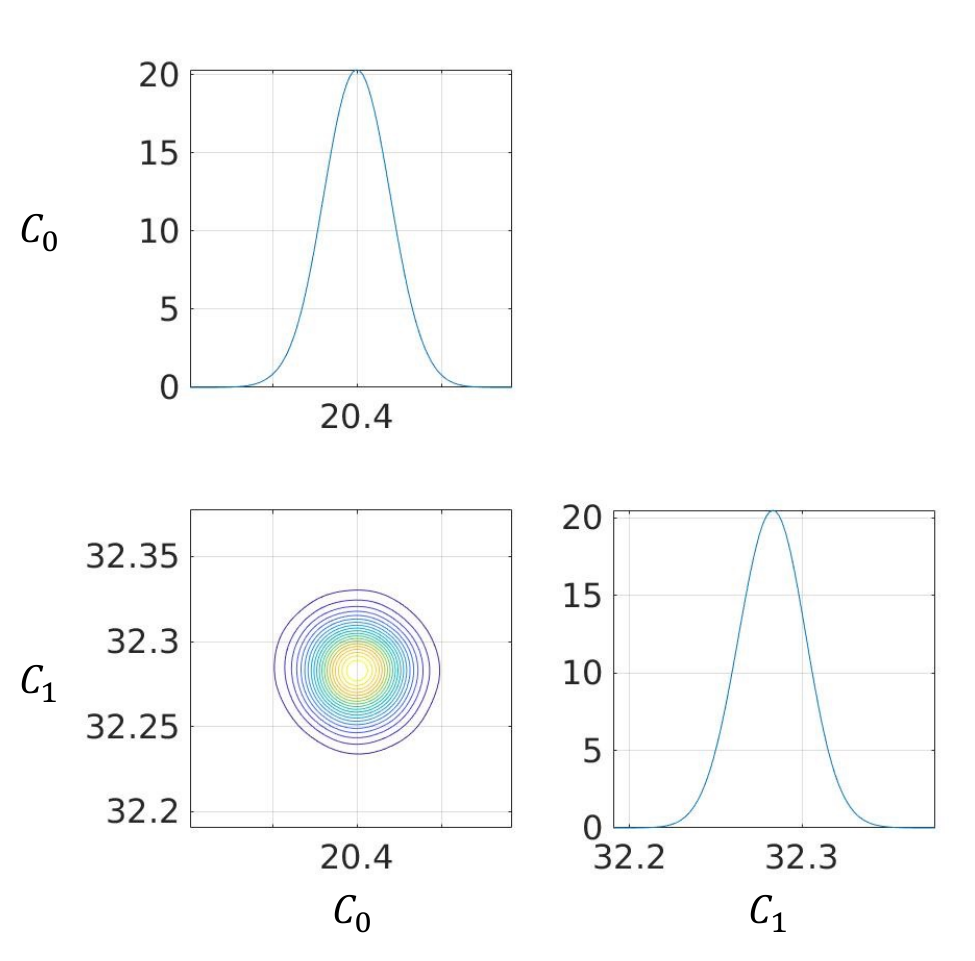}}
\subfigure[Joint distribution of of the third parameter in $\Cb$ and $\ln(\sigma_{gp})$]{\includegraphics[width=.48\columnwidth]{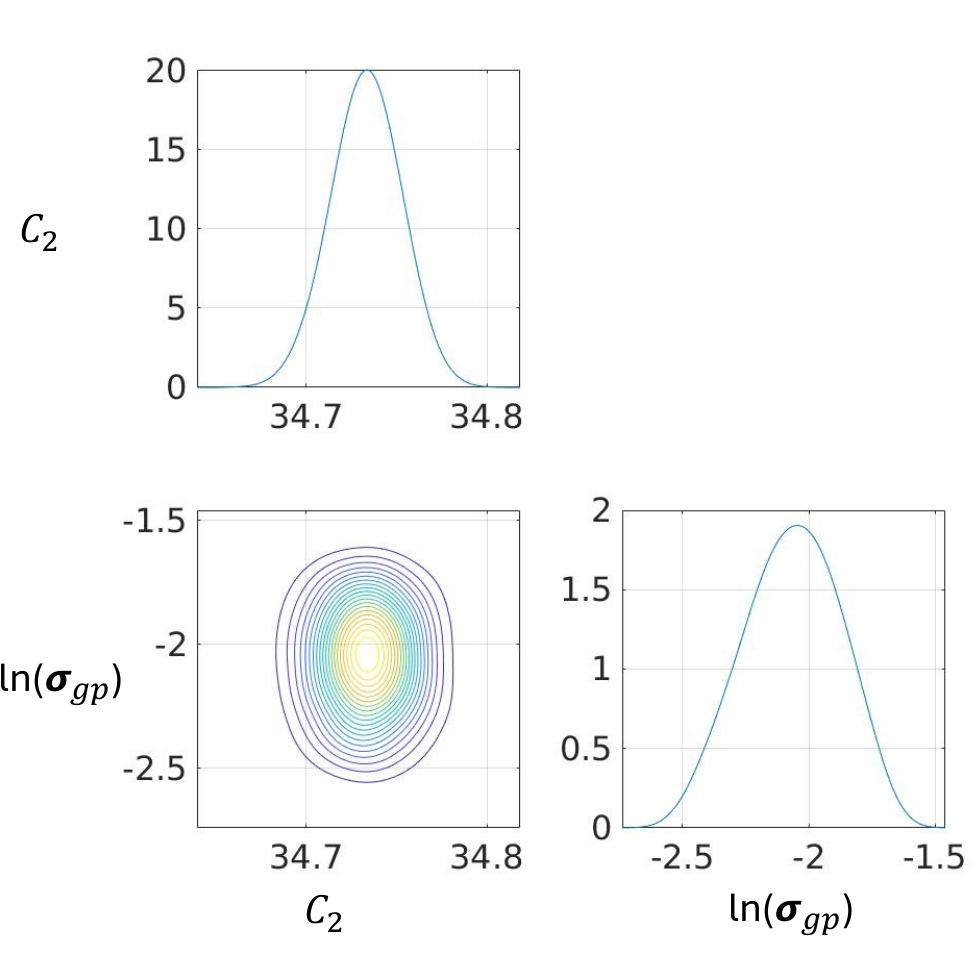}}
\caption{Trace plot and PDF for the combined trace. The acceptance rate $\approx$ 0.42, ESS $\approx$ 6,000 (out of 24,000 draws). }
\label{fig:periodic_trace}
\end{figure}

The chains have 24,000 draws in total, with approximately 42\% acceptance rate on average. To present the aggregate results, we use roughly 6,000 equally spaced samples out of the total, where this effective sample size (ESS) was calculated using the method in \cite{vats2019multivariate}. The trace plots shown in Fig.\ref{fig:periodic_trace} indicate good mixing, and, with the high ESS, we have reliable probability density functions (PDFs) and associated statistics. 

\subsubsection{Impact of GP Correlation Lengths}\label{sec:peri_gp}

As discussed in Section \ref{sec:math}, to avoid the repeated cost of \eqref{eq:lamphitau} and \eqref{eqn:eigen_eq} at each MCMC step, we use the GP correlation length, $l_{gp}$, as a tunable hyperparameter. In this section, we investigate the impact of different values of $l_{gp}$ on the learnt corrected kernels and the corresponding nonlocal solution behaviors.

\begin{figure}[htp]
\centering
\subfigure[Kernel at point $x=0.0$ for $l_{gp}=2L$=40.]{\includegraphics[width=.48\columnwidth]{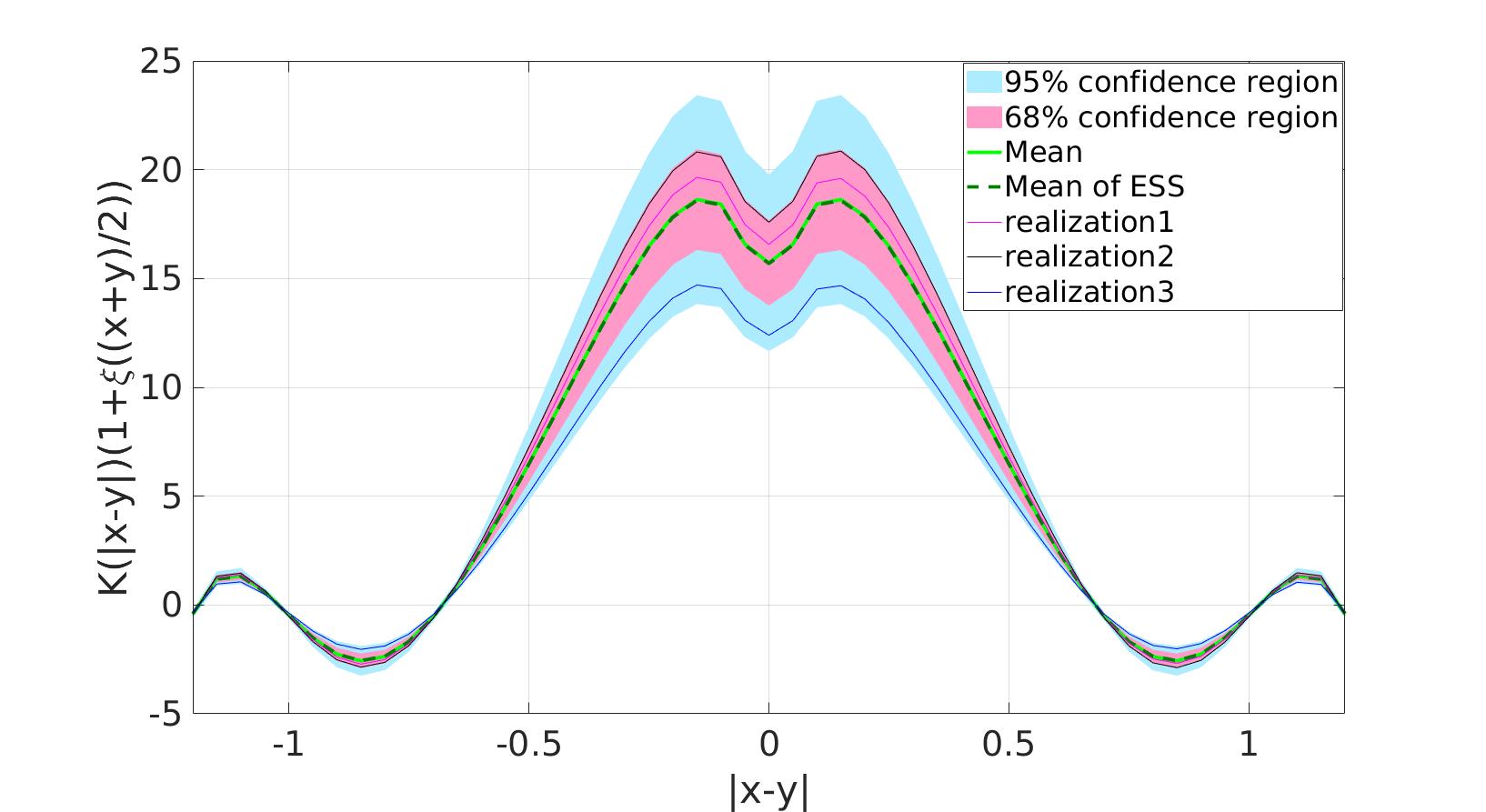}}
\subfigure[Kernel at point $x=0.0$ for $l_{gp}=\dfrac{L}{128}$=0.15625.]{\includegraphics[width=.48\columnwidth]{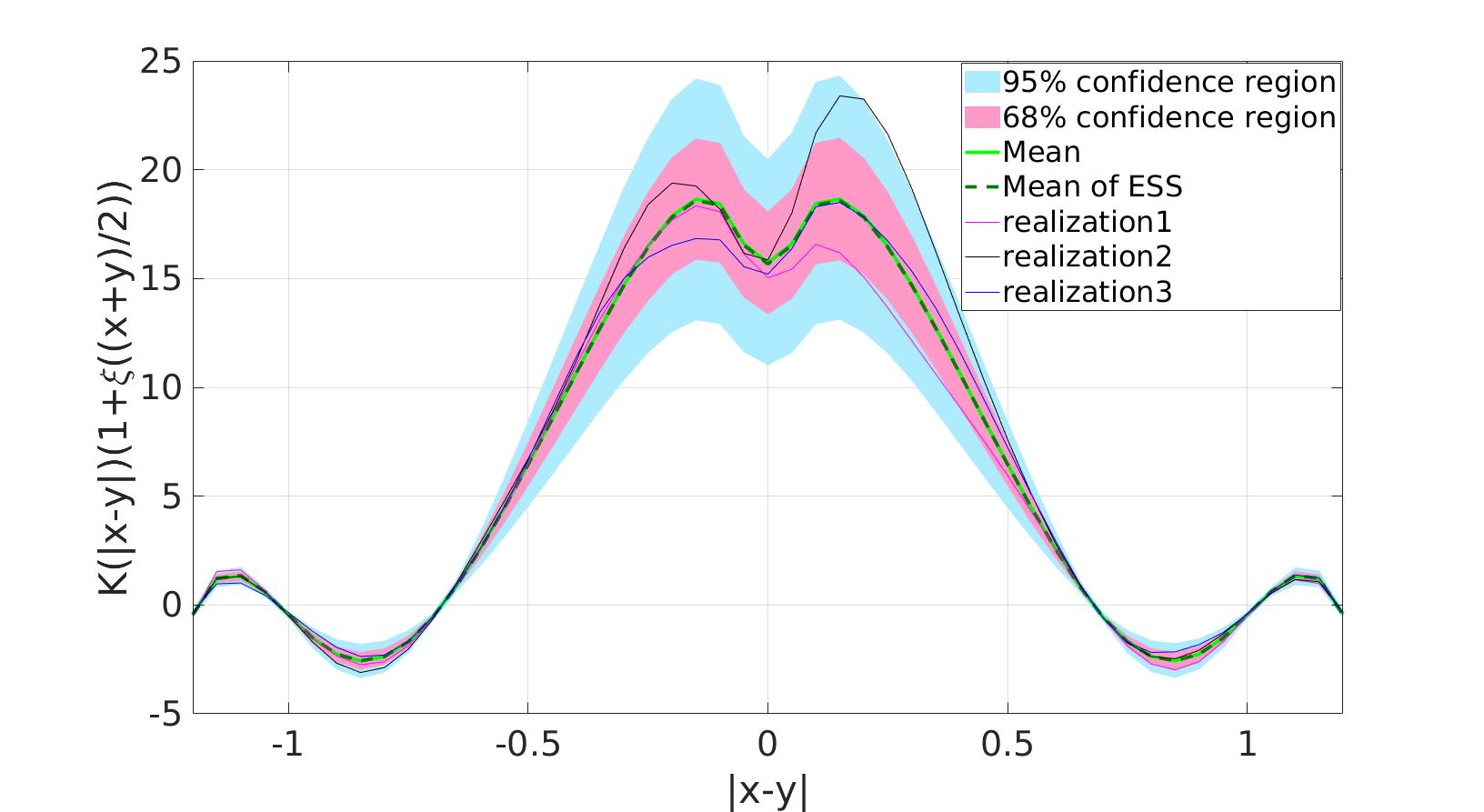}}
\caption{Kernel with uncertainty for a bar with periodic microstructure using two different correlation length.}
\label{fig:periodic_ker}
\end{figure}

In Fig.~\ref{fig:periodic_ker}, the uncertainty with several realizations for the kernel with GP, $K(|x-y|)(1+\xi(\dfrac{x+y}{2}))$, is plotted for correlation lengths $l_{gp}=2L=40$ and $l_{gp}=L/128=0.15625$, using 10,000 realizations in total. Since the behavior of the kernel is similar at each point in the one-dimensional bar, we pick the location $x=0.0$ as an instance for illustration. 
%We point out that the value at $|x-y|=0.0$ is obtained by spline interpolation, to make the plot smooth. 
We observe that the curve labeled `Mean', which is the mean value of all the realizations, is consistent with the corresponding kernel as the mean of the ESS, which is a single push-forward of the embedded model \eqref{eqn:enl} using the mean value of the effective samples and setting the GP to 0. In fact, without the GP the embedded model degenerates to the original nonlocal model \eqref{eqn:homonl}. This consistency is because of the linearity of the nonlocal kernel and the independence of the $\xi$ components.
Further, for different $l_{gp}$ values, one can barely see changes in the mean and the confidence region. On the other hand, sampled GP realizations exhibit significant structural differences between the two cases, in Figs.~\ref{fig:periodic_ker}(a) and \ref{fig:periodic_ker}(b), consistent with the large change in $l_{gp}$ between the two cases.
%However, single realization is of huge difference, since the GP of different correlation length is supposed to capture the uncertainty of different levels. In Fig.\ref{fig:periodic_ker}(a) we plot a realization with $l_{gp}=40$, and find that the kernel is almost symmetric with respect to the origin point, $|x-y|=0$. Additionally, the kernel with correction has the same shape as the original kernel $K(|x-y|)$. This is because the GP correlation length $l_{gp}=40$ in this example is much larger than the scale of the horizon $\delta=1.2$, which makes each generated sample kernel correction term almost flat inside the horizon. On the contrary, in Fig.\ref{fig:periodic_ker}(b) we demonstrate the results on $l_{gp}=0.12625$. One can see that the corrected kernel samples are no longer symmetric, and the shape of each realization varies drastically, due to the dramatic oscillation of a GP with a small correlation length $l_{gp}=0.15625$ inside the horizon. 

\begin{figure}[htp]
\centering
\subfigure[Group velocity using $l_{gp}=2L$=40.]{\includegraphics[width=.65\columnwidth]{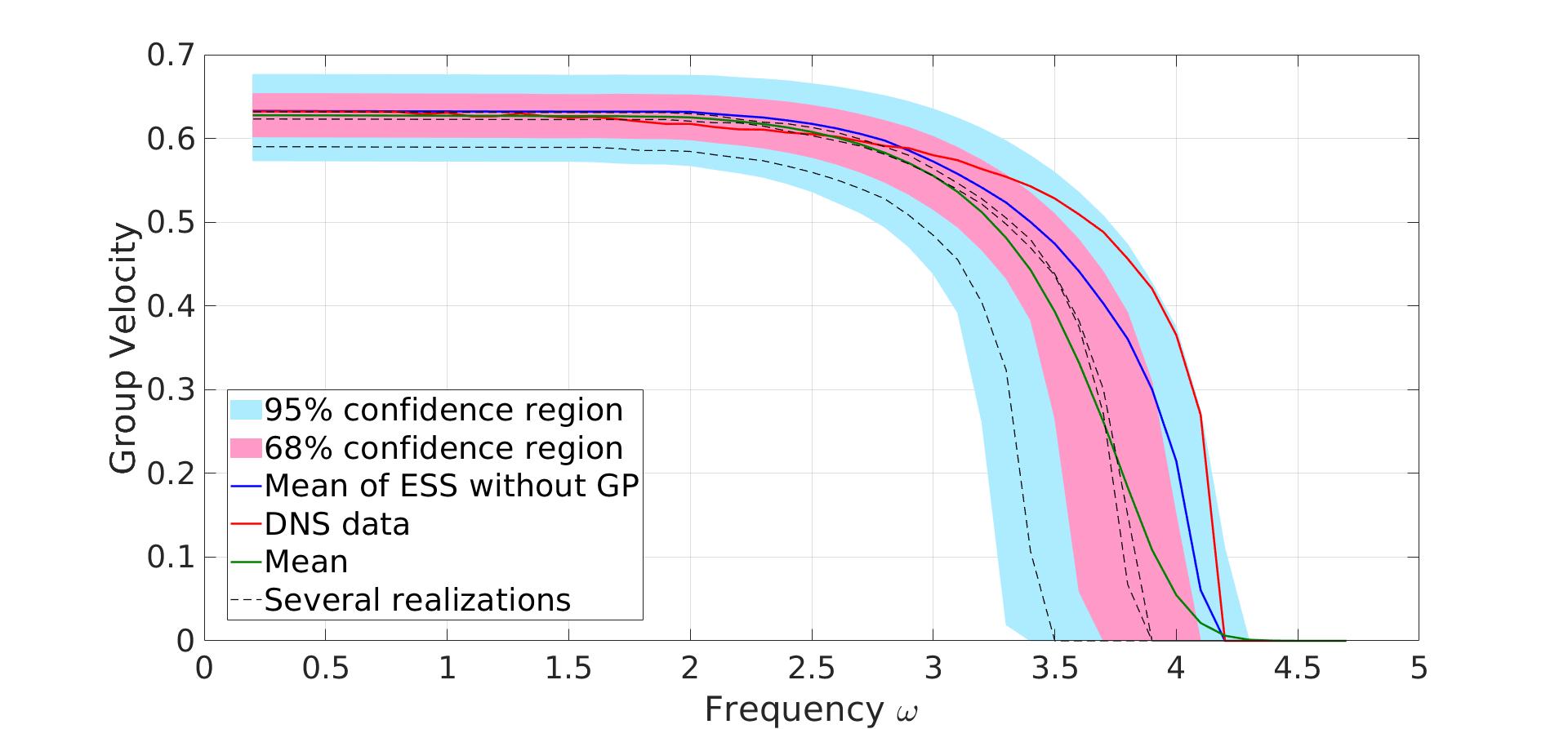}}
\subfigure[Dispersion curve using $l_{gp}=2L$=40.]{\includegraphics[width=.30\columnwidth]{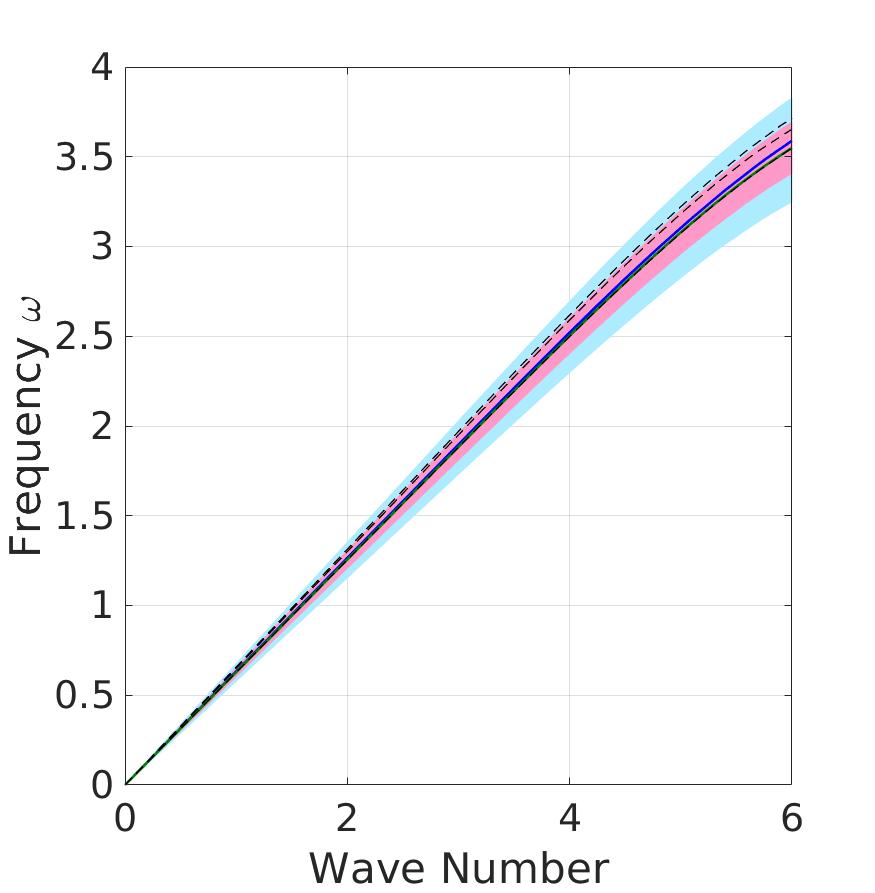}}
%\subfigure[Group velocity using $l_{gp}=\dfrac{L}{32}$=0.625.]{\includegraphics[width=.65\columnwidth]{Figures/vg_lam32.jpg}}
%\subfigure[Dispersion curve using $l_{gp}=\dfrac{L}{32}$=0.625.]{\includegraphics[width=.30\columnwidth]{Figures/dispersion_lam32_jun25.jpg}}
\subfigure[Group velocity using $l_{gp}=\dfrac{L}{32}$=0.625.]{\includegraphics[width=.65\columnwidth]{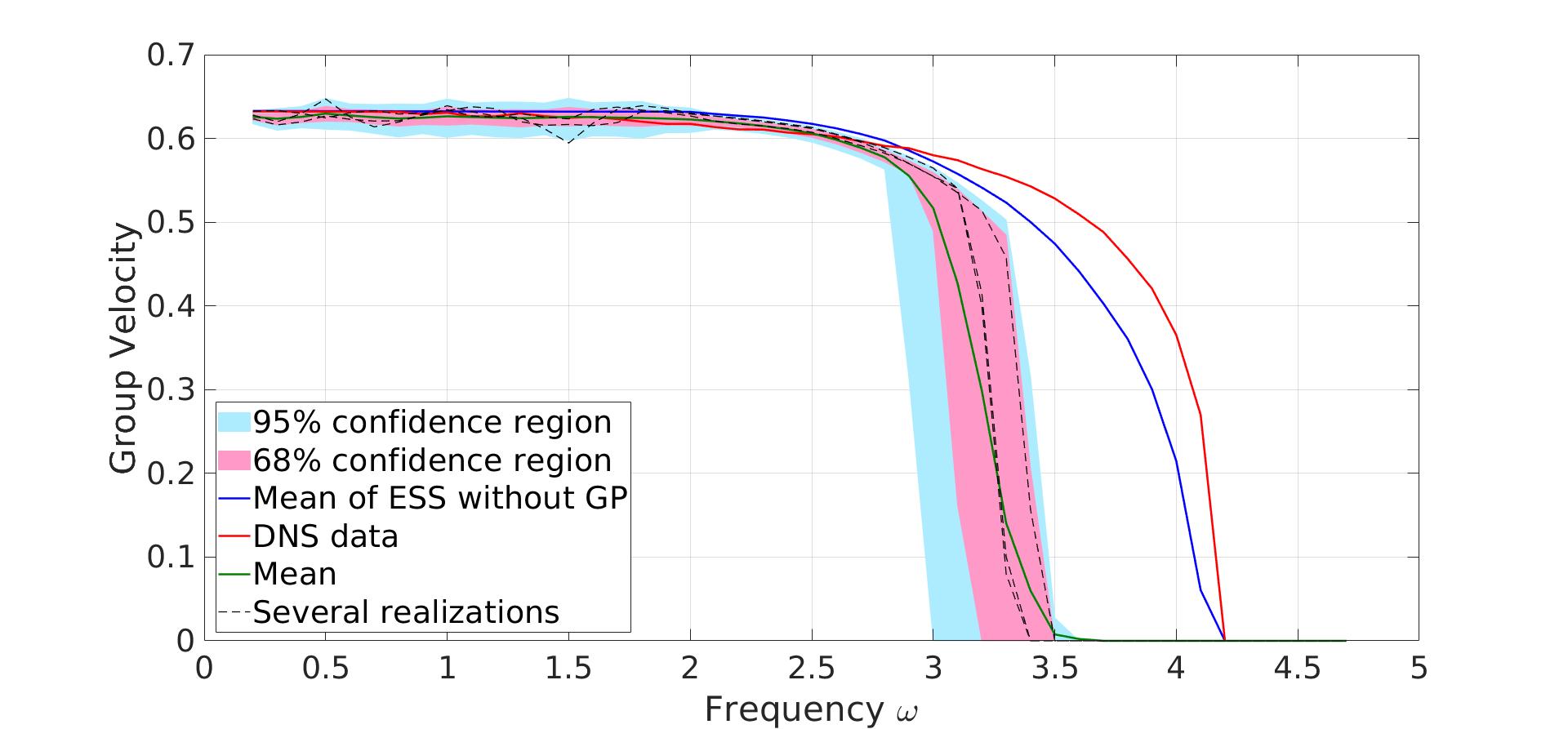}}
\subfigure[Dispersion curve using $l_{gp}=\dfrac{L}{32}$=0.625.]{\includegraphics[width=.30\columnwidth]{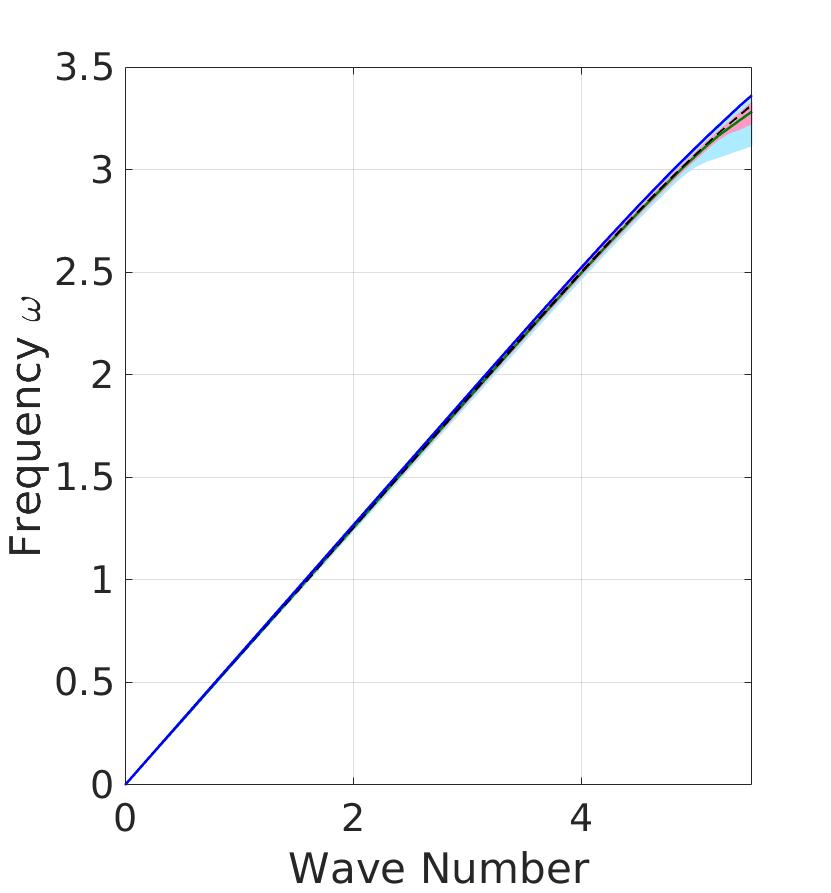}}
\caption{Group velocity and dispersion curve with uncertainty for a bar with periodic microstructure using two different correlation lengths. The curve labeled `Mean of ESS without GP' denotes the single push-forward with the GP set to $0$ and using the mean value of the effective samples. The curve labeled `Mean' denotes the mean value of all the realizations.}
\label{fig:periodic_vg}
\end{figure}

To examine the dispersion behavior of these kernel realizations, we plot in Figs.~\ref{fig:periodic_vg}(a,c) the group velocity of kernels corresponding to a GP with a large ($l_{gp}=40$) and relatively small ($l_{gp}=0.625$) correlation length. For each realization of the kernel, the group velocity is computed using a wave packet that travels a long enough distance such that the wave is away from the ends of the bar. In order to do this in numerical simulation, we set the bar length equal to 400 and generate realizations of the kernels with the same correlation length on this longer bar. We note that the reduction in $l_{gp}$ results in a reduced band stop location (the smallest frequency where the group velocity drops to zero). Further, it is evident that the confidence region matches the dispersion behavior better when using a large correlation length. At the same time, oscillation is observed in the low frequency ($\omega<2.0$) region in the small $l_{gp}$ case. Finally, in Figs.~\ref{fig:periodic_vg}(b,d), we plot the dispersion curves of these kernels. Note that all learnt kernels are positive, highlighting the physical stability of the corresponding nonlocal models.

\begin{figure}[htp]
\centering
\subfigure[Sample 10 from type 1 data at t=2.0, $l_{gp}=40$.]{\includegraphics[width=.48\columnwidth]{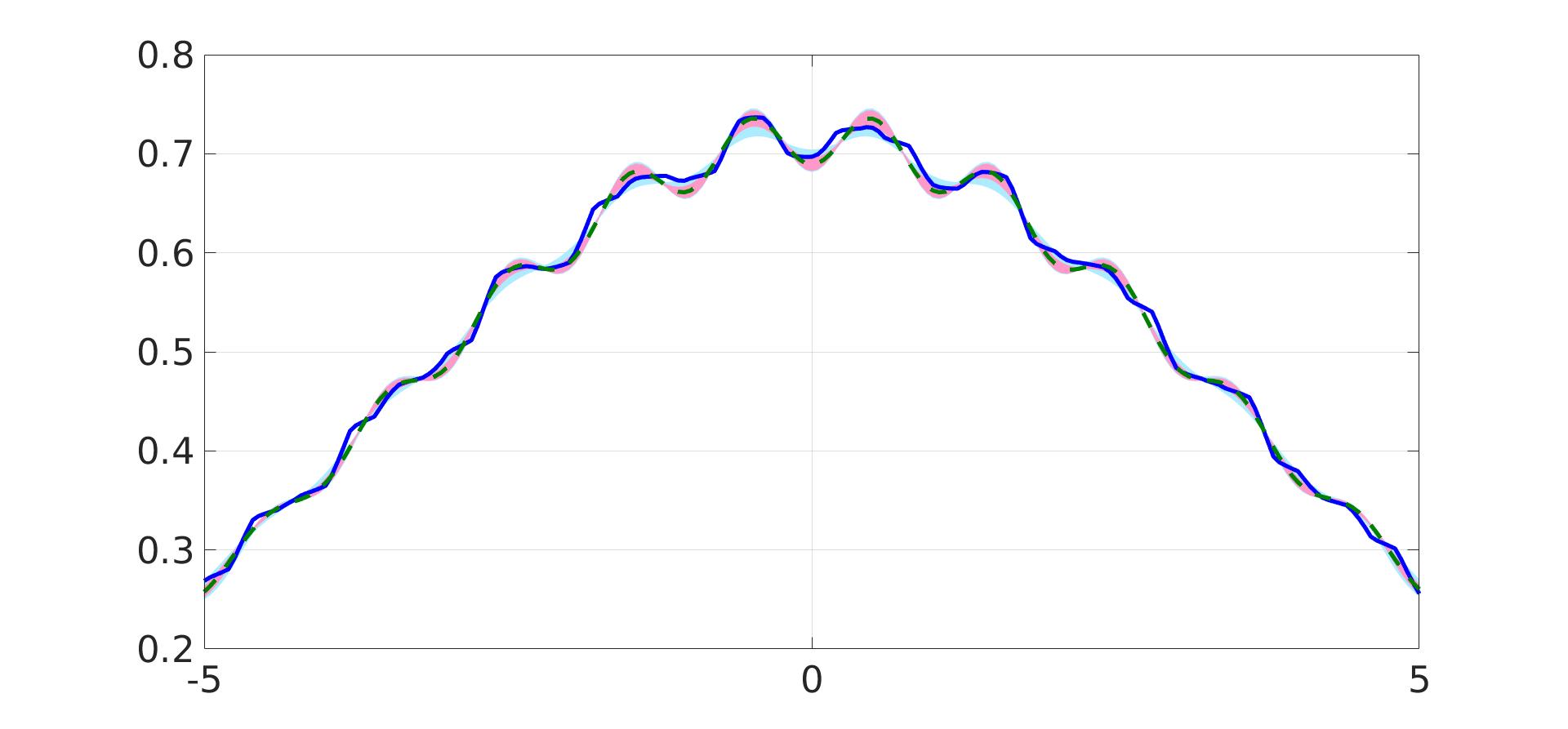}}
\subfigure[Sample 3 from type 2 data at t=2.0, $l_{gp}=40$.]{\includegraphics[width=.48\columnwidth]{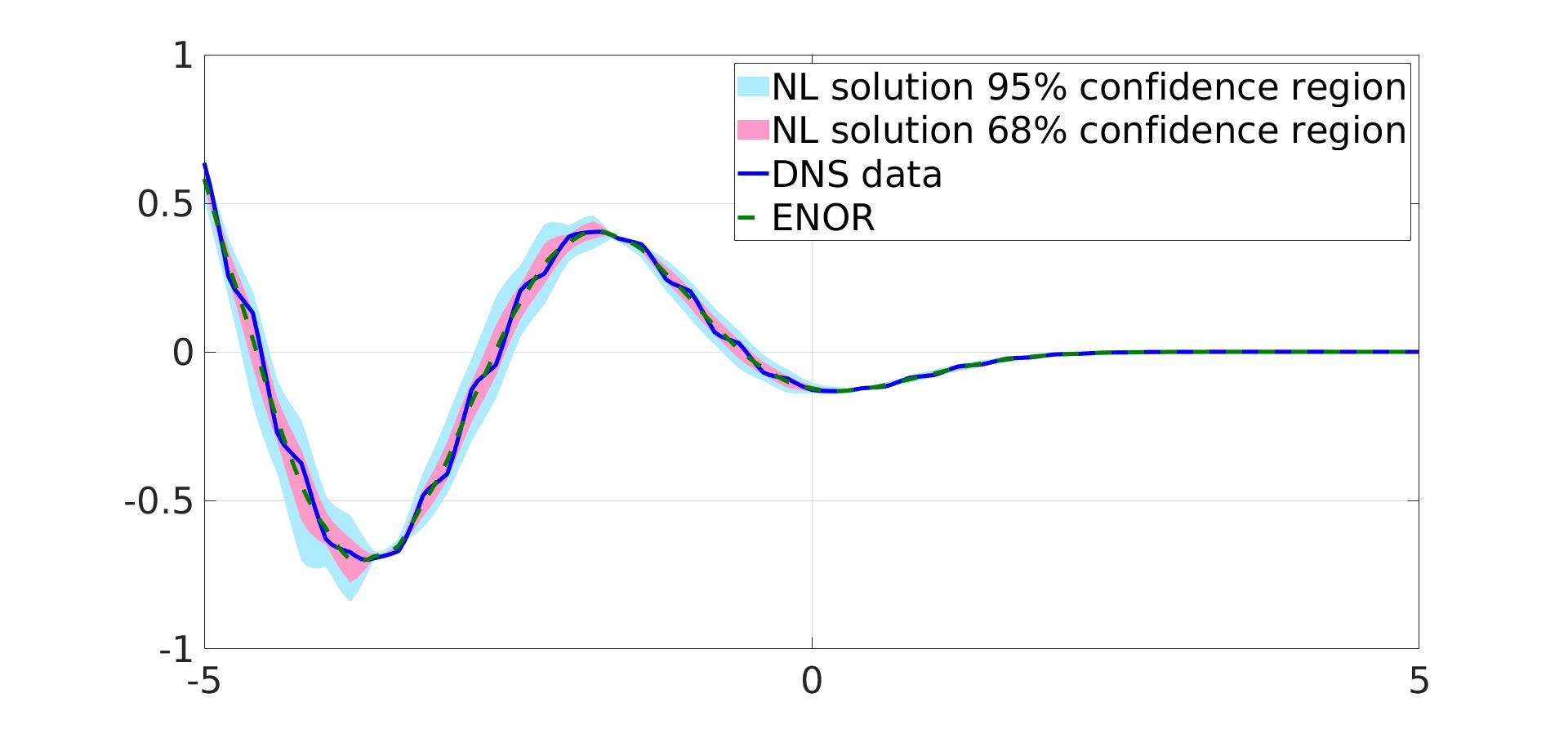}}
\subfigure[Sample 10 from type 1 data at t=2.0, $l_{gp}=0.15625$.]{\includegraphics[width=.48\columnwidth]{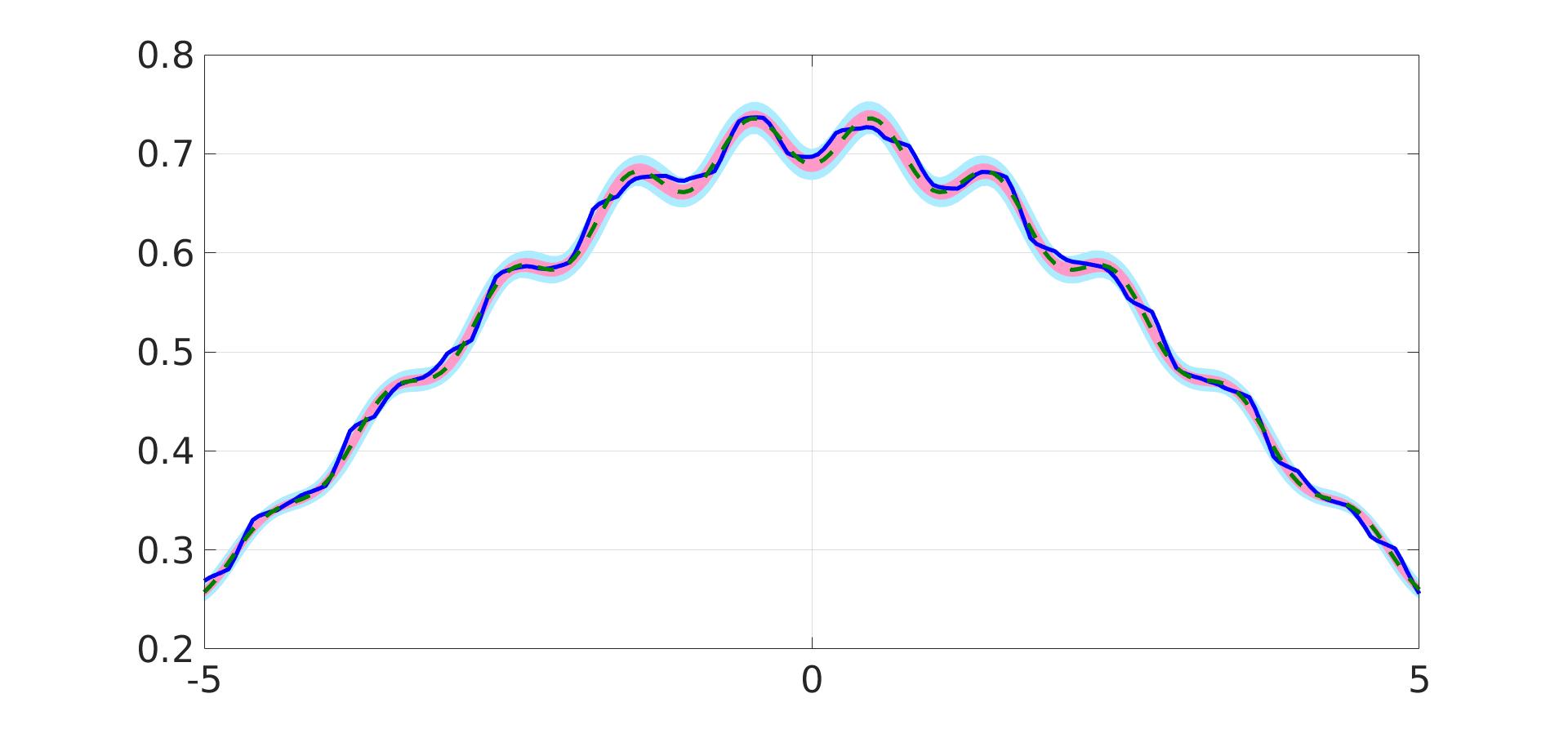}}
\subfigure[Sample 3 from type 2 data at t=2.0, $l_{gp}=0.15625$.]{\includegraphics[width=.48\columnwidth]{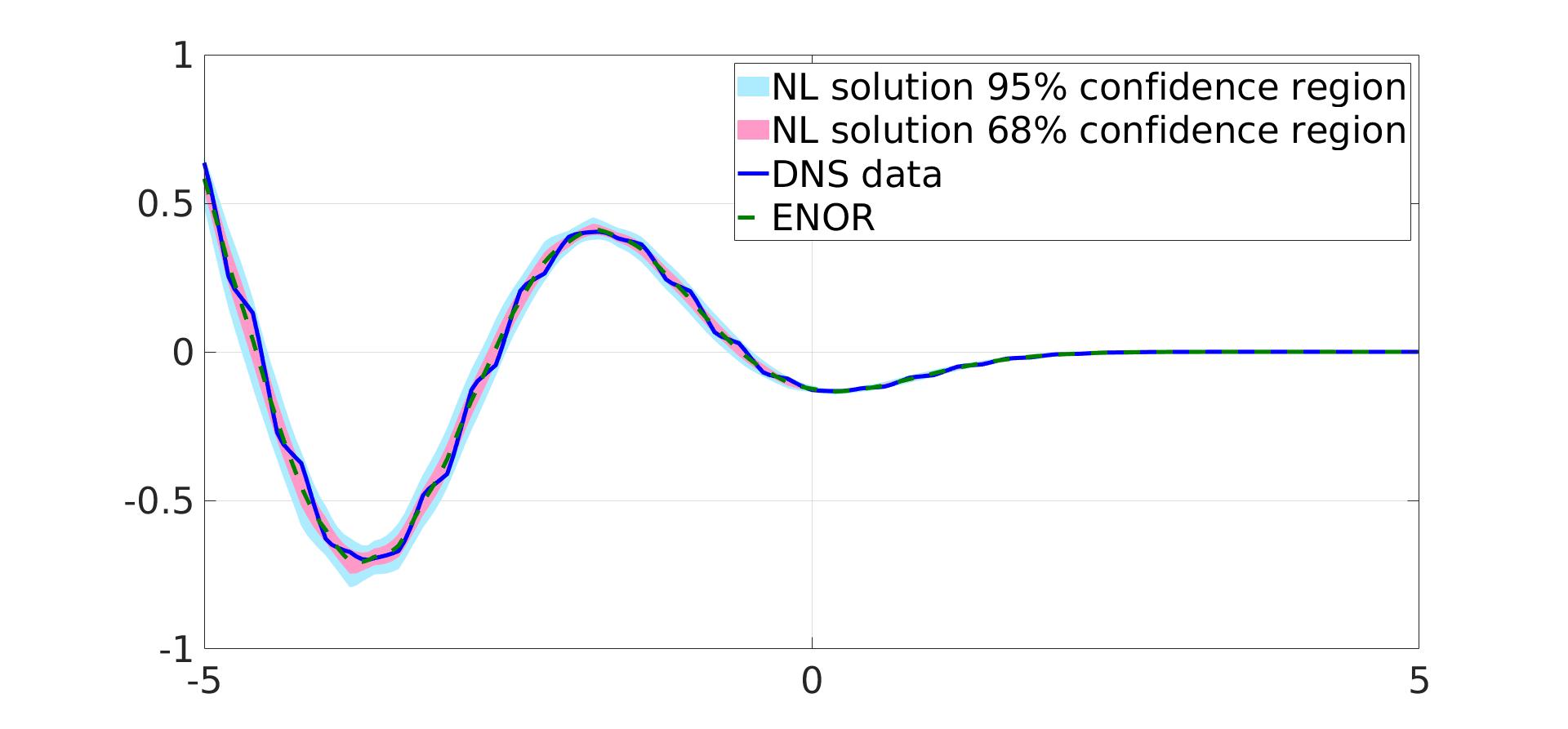}}
\caption{Posterior uncertainty on training samples for a bar with periodic microsturcture using two different correlation length.}
\label{fig:peri_tpf}
\end{figure}

\begin{figure}[htp]
\centering
\subfigure[Best $l_{gp}$ for training samples]{\includegraphics[width=.75\columnwidth]{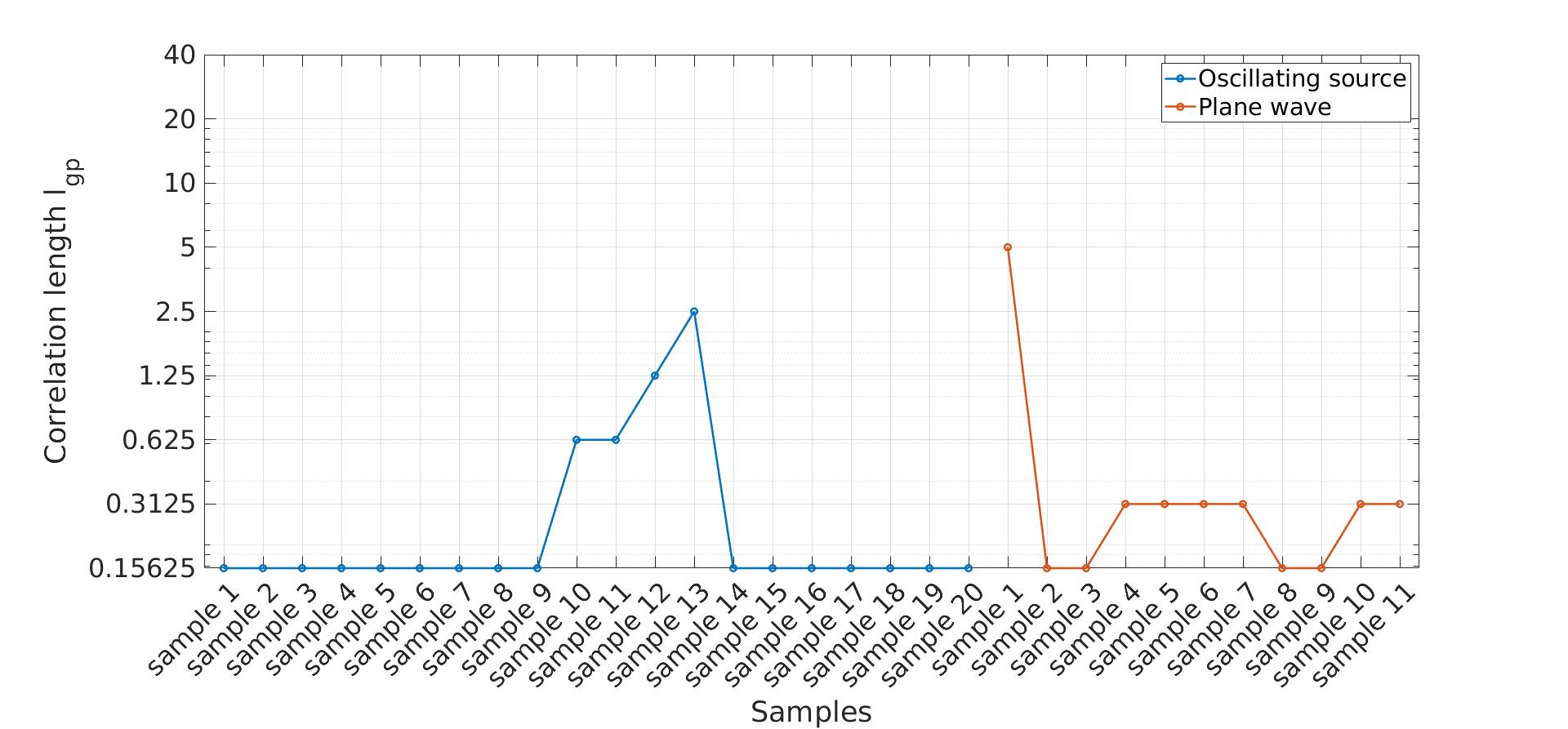}}
\subfigure[Best $l_{gp}$ for validation samples]{\includegraphics[width=.23\columnwidth]{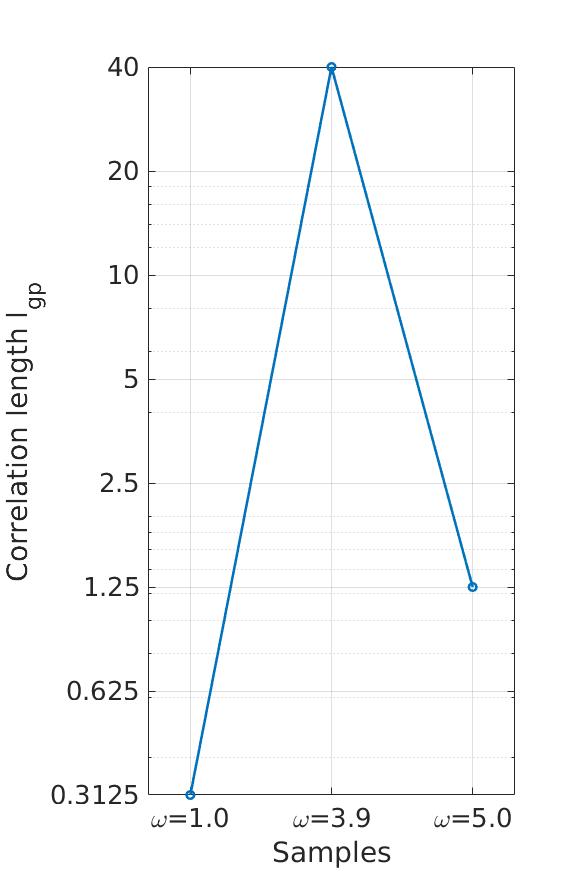}}
\caption{Best $l_{gp}$ (with the lowest CRPS) on training and validation samples for a bar with periodic material. }
\label{fig:periodic_crps}
\end{figure}

Next, we investigate the posterior uncertainty on the predicted displacements from the learnt models. Two samples from setting 1 and 2 datasets are considered, with forcing at $k=10$ in setting 1 and $\omega=1.05$ in setting 2. 
Let us first recall the definitions of two relevant push-forward posteriors for a typical additive data noise problem setup $y=f_d(x,\lambda,\sigma)=f(x,\lambda)+\epsilon(\sigma)$ given the parameter posterior $p(\lambda,\sigma|D)$. We have the push-forward posterior (PFP) as the push-forward of the parameter posterior through the predictive model $f(x,\lambda)$, while the posterior predictive (PP) is defined as the push-forward of the parameter posterior through the full data model $f_d(x,\lambda,\sigma)$~\cite{Gelman:1995,Gelman:1996}.
Note that, in an embedded model-error construction with no additive data noise, the PFP and PP are equivalent. 
In Fig.\ref{fig:peri_tpf}, the PFP, which is the push-forward of the posterior $p(\Cb,\sigma_{gp}|\mathcal{D})$ through the nonlocal model with embedded model error, is plotted for two different $l_{gp}$ values. 
For each case, we see that both the 95\% and 68\% PFP confidence regions generally cover the majority of the ground truth data over the spatial domain. However, we do observe that, visually, and for these two samples, the kernel with a smaller $l_{gp}$ does a better job of spanning the ground truth data discrepancy from the mean model. This is also illustrated in Fig.\ref{fig:periodic_crps}, where the best choices of $l_{gp}$ for different training samples are plotted according to the Continuous Ranked Probability Score (CRPS) \cite{zamo2018estimation}. The CRPS compares a single ground truth value to a distribution. Assume that we have a ground truth $\bar{y}$ and a cumulative distribution function (CDF) $H$ for a variable $\bar{x}$, then the CRPS can be analytically written as $$\text{CRPS}(H,\bar{y})=\int (H(\bar{x})-\mathbbm{1}_{\{\bar{x}\geq\bar{y}\}})^2\text{d}\bar{x}. $$ In a numerical setting where only a sampling-based empirical CDF is available, \cite{zamo2018estimation} provides alternative forms of the CRPS which are feasible to estimate 
\begin{align}
\text{CRPS}(H,\bar{y})&=\mathbb{E}[|\bar{X}-\bar{y}|]-\dfrac{1} {2}\mathbb{E}[|\bar{X}-\bar{X}'|]\label{eqn:nrg}\\
&=\mathbb{E}[|\bar{X}-\bar{y}|]+\mathbb{E}[\bar{X}]-2\mathbb{E}[\bar{X}\cdot H(\bar{X})]\label{eqn:pwm},
\end{align} 
where $\bar{X}$, $\bar{X}'$ are independently and identically distributed according to $H$. %Equations \eqref{eqn:nrg} and \eqref{eqn:pwm} are named NRG and PWM formula, respectively.
Per the definition of the CRPS, the lower the score is, the better does our predicted displacement match the DNS data in distribution. Specifically, we use equation \eqref{eqn:nrg} here, but in principle the two expressions are equivalent. The average CRPS values across different training samples, at $t=2.0$, are summarized in Table.\ref{table:crps_tpf}. As the correlation length $l_{gp}$ decreases, only a slight reduction in the CRPS is obtained, suggesting that all studied kernels have comparable performance in obtaining the correct distribution of the displacement.

\begin{table}[htp]
\begin{tabular}{|l|l|l|l|l|l|l|l|l|l|}
\hline
Material & $2L$ & $L$ & $L/2$ & $L/4$ & $L/8$ & $L/16$ & $L/32$ & $L/64$ & $L/128$ \\ \hline
Periodic & 0.0048 & 0.0048 & 0.0048 & 0.0048 & 0.0047 & 0.0046 & 0.0045 & 0.0044 & 0.0043 \\ \hline
Disorder & 0.0080 & 0.0080 & 0.0080 & 0.0079 & 0.0079 & 0.0077 & 0.0076 & 0.0076 & 0.0076 \\ \hline
\end{tabular}
\caption{Average CRPS of training samples for different $l_{gp}$. The value of the CRPS is evaluated for all the training samples on each grid in the physical domain at the last time step $t=2.0$.}
\label{table:crps_tpf}
\end{table}

\begin{figure}[htp]
\centering
\subfigure[$\omega=2$, t=100.0, $l_{gp}=40$.]{\includegraphics[width=.48\columnwidth]{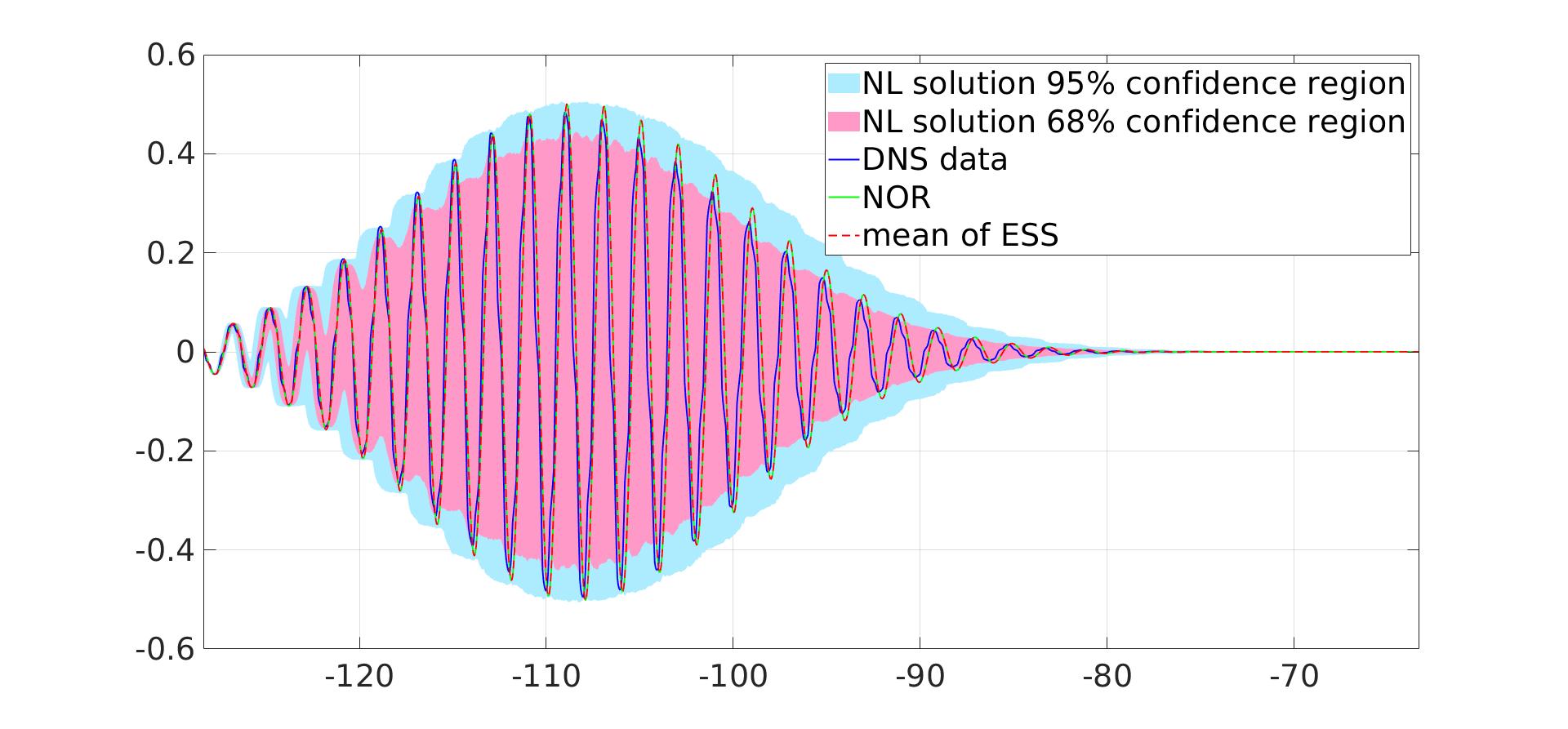}}
\subfigure[$\omega=2$, t=100.0, $l_{gp}=0.15625$.]{\includegraphics[width=.48\columnwidth]{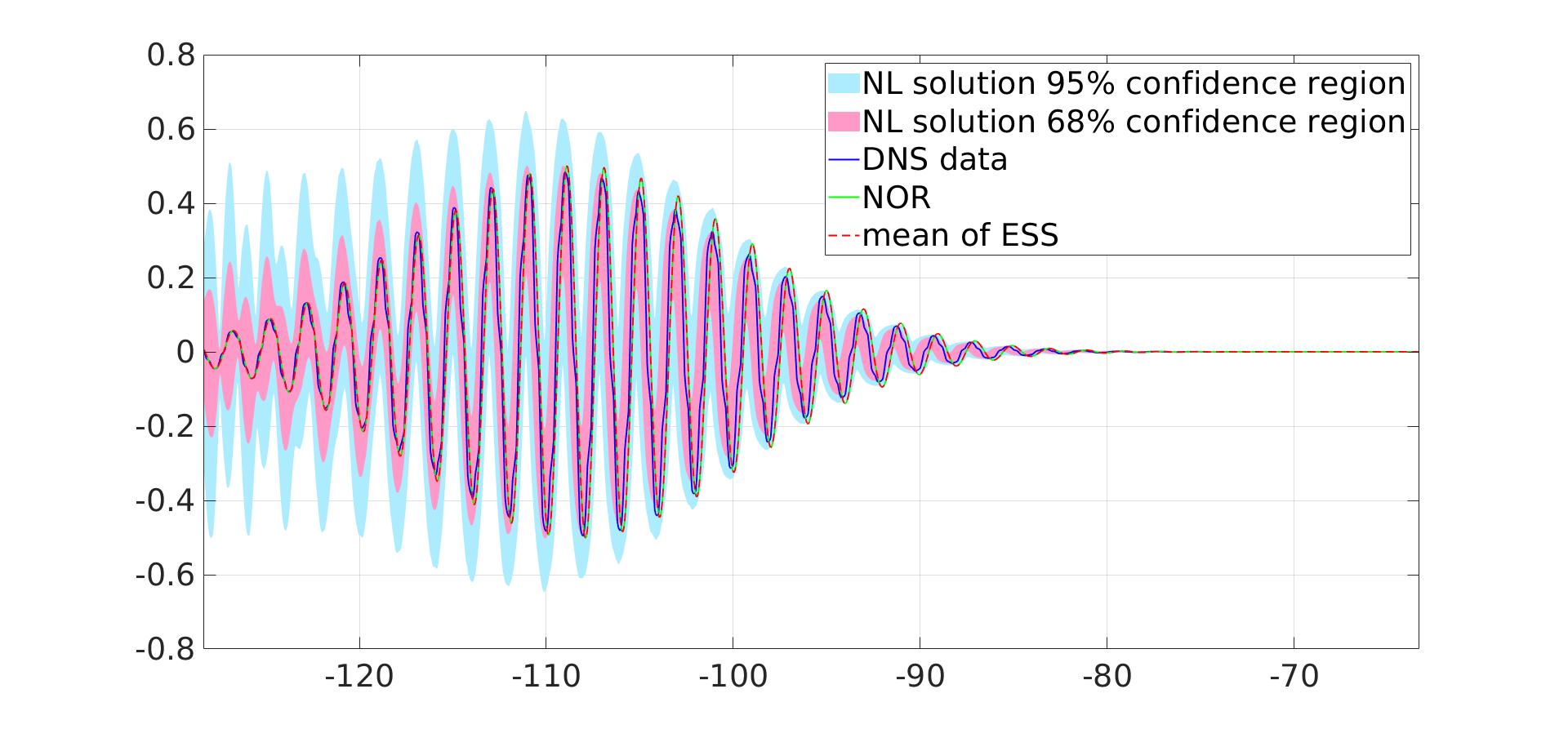}}
\subfigure[$\omega=3.9$, t=100.0, $l_{gp}=40$.]{\includegraphics[width=.48\columnwidth]{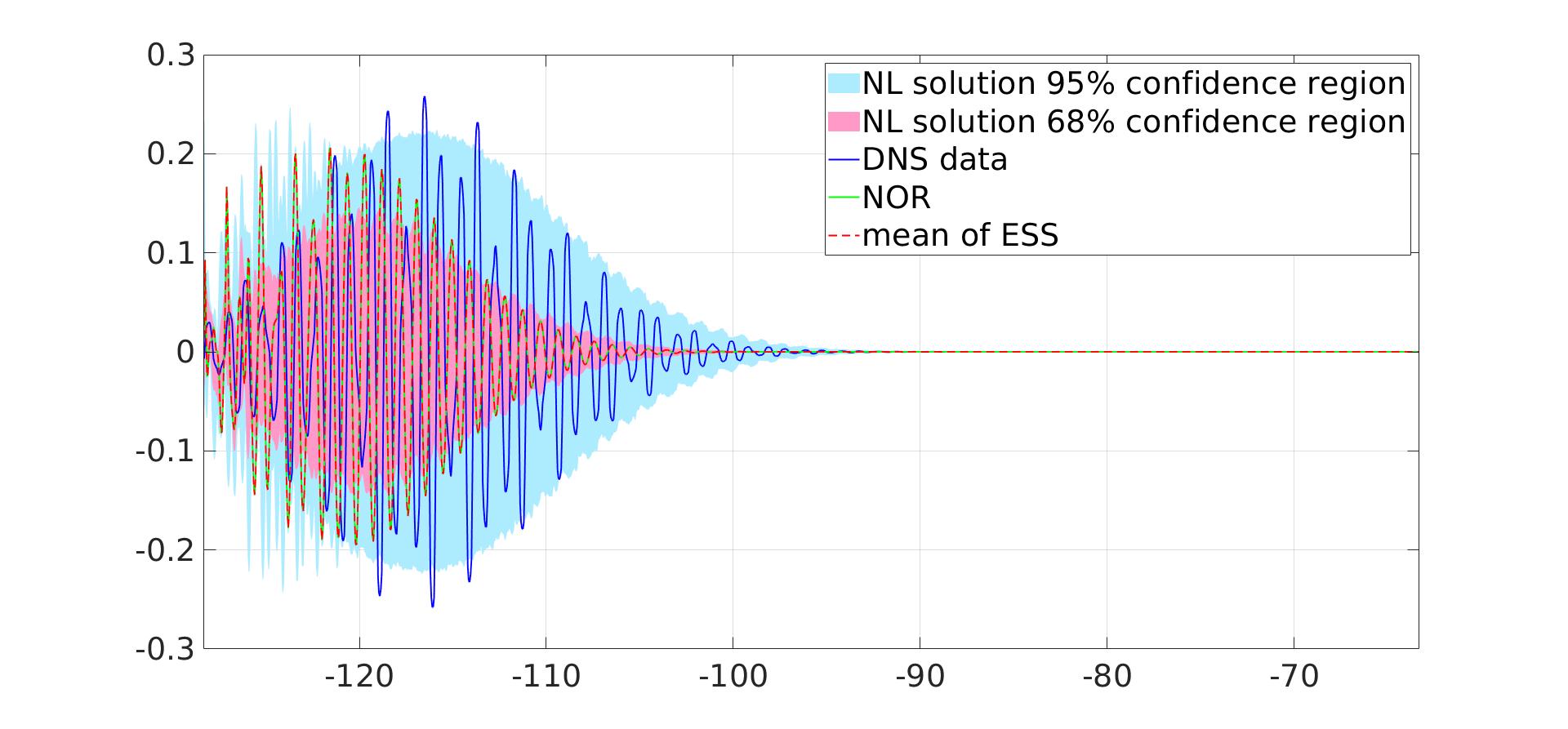}}
\subfigure[$\omega=3.9$, t=100.0, $l_{gp}=0.15625$.]{\includegraphics[width=.48\columnwidth]{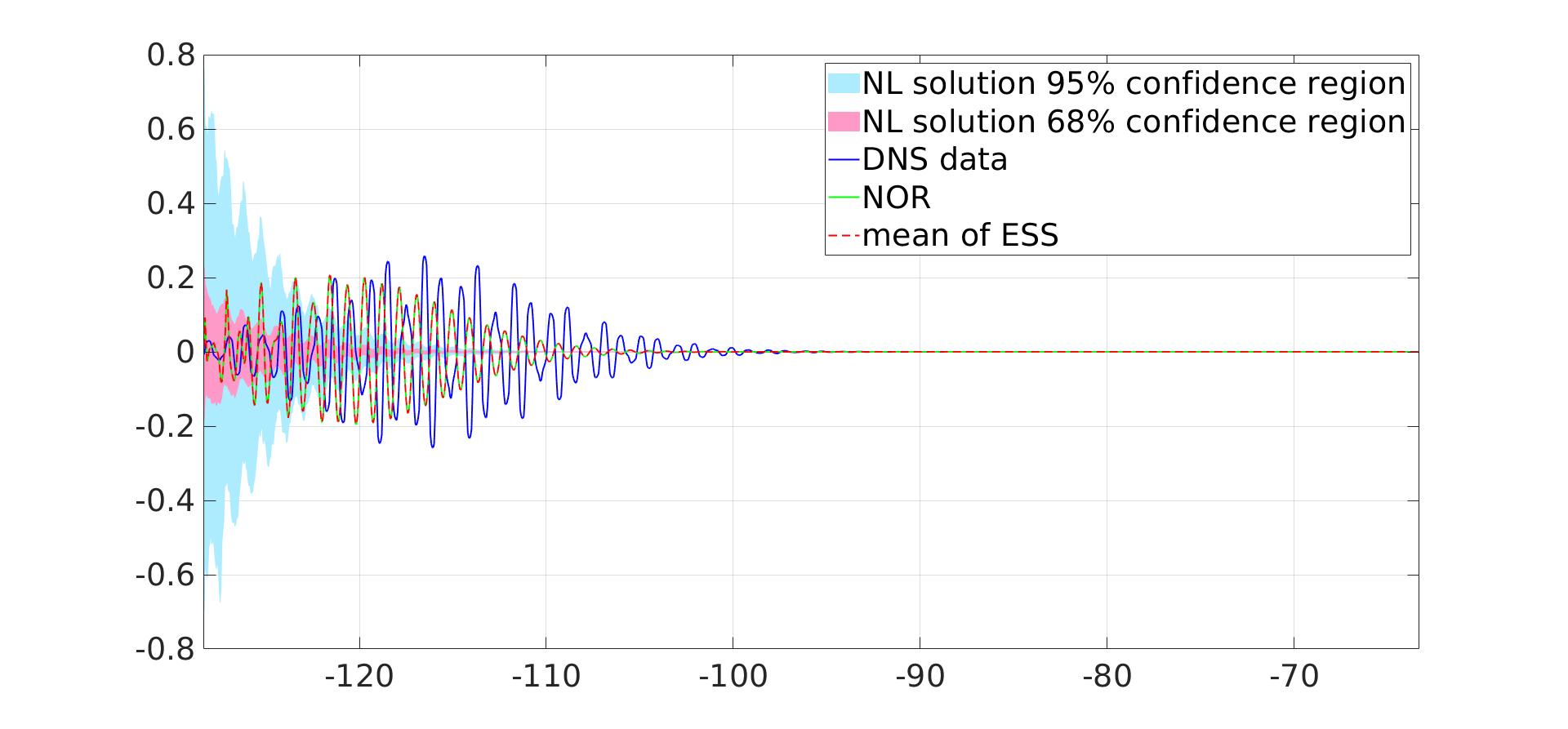}}
\subfigure[$\omega=5$, t=100.0, $l_{gp}=40$.]{\includegraphics[width=.48\columnwidth]{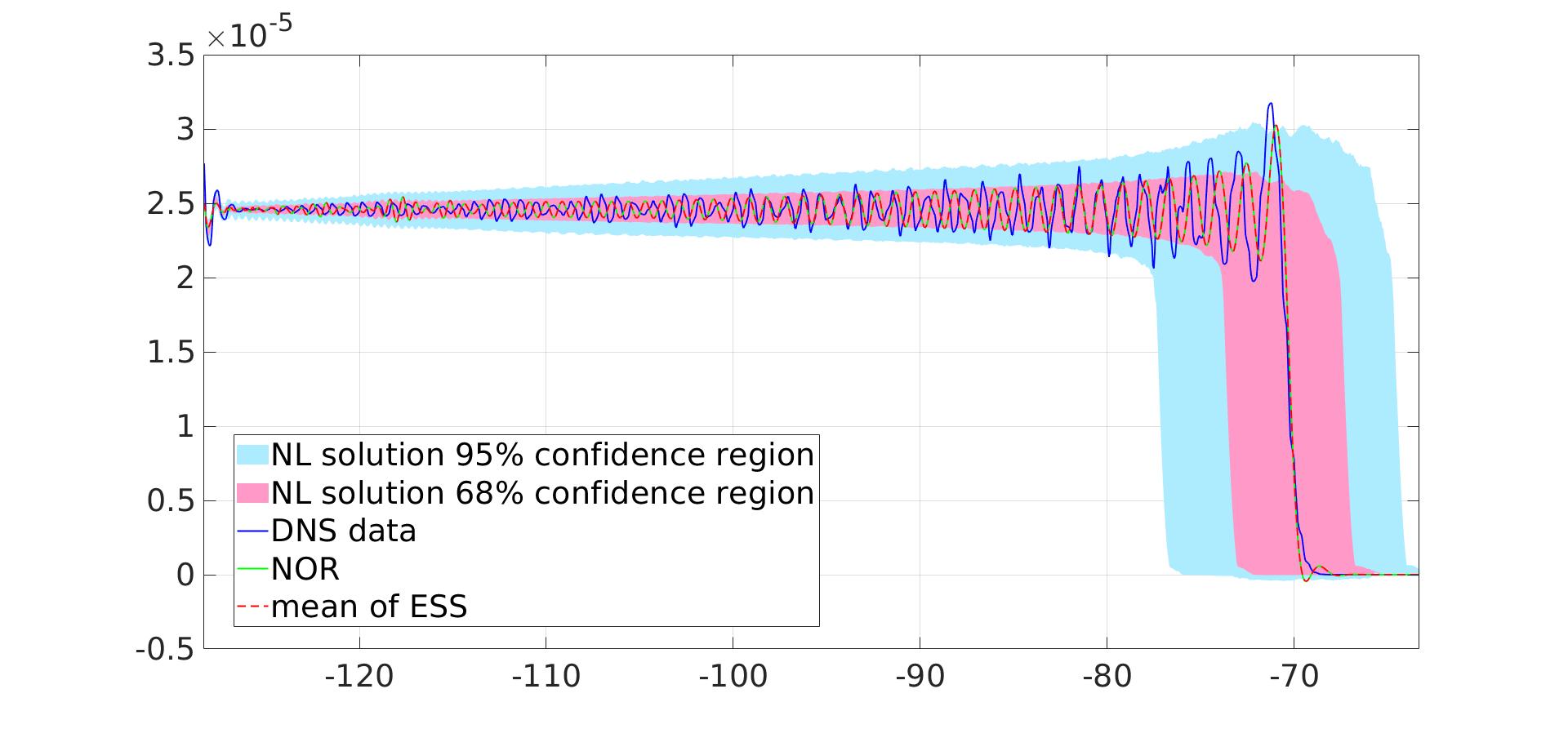}}
\subfigure[$\omega=5$, t=100.0, $l_{gp}=0.15625$.]{\includegraphics[width=.48\columnwidth]{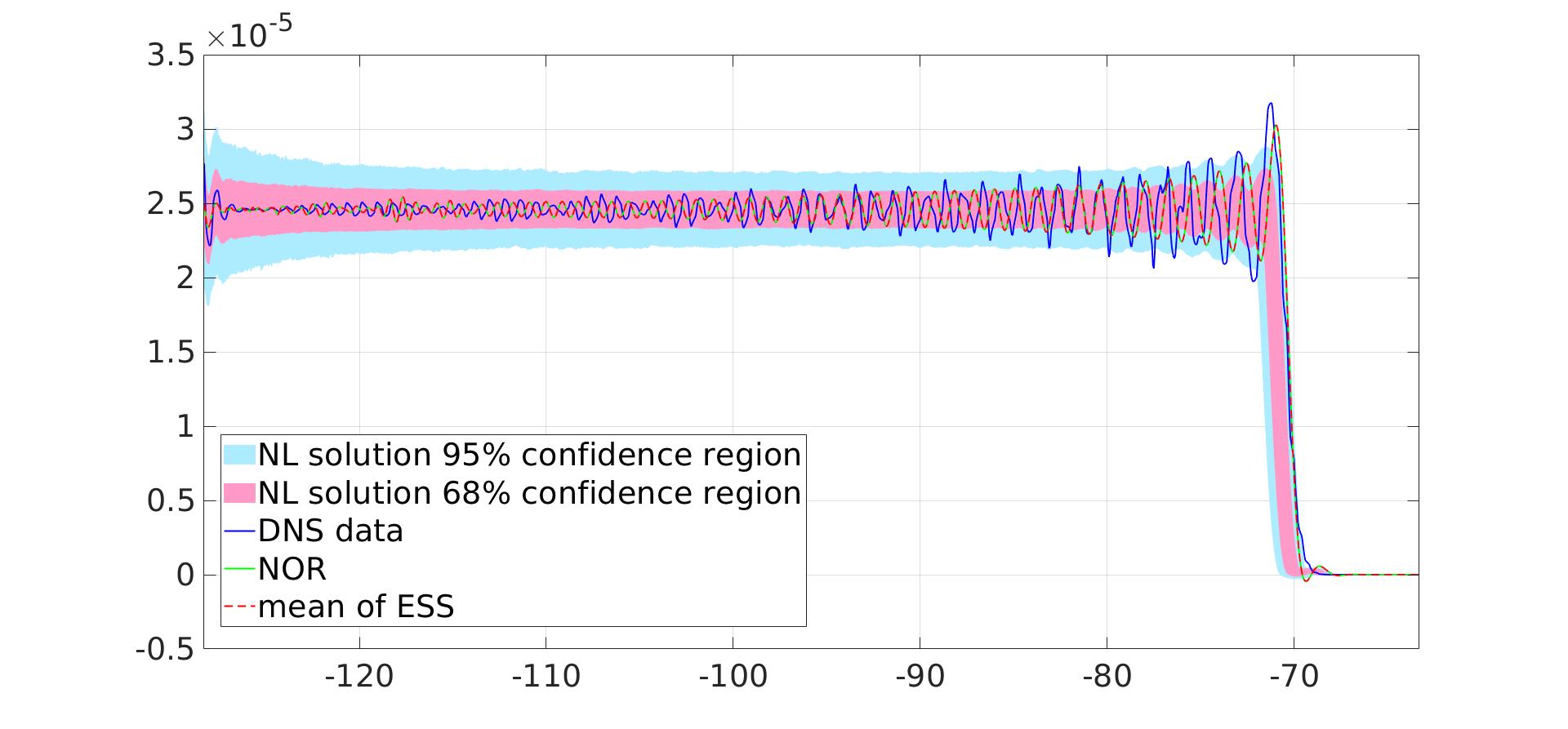}}
\caption{Validation on wave packet for the periodic material at the last time step $t=100.0$. The columns correspond to different correlation length $l_{gp}$ and the rows correspond to different frequencies $\omega$. }
\label{fig:wppf}
\end{figure}

Finally, we provide the prediction of a wave packet, a waveform that is substantially different from the training data. With this setting, we consider an extrapolation scenario: the learnt model is employed to generate a long-term simulation up to $t=100$, which is $50\times$ the training time interval. In Fig.\ref{fig:wppf}, we plot the results using $l_{gp}=40$ and $0.15625$. For a low-frequency wave ($\omega=2.0$), $l_{gp}=0.15625$ works better, in the sense that the predictive distribution more closely follows the small features of the wave, roughly reflecting the local magnitude of the discrepancy between the two solutions. On the other hand, the case with $l_{gp}=40$ fails to do so, with a predictive distribution that broadly encompasses the two solutions, but does not capture the small-scale structure. In both cases, the confidence region fully covers the ground truth, providing a conservative estimation of uncertainty, and avoiding overconfidence, as is the intent of the embedded model error construction. Considering next the $\omega=3.9$ case, a frequency close to the band gap, we find that the larger correlation length $l_{gp}=40$ provides better predictions. Here, the case with smaller correlation length $l_{gp}=0.15625$ suffers from a mismatch between the uncertainty prediction, the single push-forward of the mean of ESS, and the ground truth. This observation is consistent with the results in Fig.\ref{fig:periodic_vg}, where the band gap shifts from the ground-truth band gap and the confidence region fails to cover the DNS data. %\hnn{I note that Fig.\ref{fig:periodic_vg} has no results for the $l_{gp}=0.15625$ case. I am not sure how we're comparing the present case results to those in Fig.\ref{fig:periodic_vg}.} %\YF{[Here I was just trying to compare the results of larger and smaller correlation length. The smaller the $l_{gp}$ is, the more the group velocity shifts from the true value. We didn't provide results of $l_{gp}$=0.15625 in Fig.7 because at that $l_{gp}$, the results oscillate so much and only provides a few meaningful curves (lots of them give negative velocity). Currently I'm tuning the parameters in the group velocity code to see if it could help to give more meaningful samples. ]}
%Therefore, at this frequency of $\omega=3.9$ the predicted wave generated by the ENOR model almost stops propagating, thus the confidence region fails to cover the ground truth. %which means the wave travels for a much shorter distance compared with DNS data. 
%\hnn{I suggest removing this last sentence ... I am not sure that it's adding anything frankly.} \YF{[Removed.]}
For the frequency ($\omega=5.0$) that is much larger than the band gap frequency, the stress wave is anticipated to stop propagating. In this setting we find that both small and large correlation length cases work well. The best $l_{gp}$ values for the validation samples are also provided in Fig.\ref{fig:periodic_crps}, where the optimal correlation length $l_{gp}$ varies depending on the frequency and wave type.
%\hnn{Is this figure useful? If we're not discussing any findings from it, then it's best to remove it.}\YF{[The discussion is in the below paragraph.]} 
%It is worth mentioning that for $\omega=4.0$, the value of the CRPS is in the scale of $1e-7$ for all the $l_{gp}$, therefore there's no big difference between the cases. 
%\hnn{I suggest removing this last sentence. To have the CRPS so small makes me think it might be somehow a degenerate case?} \YF{[Sentence removed. I think the reason for small CRPS is because of the magnitude of the wave in that case is so small (as in Fig.9), which is always considered as 'propagation stopped'.]}

In conclusion, the correlation length for the embedded model error impacts the results differently for different waves, so that the optimal choice would need to be selected according to the purpose of the task. A frequency/waveform-dependent $l_{gp}$ might be of interest. We leave this as a possible future direction. In the following, we choose the best $l_{gp}$ according to the fidelity of prediction of group velocity in Fig.\ref{fig:periodic_vg}, choosing $l_{gp}=2L=40$ as the optimal correlation length.

\subsubsection{Comparison with the Baseline}

\begin{figure}[htp]
\centering
\subfigure[ENOR PP (or PFP) for sample 10 from type 1 data at $t$=2.0.]{\includegraphics[width=.48\columnwidth]{./tpf_s10_lam05.jpg}}
\hspace{0.5em}
\subfigure[ENOR PP (or PFP) for sample 3 from type 2 data at $t$=2.0.]{\includegraphics[width=.48\columnwidth]{./tpf_s23_lam05_jun25.jpg}}
\subfigure[BNOR PFP for sample 10 from type 1 data at $t$=2.0.]{\includegraphics[width=.48\columnwidth]{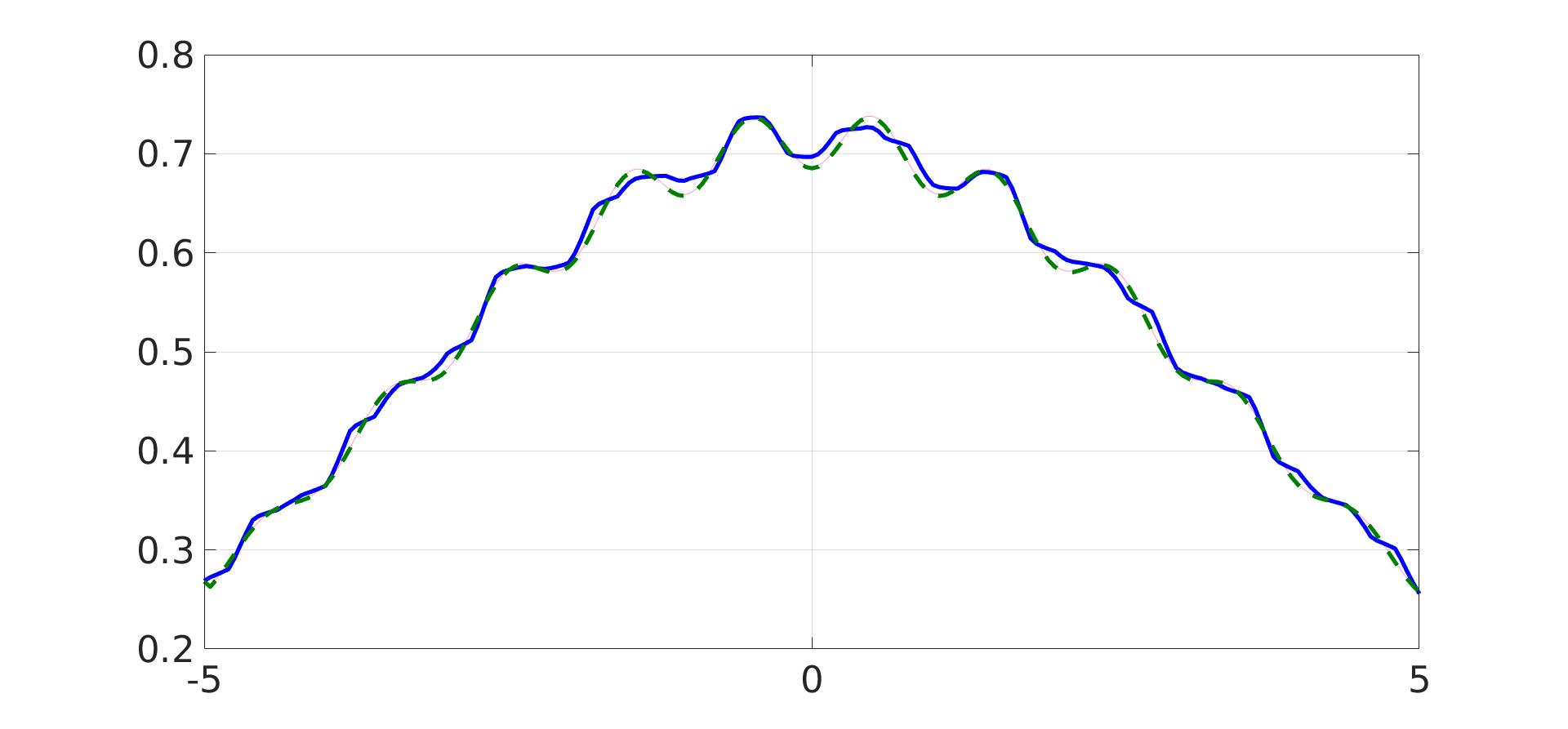}}
\hspace{0.5em}
\subfigure[BNOR PFP for sample 3 from type 2 data at $t$=2.0.]{\includegraphics[width=.48\columnwidth]{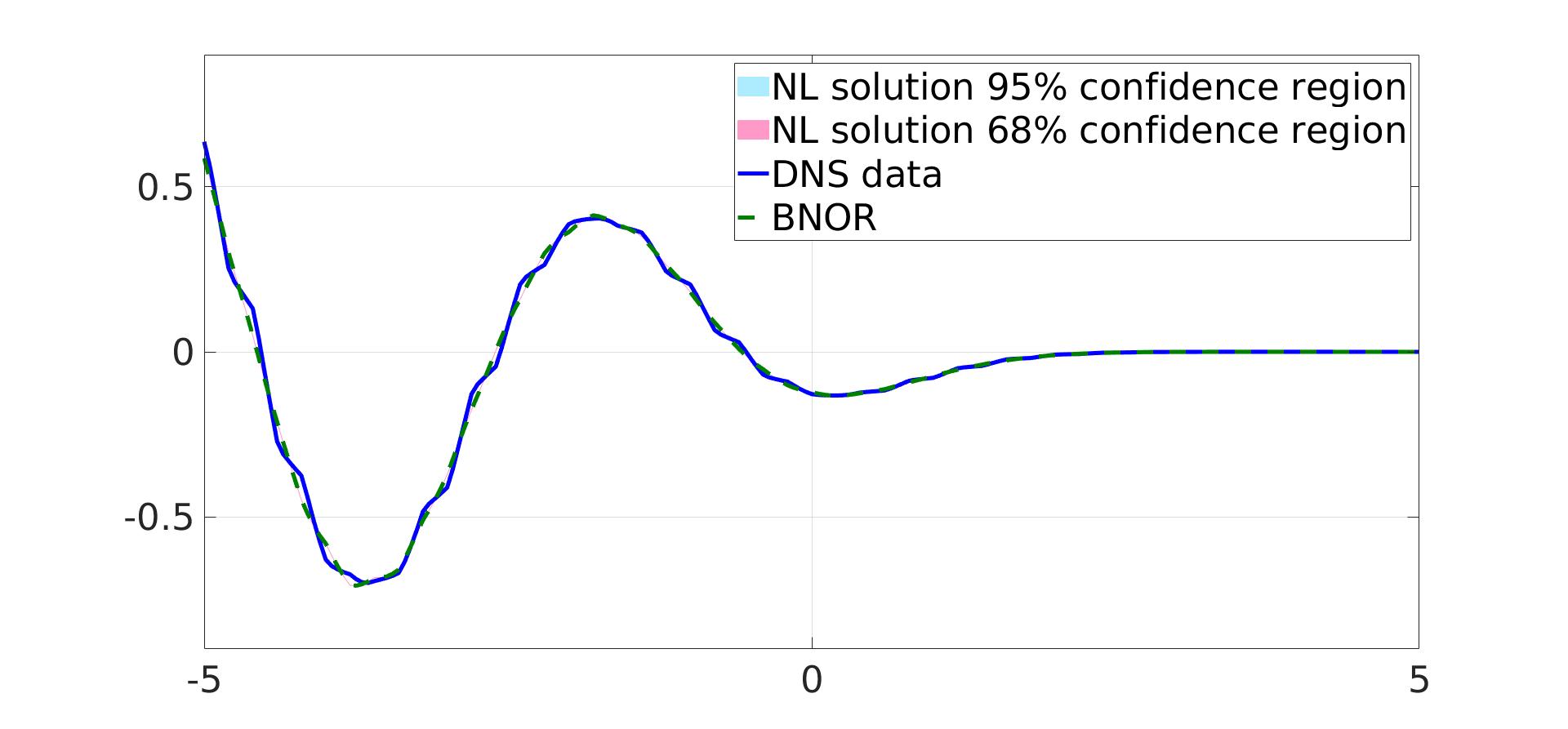}}
\subfigure[BNOR PP for sample 10 from type 1 data at $t$=2.0.]{\includegraphics[width=.48\columnwidth]{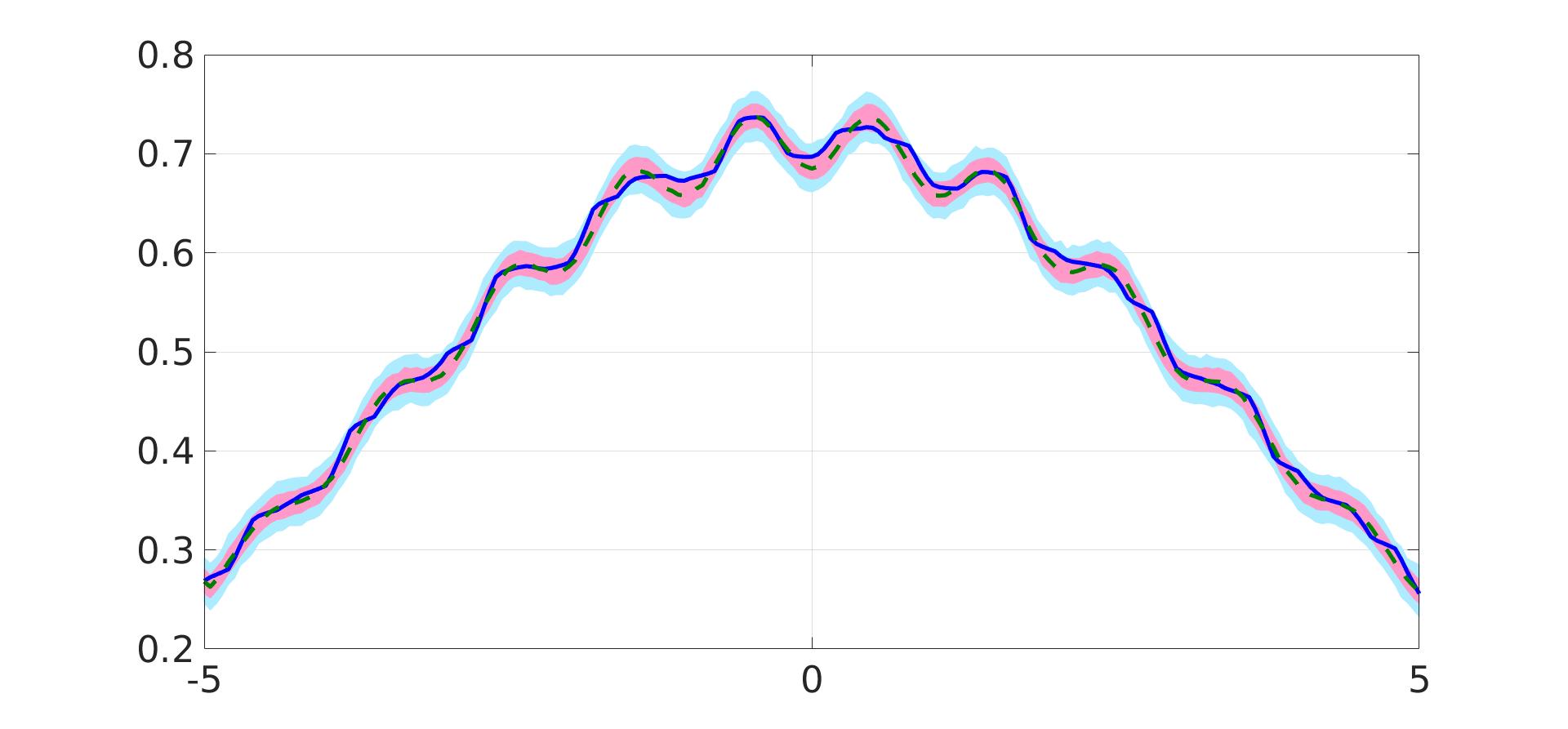}}
\hspace{0.5em}
\subfigure[BNOR PP for sample 3 from type 2 data at $t$=2.0.]
{\includegraphics[width=.48\columnwidth]{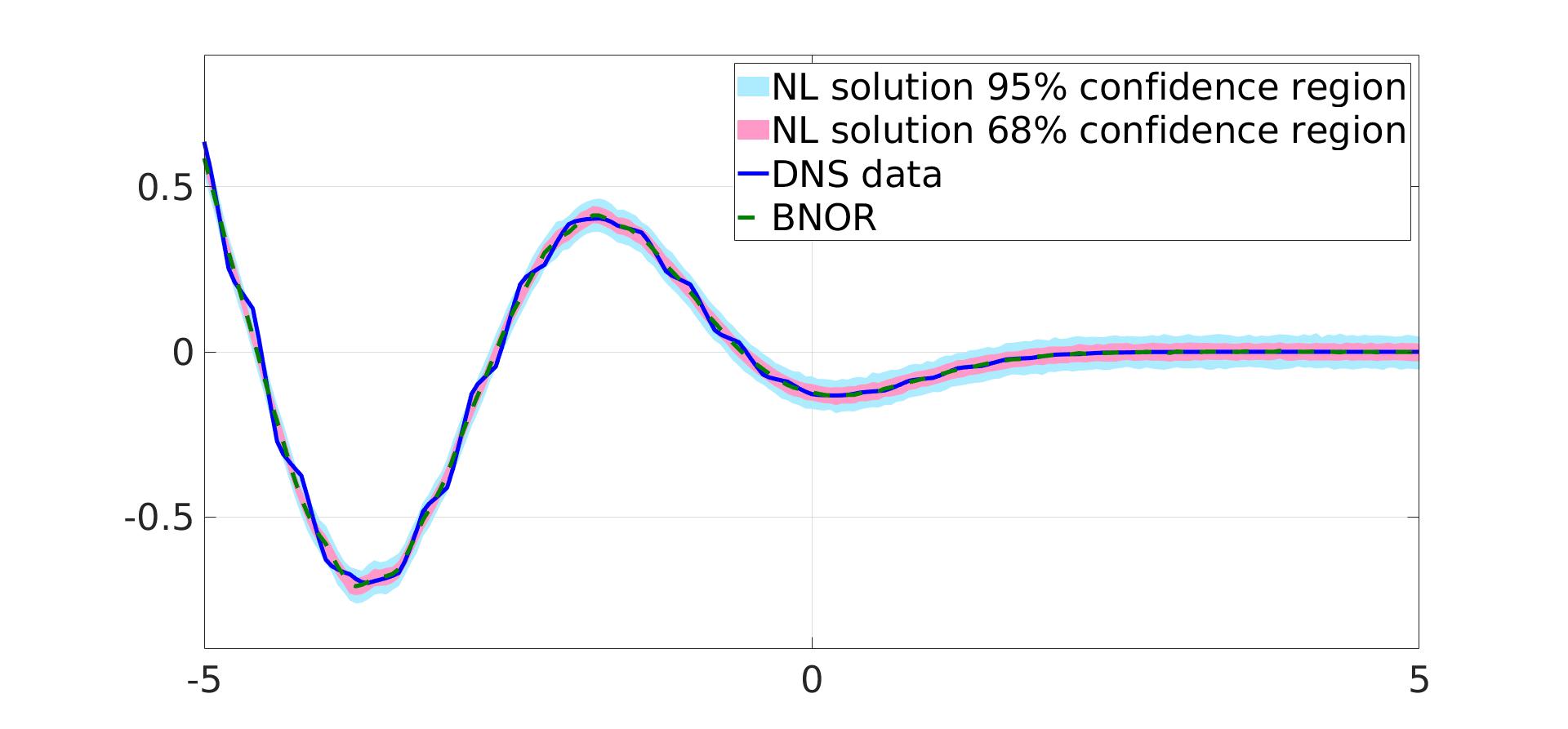}}
\caption{Comparison of posterior uncertainty (PFP and PP) between ENOR and BNOR for periodic material. The columns correspond to different samples in training data and the rows correspond to different methods.}
\label{fig:pfppp}
\end{figure}

To further illustrate the efficacy of the present ENOR construction in capturing model error, we compare the posterior uncertainty on model predictions (with $l_{gp}=40$) to the results from BNOR~\cite{fan2023bayesian}, where an additive \emph{iid} noise model was used. Results are shown in Fig.\ref{fig:pfppp} for two training data samples at a given time instant $t=2.0$. We compare the equivalent PP/PFP ENOR results with the corresponding BNOR PP and PFP results.
%Note that even this case has the highest CRPS (0.0048) among the $l_{gp}$'s, 
From Figs.\ref{fig:pfppp} (c) and (d), one can see that the BNOR PFP exhibits an almost negligible confidence region. On the other hand, the BNOR PP, shown in Figs.\ref{fig:pfppp} (e) and (f), exhibits higher uncertainty with a uniform-width confidence region around the mean prediction, as is expected given the additive \emph{iid} noise. Compared with these two baseline results, the uncertainty given by the ENOR PP/PFP, shown in Figs.\ref{fig:pfppp} (a) and (b), is somewhat more adaptive, exhibiting a degree of uncertainty that approximately tracks the discrepancy between the mean prediction and the DNS data, highlighting the effectiveness of the embedded model error setting. 
%This can be addressed especially in the zeros region, such as in sample 23, where the surrogate model matches closely with the DNS data. 
To provide a further quantitative comparison, we note that the BNOR results at $t=2.0$ in \cite{fan2023bayesian} exhibit a PFP CRPS of 0.051 and a PP CRPS of 0.039, which are both roughly 10$\times$ higher than the present ENOR result in the worst case (CRPS of 0.0048). %\hnn{We shouldn't report the best case, but rather the worst case, to be convincing.} 
This again illustrates that the ENOR model provides an uncertainty that better reflects the discrepancy between the mean model prediction and the ground truth.  

\subsubsection{Parametric Uncertainty versus Model Error}\label{sec:puvme}
\begin{figure}[htp]
\centering
\subfigure[Sample 10 from type 1 data at $t=2.0$. Confidence region using 100 effective samples and 100 realization each.]{\includegraphics[width=.48\columnwidth]{./tpf_s10_lam05.jpg}}
\hspace{0.5em}
\subfigure[Sample 3 from type 2 data at $t=2.0$. Confidence region using 100 effective samples and 100 realization each.]{\includegraphics[width=.48\columnwidth]{./tpf_s23_lam05_jun25.jpg}}
\subfigure[Sample 10 from type 1 data at $t=2.0$. Confidence region using fixed values of kernel parameters $\Cb$ and $\sigma_{gp}$ with 1000 GP realizations.]{\includegraphics[width=.48\columnwidth]{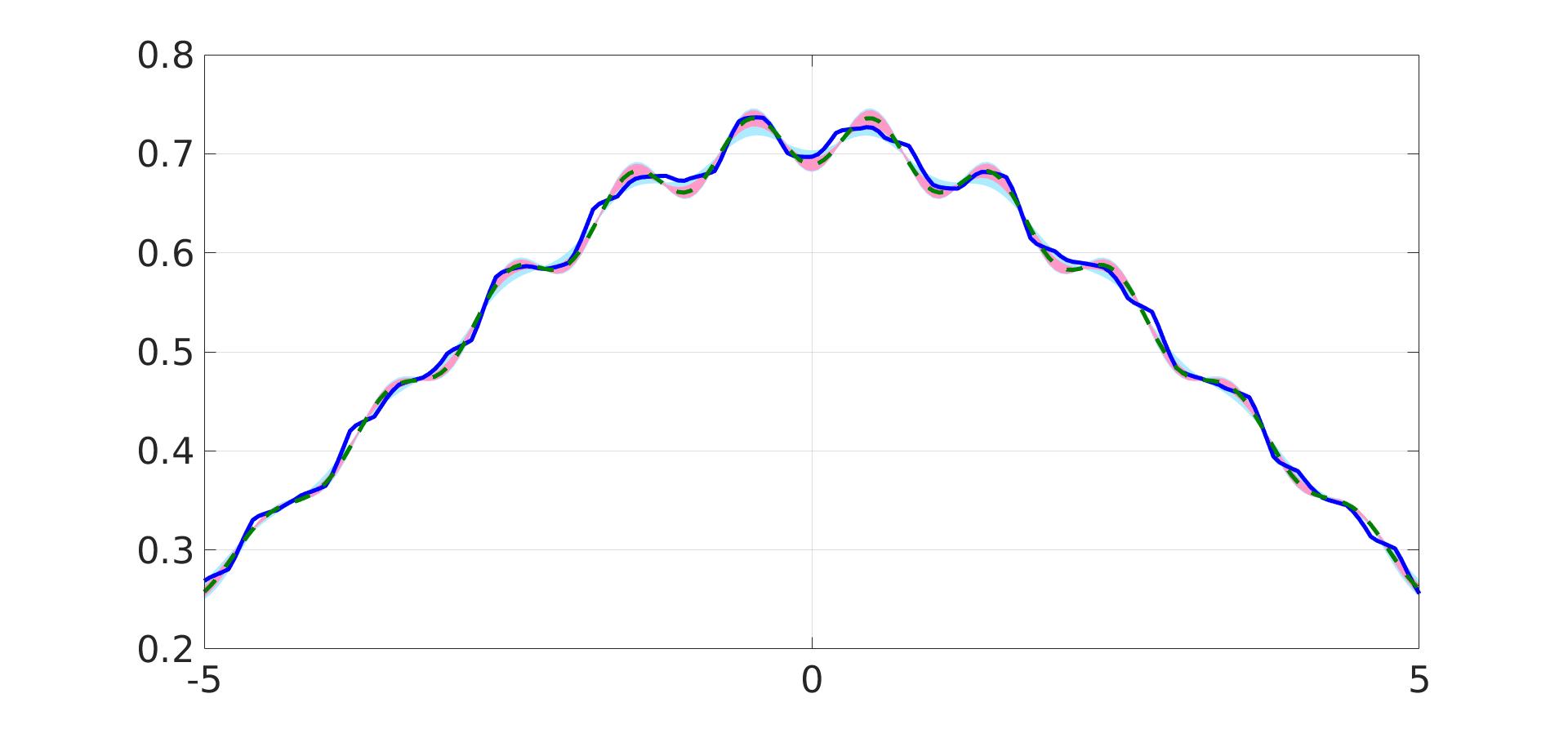}}
\hspace{0.5em}
\subfigure[Sample 3 from type 2 data at $t=2.0$. Confidence region using fixed values of kernel parameters $\Cb$ and $\sigma_{gp}$ with 1000 GP realizations.]{\includegraphics[width=.48\columnwidth]{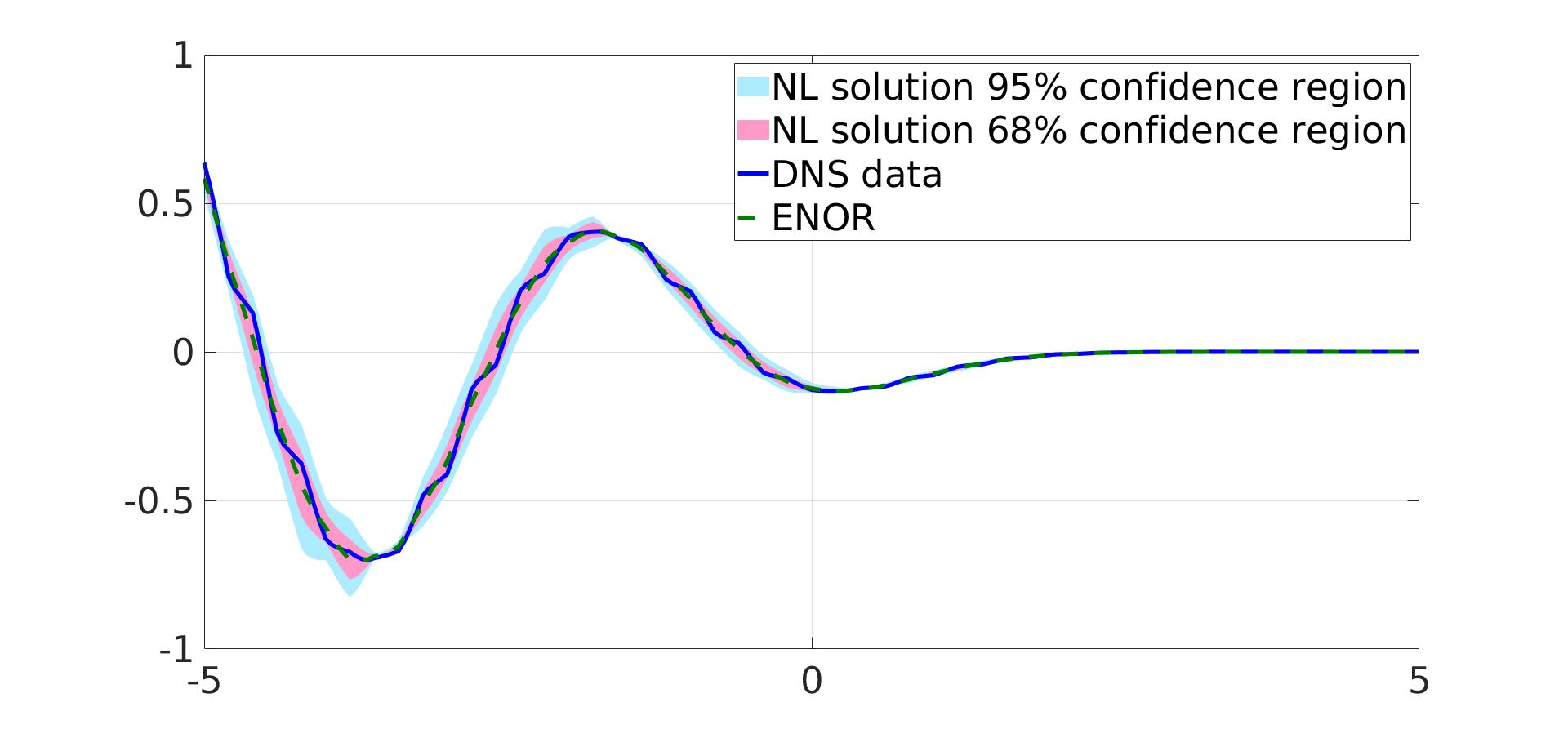}}
\subfigure[Sample 10 from type 1 data at $t=2.0$. Confidence region using 100 effective samples of $\Cb$ and $\sigma_{gp}$ without GP (i.e. with $\xi_m$=0).]{\includegraphics[width=.48\columnwidth]{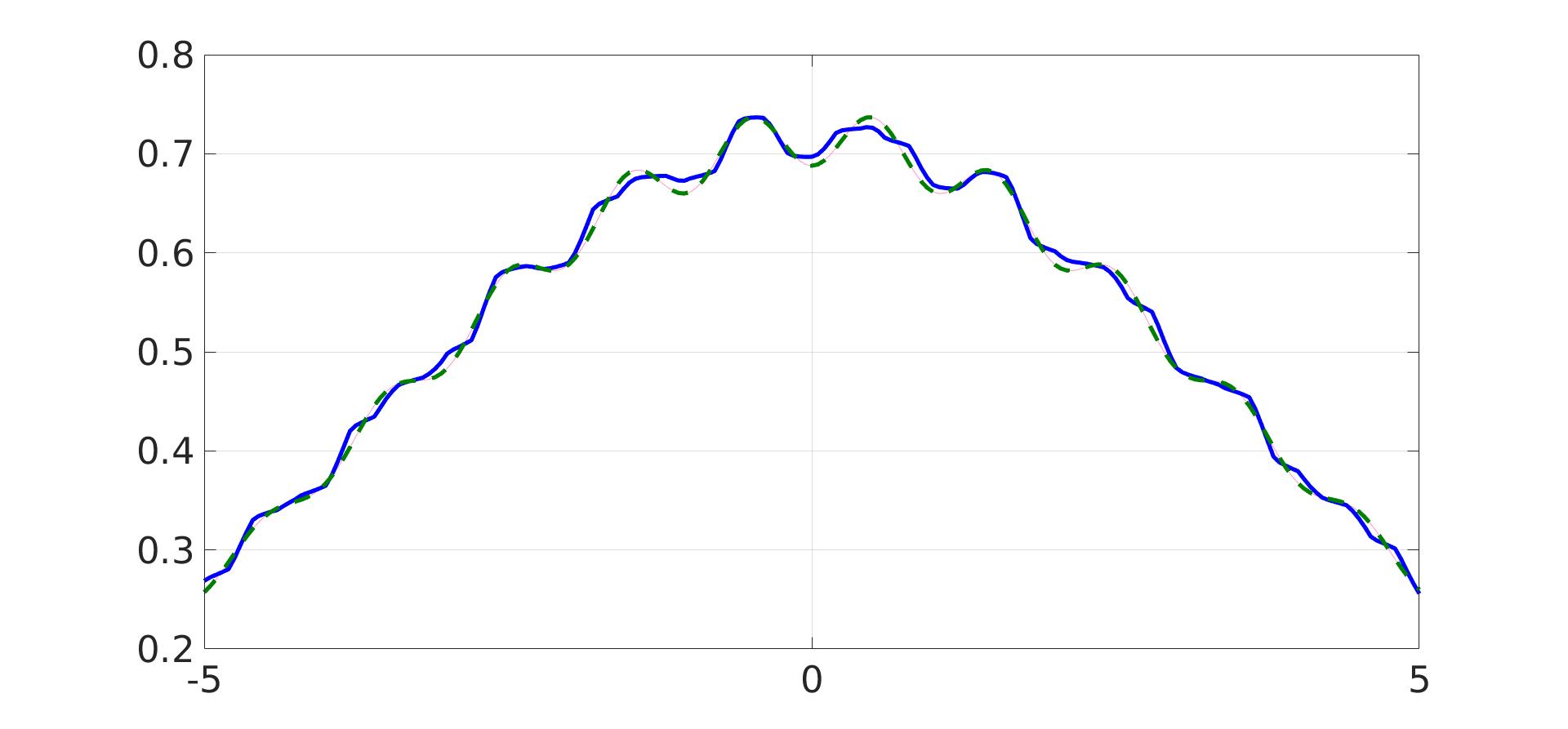}}
\hspace{0.5em}
\subfigure[Sample 3 from type 2 data at $t=2.0$. Confidence region using 100 effective samples of $\Cb$ and $\sigma_{gp}$ without GP (i.e. with $\xi_m$=0).]{\includegraphics[width=.48\columnwidth]{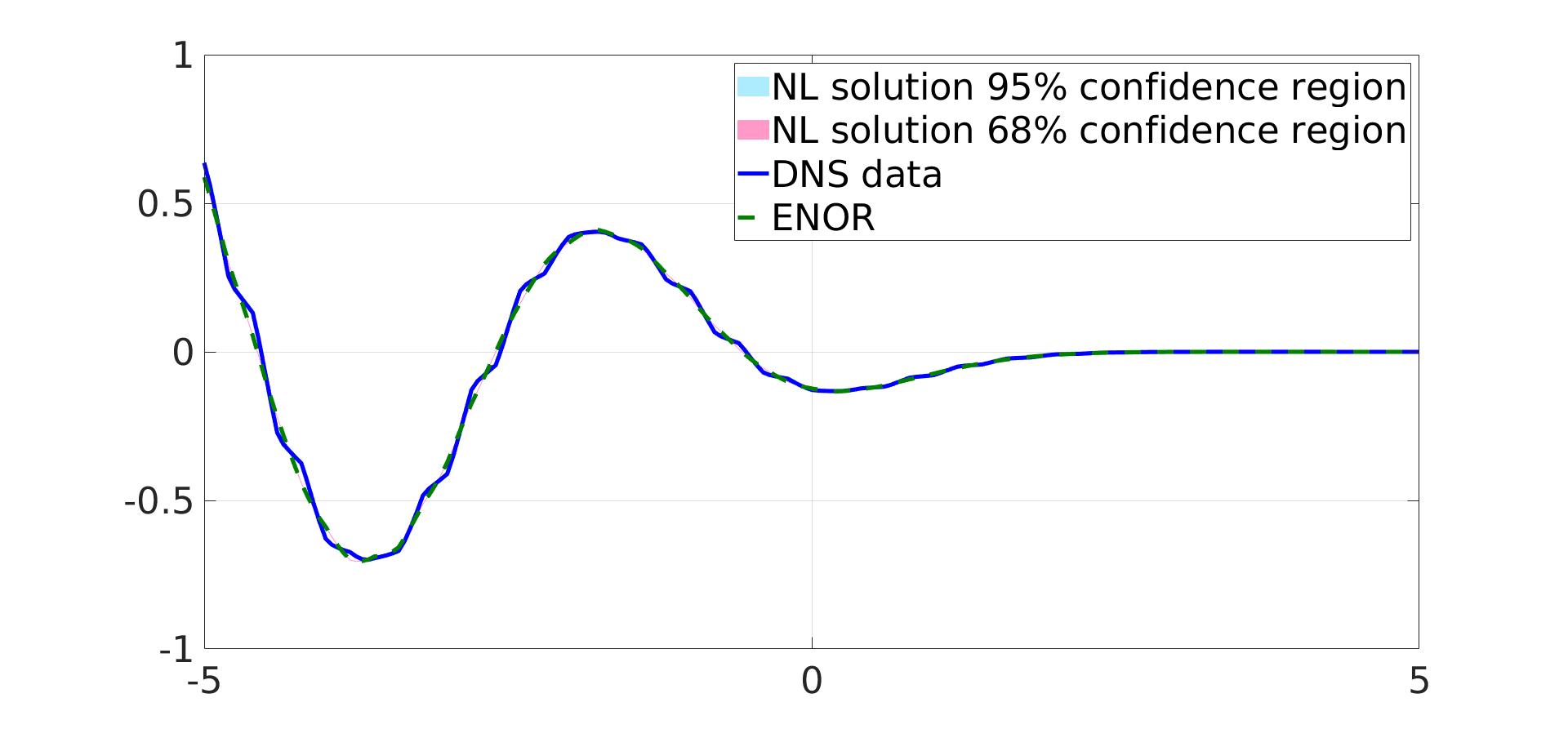}}
\caption{An illustration of the relative impact of parametric uncertainty and model error on resulting predictive uncertainty, in the periodic material. The columns correspond to different samples in the training data and the rows correspond to posterior prediction using different sources of uncertainty. }
\label{fig:gsa}
\end{figure}

It is instructive to examine the role of parametric uncertainty versus model error in the resulting uncertainty in model predictions. We illustrate in Fig.\ref{fig:gsa} the posterior predictive uncertainty under the following three scenarios.
\begin{enumerate}
\item In plots (a) and (b), we consider uncertainty from all sources, by sampling realizations from both the embedded error correction term $\xi$ and the marginal distribution on kernel parameters.
\item In plots (c) and (d), we consider the uncertainty from the embedded error correction term $\xi$ only, using deterministic kernel parameters.
\item In plots (e) and (f), we neglect the embedded error correction term but sample the kernel parameter $\Cb$ from the learnt marginal distribution. As such, only the uncertainty from the marginal distribution on kernel parameters is considered. 
\end{enumerate}
By comparing the confidence regions from these three settings, one can observe that results from scenarios 1 and 2 almost coincide, while the predicted uncertainty in scenario 3 is negligible. This indicates that nearly all of the predictive uncertainty comes from the embedded model error. This is typical in Bayesian estimation of one model against another in the absence of data noise and when there is a sufficiently large amount of data.

\subsection{Example 2: Random Microstructure}
%\subsubsection{Data generation and settings}

To investigate the performance of the ENOR model on more complicated microstructures, we consider here a disordered heterogeneous bar, where the layer lengths of the two materials are random. In particular, the layer sizes are two uniformly distributed random variables: $b_1,b_2\sim \mathcal{U}[(1-D){b},(1+D){b}]$, with the average layer size $b=0.2$ and the disorder parameter $D=0.5$. The density for both components is still set as $\rho=1$ and their Young's moduli are set at $E_1=1$ and $E_2=0.25$.

To generate the training dataset, we consider the high-fidelity data under the same settings as in data types 1 and 2 of the periodic bar case. Compared with the periodic microstructure case, from the group velocity generated by the DNS simulations we note that the band stop generally occurs at a lower frequency in the random microstructure case. In fact, for the microstructure considered here, an estimated band stop frequency $\omega_{bs}\approx 3$ can be obtained from the DNS simulations. Therefore, for the validation data set, we study wave packets with frequencies $\omega=1,2,3$ and $4$, with the purpose of investigating the performance of our nonlocal surrogate model when the loading frequencies are below ($\omega=1,2$), around ($\omega=3$), and above ($\omega=4$) the estimated band stop frequency $\omega_{bs}$. 

\subsubsection{Results from MCMC Experiments}

To demonstrate the convergence of MCMC in this case, we again pick correlation length $l_{gp}=L/2$ as an example case to test the utility of ENOR in disordered materials. Following the same settings of the periodic microstructure study, we run 6 independent chains. The results pass the same convergence check both visually and quantitatively, where the improved $\hat{R}$ statistics for all the 24 parameters are very close to $1$, with a maximum of $1.0038$, which again verifies the convergence. The chains provide 24,000 draws in total, with approximately $39\%$ acceptance rate on average and the combined chain shows  good mixing. Compared with the periodic bar with inferred $\ln(\sigma_{gp})\approx-2.0$, a slightly larger kernel variation, $\ln(\sigma_{gp})\approx-1.8$, is obtained here. That suggests that the randomness in the material with disordered microstructure results in a larger model discrepancy relative to the DNS results.

%\begin{figure}[h!]
%\centering
%\subfigure[First parameter]{\includegraphics[width=.30\columnwidth]{Figures/dis_conv_trace0.jpg}}
%\subfigure[Second parameter]{\includegraphics[width=.30\columnwidth]{Figures/dis_conv_trace1.jpg}}
%\subfigure[$\ln(\sigma_{gp})$]{\includegraphics[width=.30\columnwidth]{Figures/dis_conv_traces.jpg}}
%\subfigure[Joint distribution of param 1 and param 2]{\includegraphics[width=.45\columnwidth]{Figures/dis_conv_pdf12.jpg}}
%\subfigure[Joint distribution of param 3 and sigma$_{gp}$]{\includegraphics[width=.45\columnwidth]{Figures/dis_conv_pdf324.jpg}}
%\caption{Convergence check: Trace plot and PDF for the traces. For each trace, acceptance rate $\approx$ 0.42, ESS $\approx$ 1000 (out of 3700 draws).}
%\label{fig:dis_conv_trace}
%\end{figure}

%\begin{figure}[h!]
%\centering
%\subfigure[First parameter]{\includegraphics[width=.30\columnwidth]{Figures/mlda_dis_trace0.jpg}}
%\subfigure[Second parameter]{\includegraphics[width=.30\columnwidth]{Figures/mlda_dis_trace1.jpg}}
%\subfigure[$\ln(\sigma_{gp})$]{\includegraphics[width=.30\columnwidth]{Figures/mlda_dis_traces.jpg}}
%\subfigure[Joint distribution of param 1 and param 2]{\includegraphics[width=.45\columnwidth]{Figures/mlda_dis_pdf12.jpg}}
%\subfigure[Joint distribution of param 3 and sigma$_{gp}$]{\includegraphics[width=.45\columnwidth]{Figures/mlda_dis_pdf324.jpg}}
%\caption{Trace plot and PDF for the combined trace. The acceptance rate $\approx$ 0.42, ESS $\approx$ 6,000 (out of 22,000 draws).}
%\label{fig:disorder_trace}
%\end{figure}

\subsubsection{Impact of GP Correlation Length}

\begin{figure}[htp]
\centering
\subfigure[Kernel at point $x=0.0$ for $l_{gp}=2L$=40.]{\includegraphics[width=.48\columnwidth]{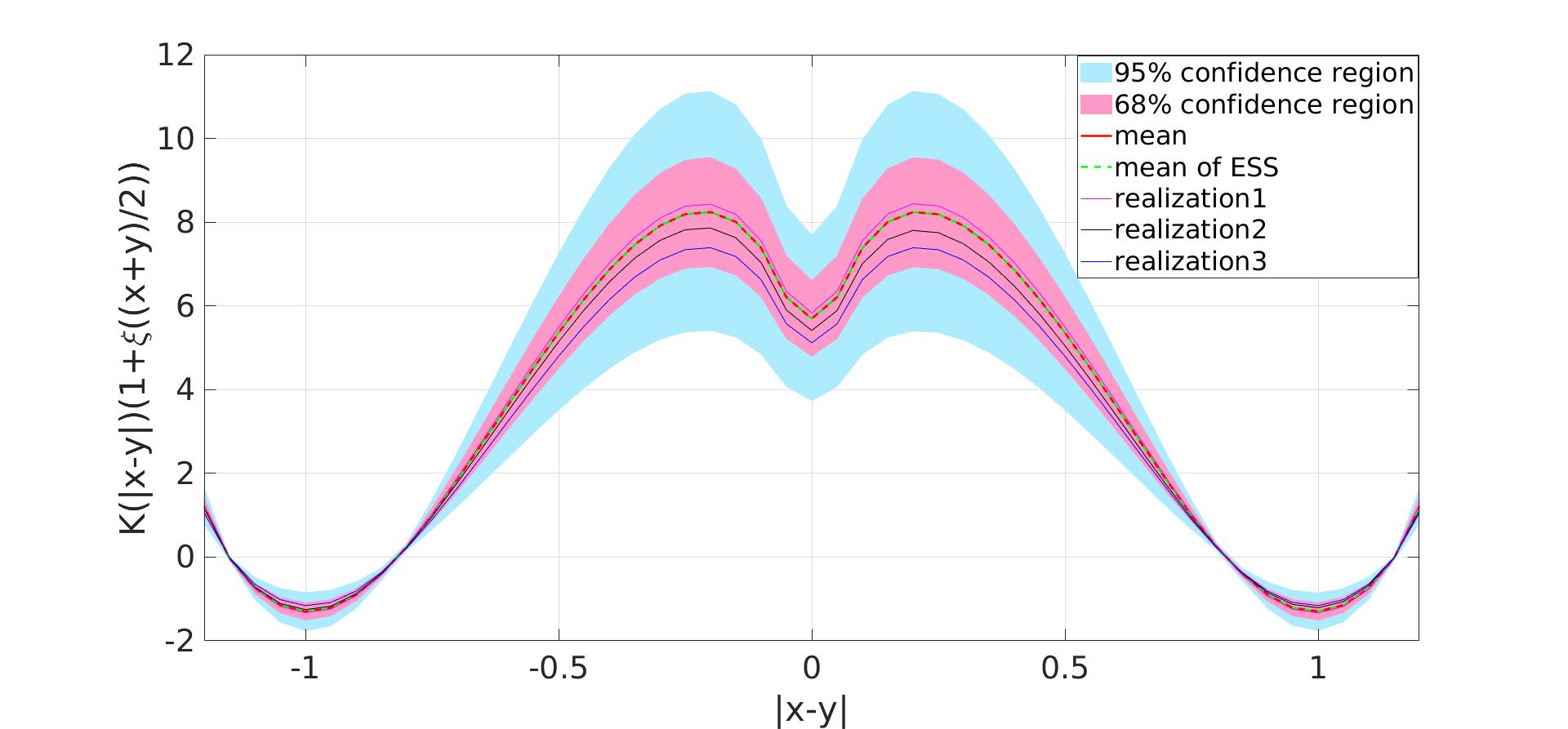}}
\subfigure[Kernel at point $x=0.0$ for $l_{gp}=\dfrac{L}{128}$=0.15625.]{\includegraphics[width=.48\columnwidth]{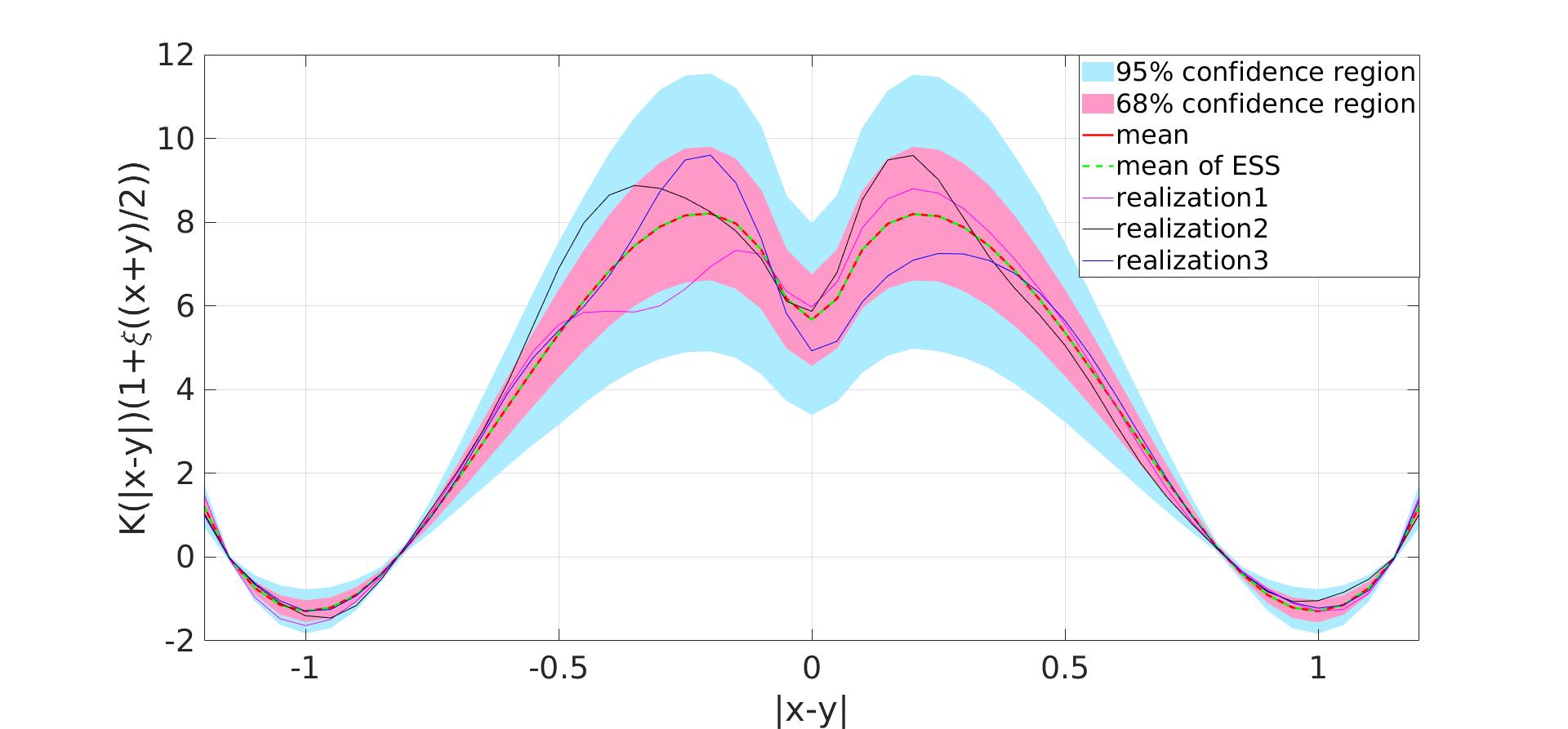}}
\caption{Kernel with uncertainty for a bar with periodic microstructure using two different correlation lengths.}
\label{fig:disorder_ker}
\end{figure}

\begin{figure}[htp]
\centering
\subfigure[Group velocity using $l_{gp}=2L$=40.]{\includegraphics[width=.65\columnwidth]{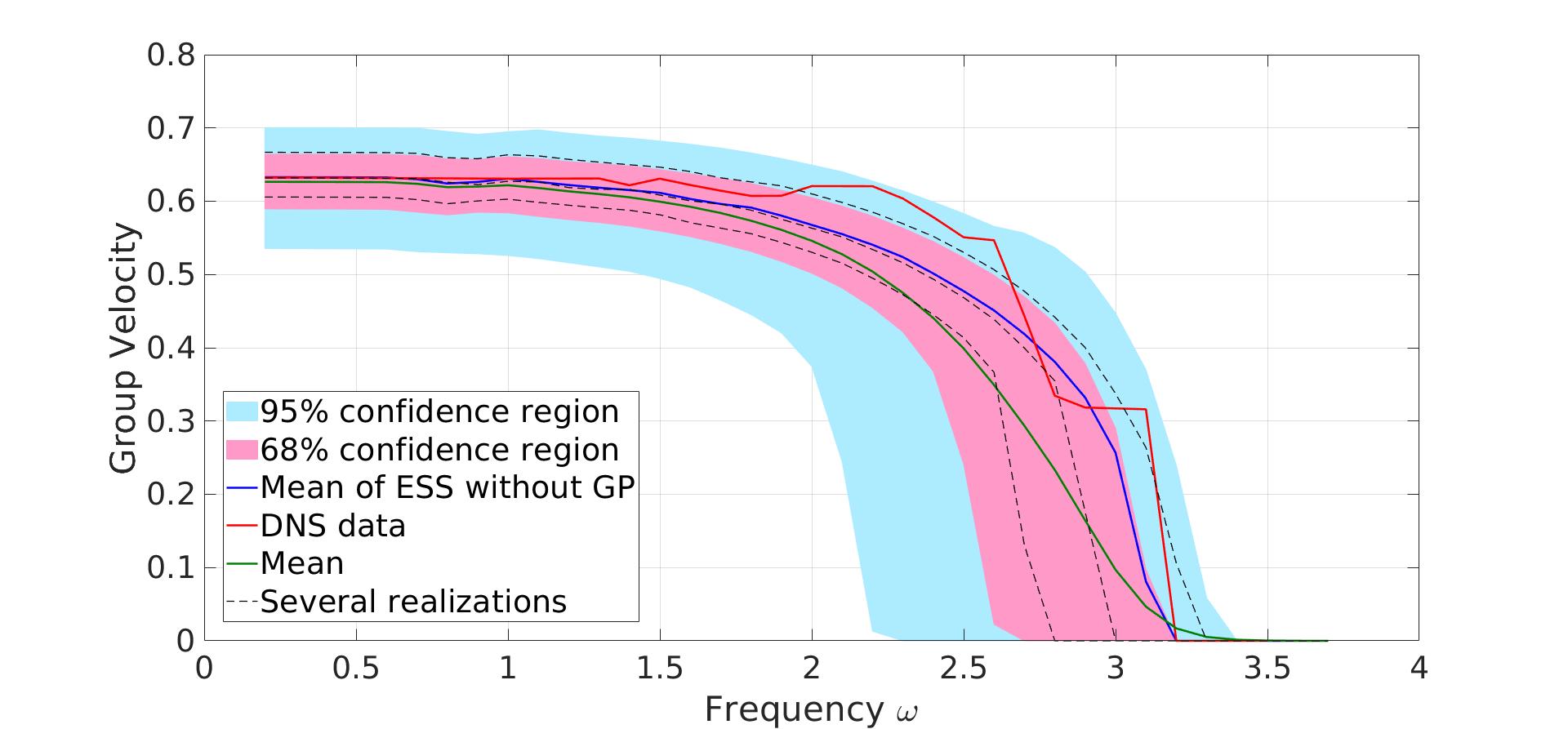}}
\subfigure[Dispersion curve using $l_{gp}=2L$=40.]{\includegraphics[width=.30\columnwidth]{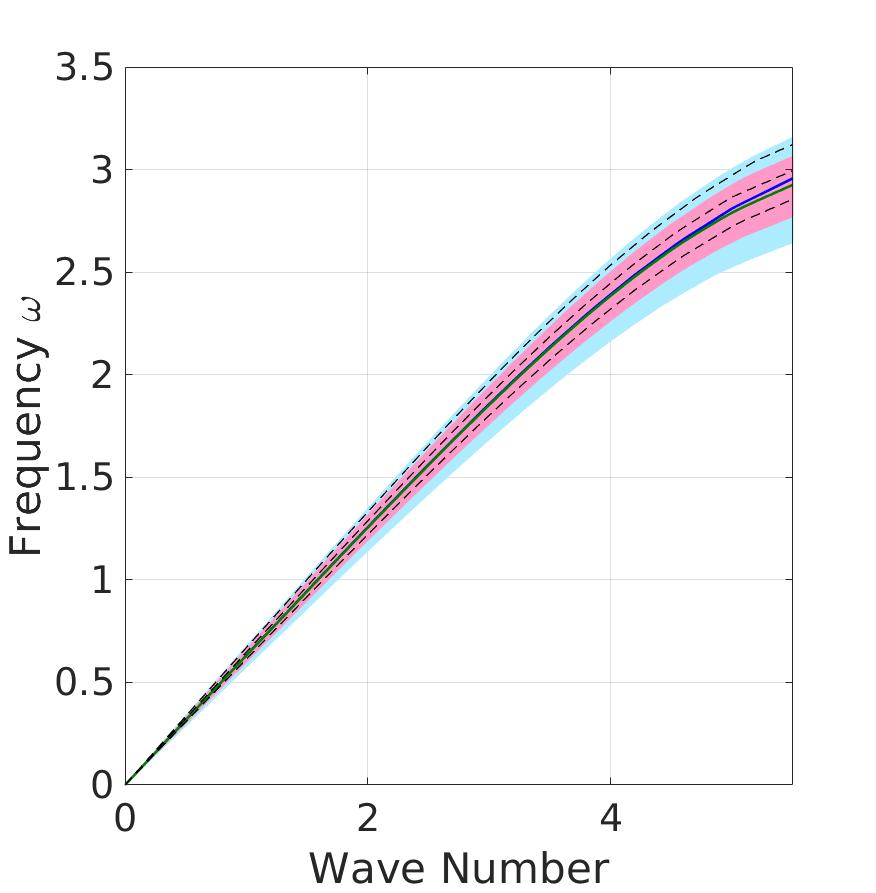}}
\subfigure[Group velocity using $l_{gp}=\dfrac{L}{32}$=0.625.]{\includegraphics[width=.65\columnwidth]{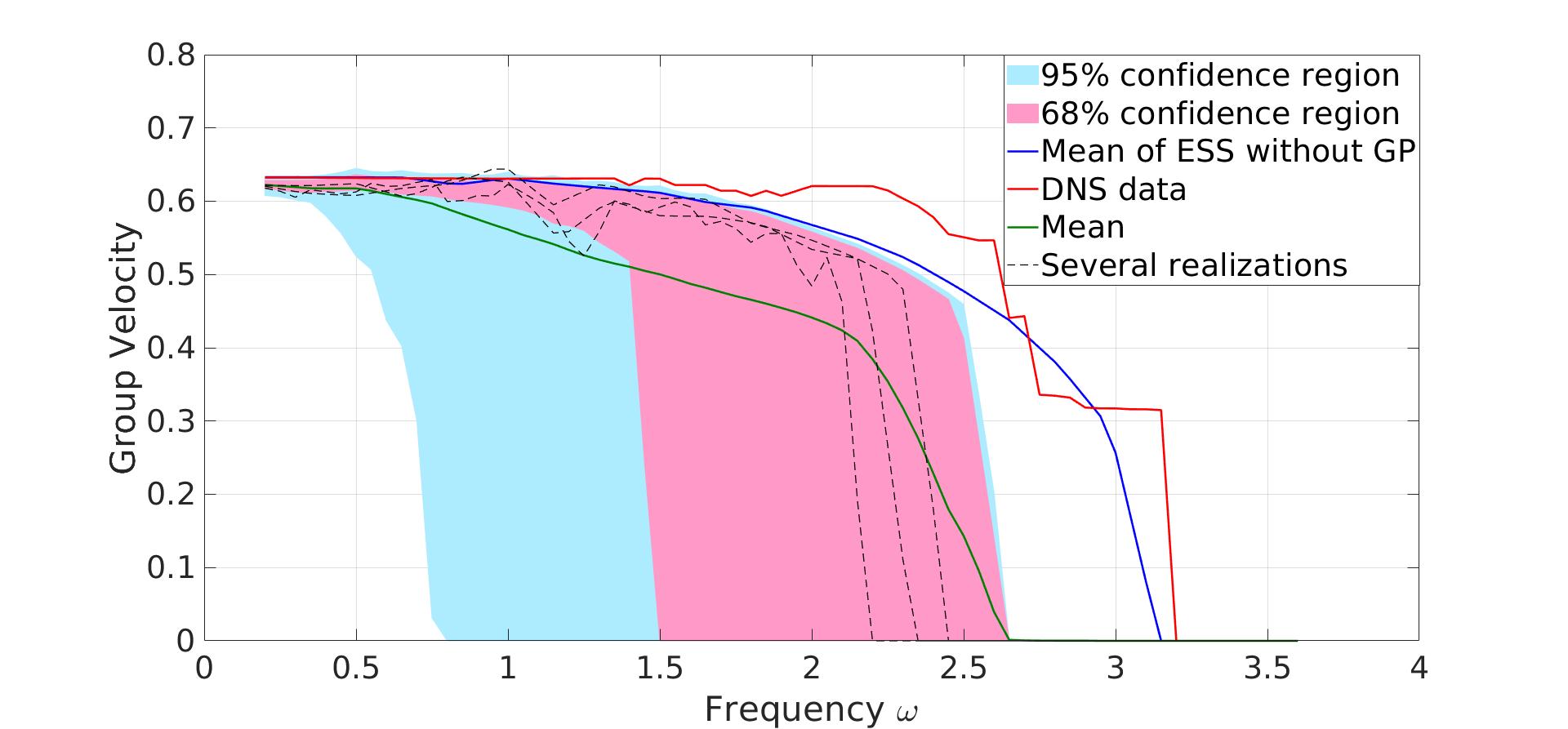}}
\subfigure[Dispersion curve using $l_{gp}=\dfrac{L}{32}$=0.625.]{\includegraphics[width=.30\columnwidth]{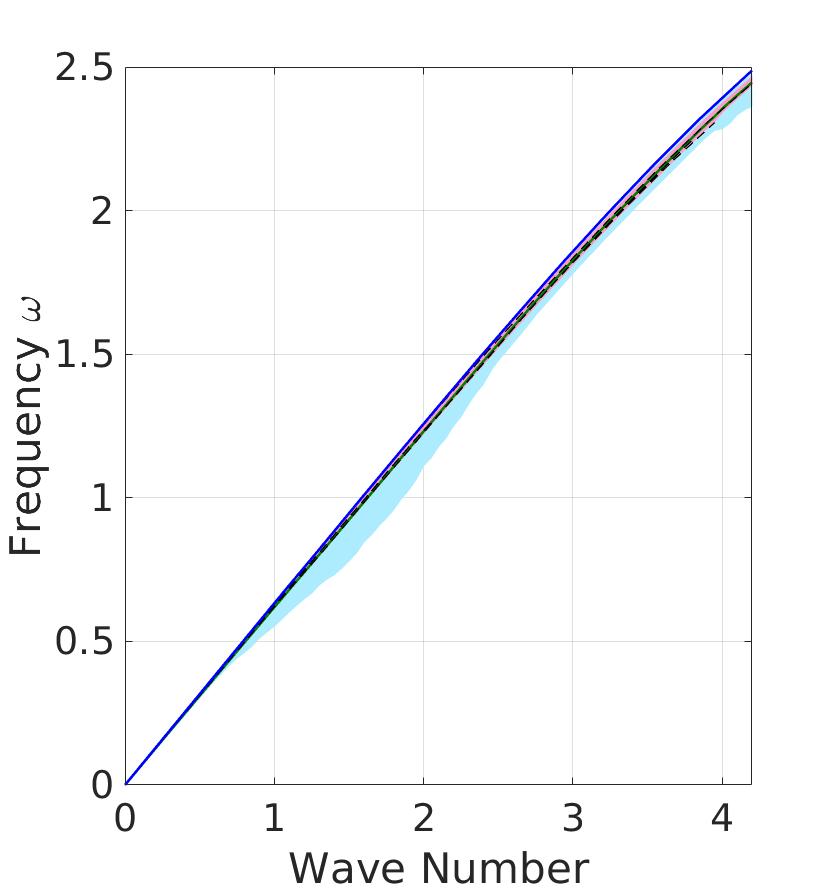}}
%\subfigure[Group velocity using $l_{gp}=\dfrac{L}{32}$=0.625.]{\includegraphics[width=.65\columnwidth]{Figures/dis_vg_lam32_oct21.jpg}}
%\subfigure[Dispersion curve using $l_{gp}=\dfrac{L}{32}$=0.625.]{\includegraphics[width=.30\columnwidth]{Figures/dis_dispersion_lam32_oct21.jpg}}
\caption{Group velocity and dispersion curve with uncertainty for a bar with disordered microstructure using two different correlation lengths.}
\label{fig:disorder_vg}
\end{figure}

We now study the impact of different correlation length ($l_{gp}$) values in the disordered material, again considering $l_{gp}$ values ranging from $0.15625$ to $40$. In Fig.\ref{fig:disorder_ker}, we observe that the uncertainty in the kernel with an embedded GP, $K(|x-y|)(1+\xi(\dfrac{x+y}{2}))$, has the same pattern of mean and confidence region variation as in the periodic bar case, given the same random kernel structure and GP properties. In Fig.\ref{fig:disorder_vg}, we investigate the behavior of the group velocity, and observe that a larger correlation length provides better fitting of the group velocity, and that reducing the correlation length $l_{gp}$ results in a left shift of the band gap in prediction. While this trend was also observed in the periodic microstructure case (see Fig.\ref{fig:periodic_vg}), severer oscillations and larger confidence regions are observed at low frequencies here in the small correlation length $l_{gp}=0.625$ case, possibly because of the higher uncertainty levels introduced by the disordered microstructure in the material. In Figs.\ref{fig:disorder_vg} (b) and (d), we show the dispersion curves of the learnt kernels, noting again that positivity of all dispersion curves indicates physical stability of the learnt models.

\begin{figure}[htp]
\centering
\subfigure[Best $l_{gp}$ for training samples]{\includegraphics[width=.75\columnwidth]{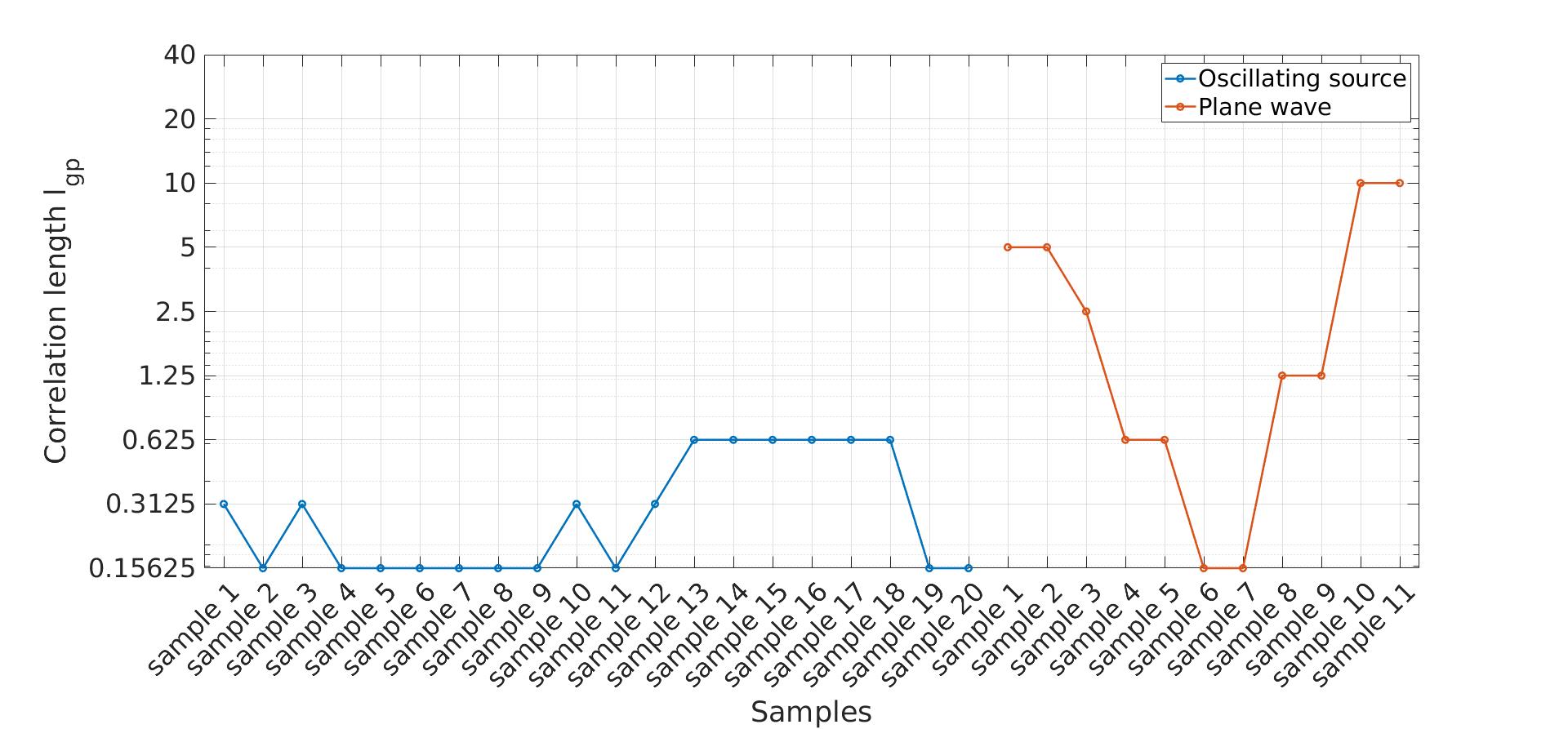}}
\subfigure[Best $l_{gp}$ for validation samples]{\includegraphics[width=.23\columnwidth]{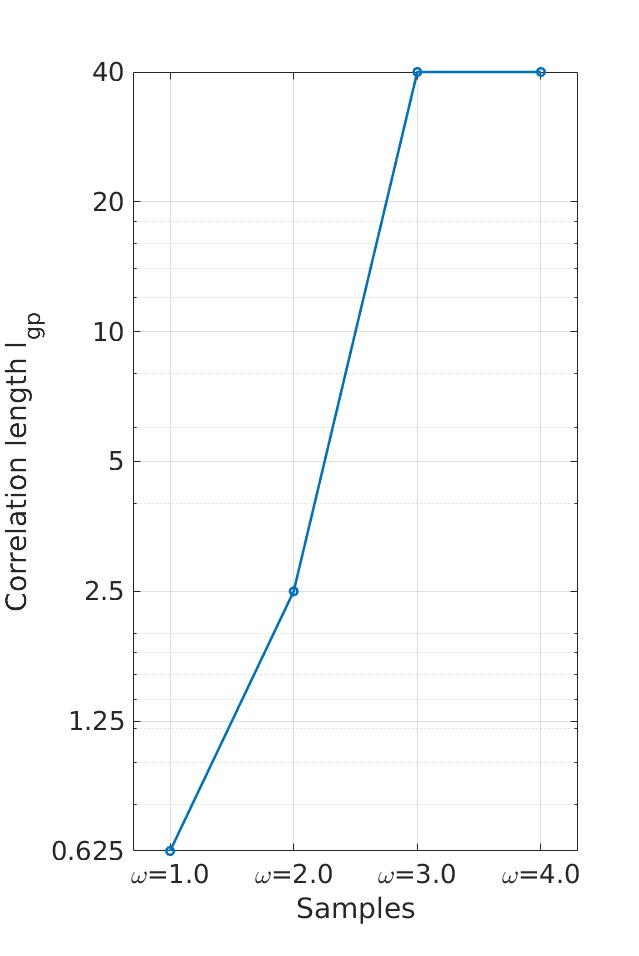}}
\caption{Best $l_{gp}$ (with the lowest CRPS) on training and validation samples for a bar with disordered microstructure.}
\label{fig:disorder_crps}
\end{figure}

Finally, we provide again the prediction of a wave packet for a larger domain and longer time as a validation, not present in the training set. %In Fig.\ref{fig:dis_wppf}, we plot solution results using $l_{gp}=40$ and $0.15625$. 
%Overall, the same conclusion can be reached here: (1) for low frequencies $\omega=1.0$ and $2.0$, the smaller correlation length works better, (2) for frequencies $\omega=3.0$ that is around the band stop, the larger correlation length works better, (3) for frequencies $\omega=4.0$ that is larger than the band stop, the two correlation lengths work equally well, with the CRPS for all the $l_{gp}$ values are in the same small range $\mathcal{O}(10^{-4})$. 
Overall, the same conclusion on the correlation length as section \ref{sec:peri_gp} can be reached for disordered material. We refer the reader to Fig\ref{fig:dis_wppf} in Appendix for more details on this experiment. 
%\hnn{If the results in the figure lead to the same conclusions as above, why not just say that and there is no need for the figure.} \YF{[To be discussed. Should we  move these figures in appendix? ]}
In Fig.\ref{fig:disorder_crps}, we provide the best $l_{gp}$ for all the training and validation samples according to the value of the CRPS. Note that the optimal $l_{gp}$s are different from the ones illustrated in Fig.\ref{fig:periodic_crps}, highlighting the dependence of the optimal $l_{gp}$ on the material microstructure. 

%\begin{figure}[htp]
%\centering
%\subfigure[$\omega=1$, t=100.0, $l_{gp}=40$.]{\includegraphics[width=.48\columnwidth]{Figures/dis_wppf_lam05_om1_t100.jpg}}
%\subfigure[$\omega=1$, t=100.0, $l_{gp}=0.15625$.]{\includegraphics[width=.48\columnwidth]{Figures/dis_wppf_lam128_om1_t100.jpg}}
%\subfigure[$\omega=2$, t=100.0, $l_{gp}=40$.]{\includegraphics[width=.48\columnwidth]{Figures/dis_wppf_lam05_om2_t100.jpg}}
%\subfigure[$\omega=2$, t=100.0, $l_{gp}=0.15625$.]{\includegraphics[width=.48\columnwidth]{Figures/dis_wppf_lam128_om2_t100.jpg}}
%\subfigure[$\omega=3$, t=100.0, $l_{gp}=40$.]{\includegraphics[width=.48\columnwidth]{Figures/dis_wppf_lam05_om3_t100.jpg}}
%\subfigure[$\omega=3$, t=100.0, $l_{gp}=0.15625$.]{\includegraphics[width=.48\columnwidth]{Figures/dis_wppf_lam128_om3_t100.jpg}}
%\subfigure[$\omega=4$, t=100.0, $l_{gp}=40$.]{\includegraphics[width=.48\columnwidth]{Figures/dis_wppf_lam05_om4_t100.jpg}}
%\subfigure[$\omega=4$, t=100.0, $l_{gp}=0.15625$.]{\includegraphics[width=.48\columnwidth]{Figures/dis_wppf_lam128_om4_t100.jpg}}
%\caption{Validation on wave packet for a bar with disordered microstructure at the last time step $t=100.0$. The columns correspond to different correlation length $l_{gp}$ and the rows correspond to different frequencies $\omega$. }
%\label{fig:dis_wppf}
%\end{figure}

\subsubsection{Comparison with the Baseline}
Comparing the CRPS for the posterior PDF on the learnt ENOR model predictions with $l_{gp}=40$, to that from the baseline model (BNOR)~\cite{fan2023bayesian}, at the same time instant $t=2.0$, we find again superior performance from the present construction. Specifically, the ENOR CRPS is 0.008 while the BNOR PFP and PP CRPS values are 0.041 and 0.025 respectively.  Thus, consistent with the findings in the periodic material, the BNOR CRPS values are much higher than the ENOR CRPS, again indicating that the embedded model error construction provides a better statistical fit for the DNS data. 

\begin{figure}[htp]
\centering
\subfigure[ENOR PP (or PFP) for sample 10 from type 1 data at $t$=2.0.]{\includegraphics[width=.48\columnwidth]{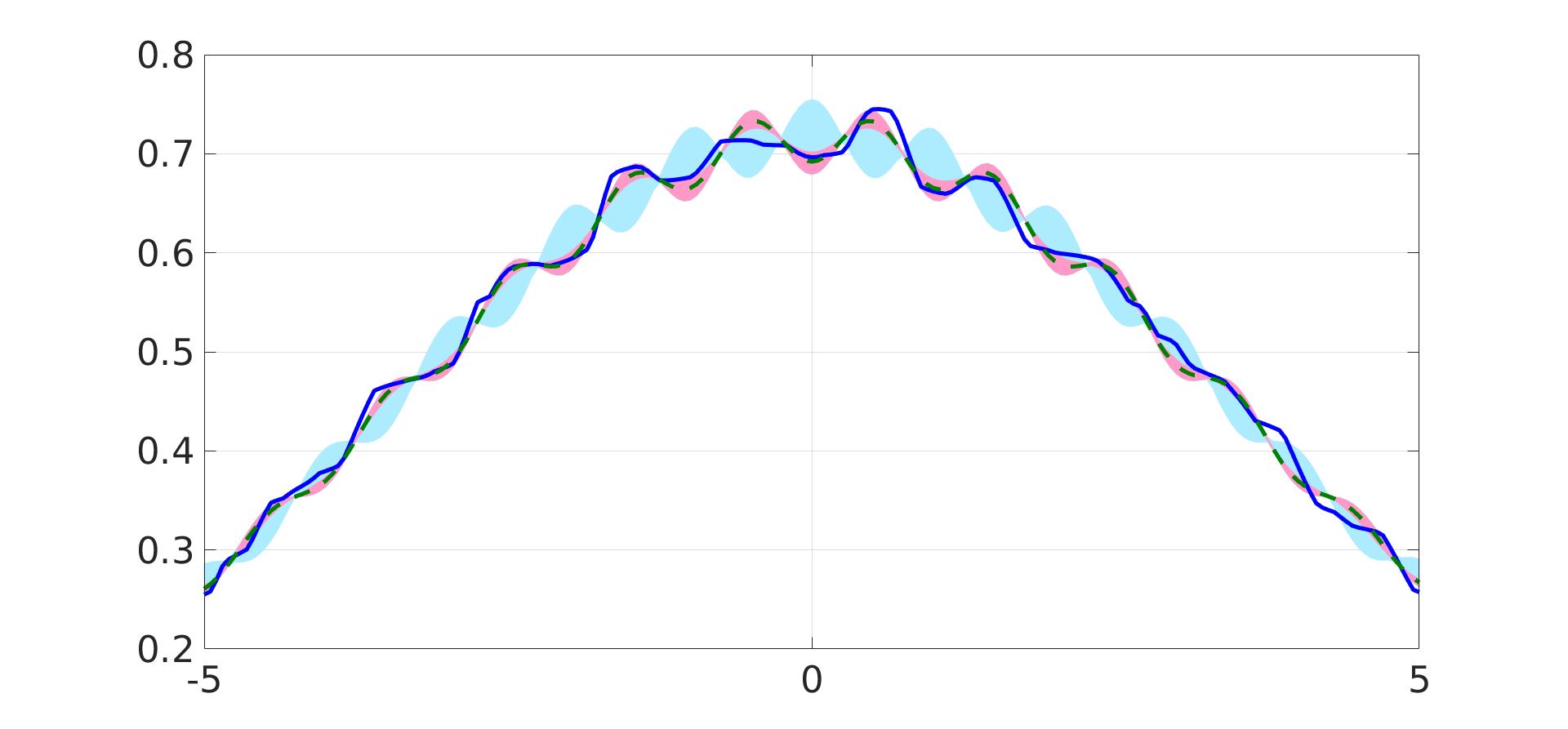}}
\hspace{0.5em}
\subfigure[ENOR PP (or PFP) for sample 3 from type 2 data at $t$=2.0.]{\includegraphics[width=.48\columnwidth]{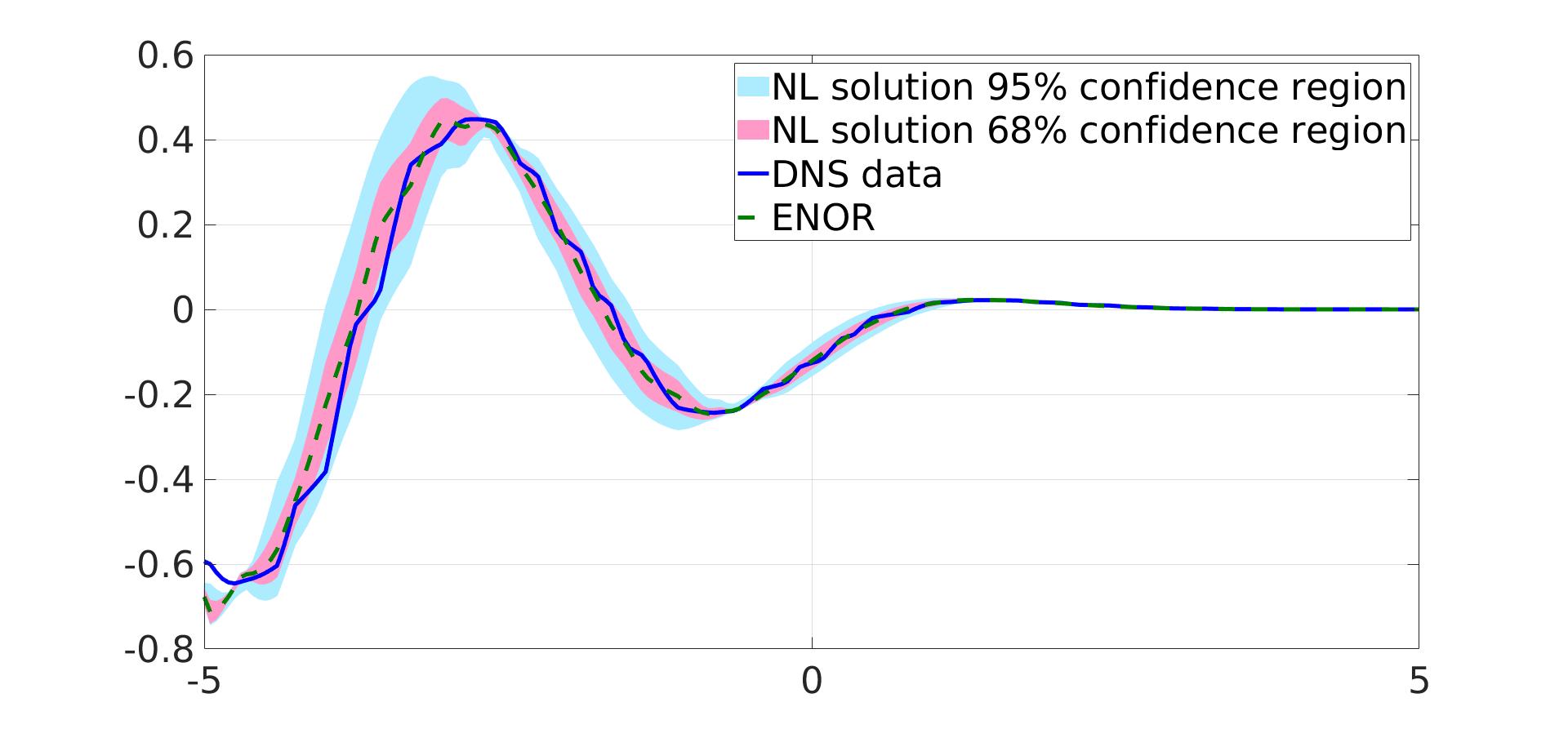}}
\subfigure[BNOR PFP for sample 10 from type 1 data at $t$=2.0.]{\includegraphics[width=.48\columnwidth]{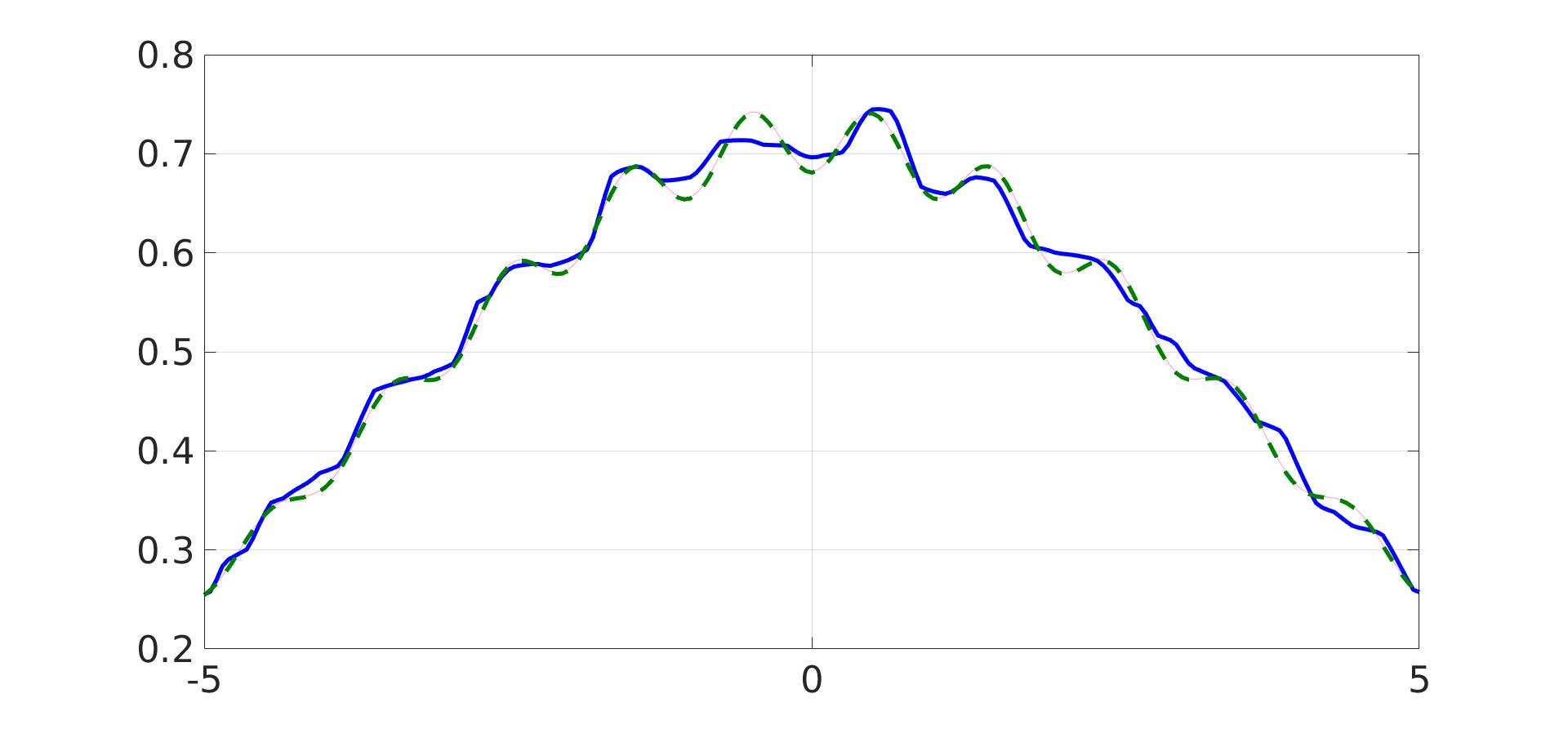}}
\hspace{0.5em}
\subfigure[BNOR PFP for sample 3 from type 2 data at $t$=2.0.]{\includegraphics[width=.48\columnwidth]{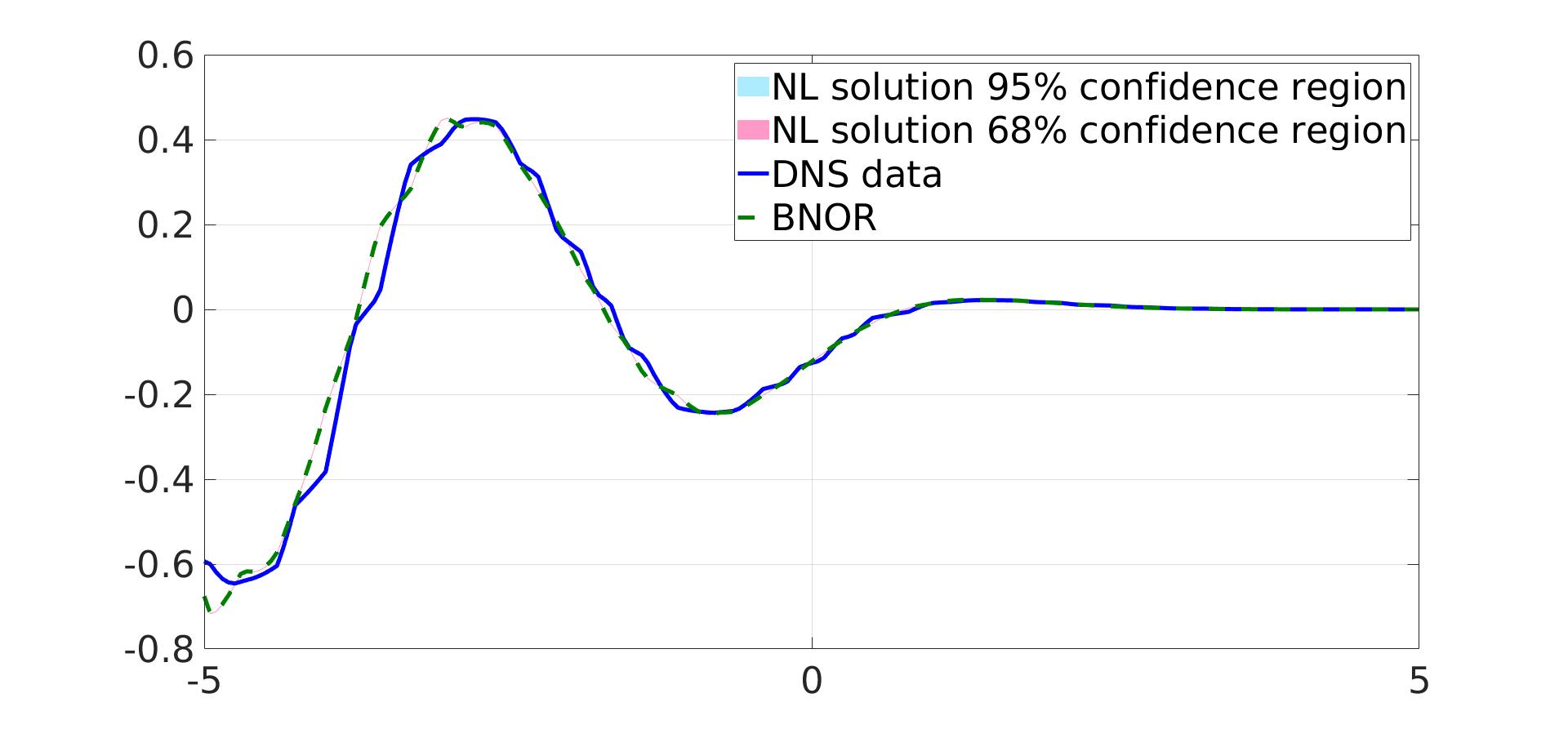}}
\subfigure[BNOR PP for sample 10 from type 1 data at $t$=2.0.]{\includegraphics[width=.48\columnwidth]{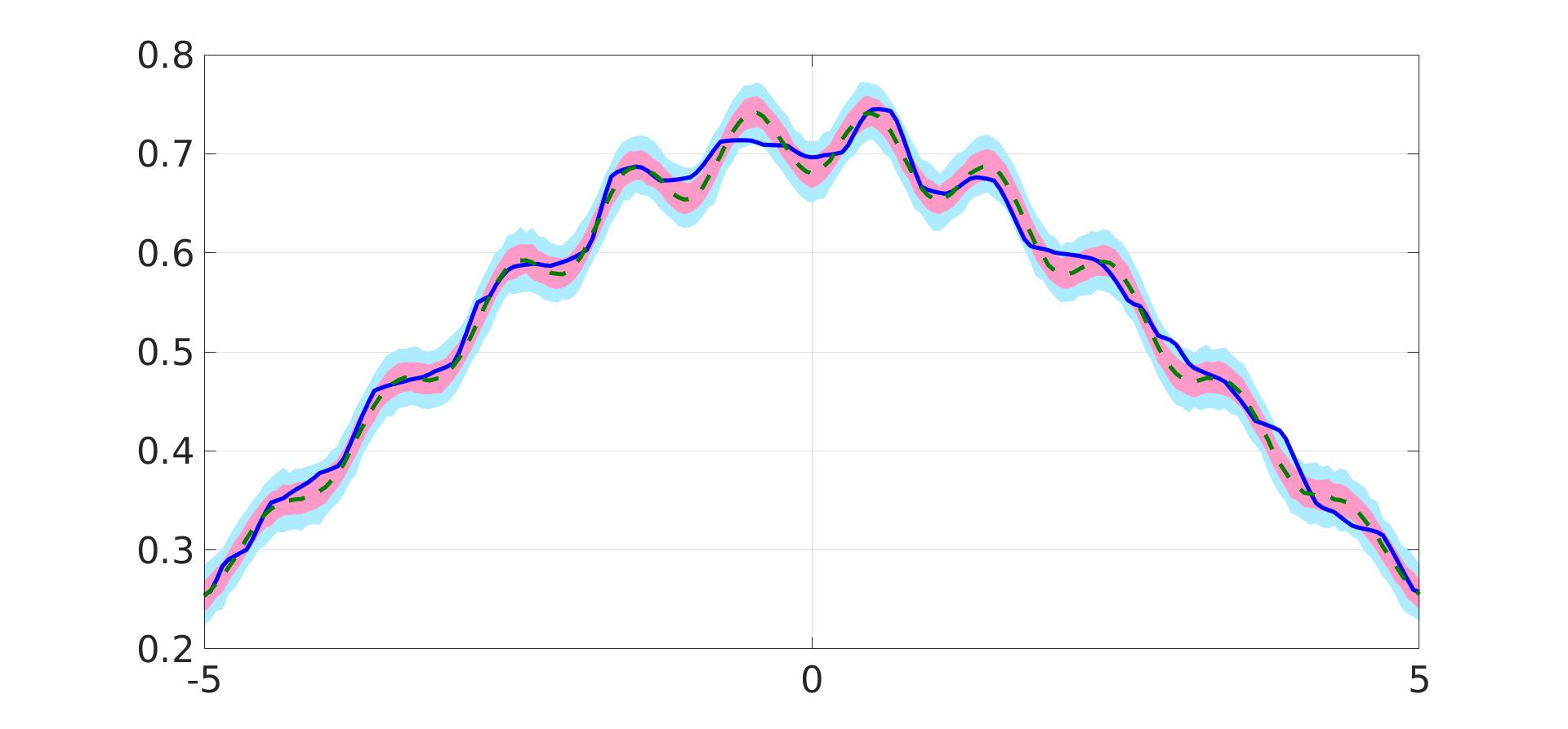}}
\hspace{0.5em}
\subfigure[BNOR PP for sample 3 from type 2 data at $t$=2.0.]{\includegraphics[width=.48\columnwidth]{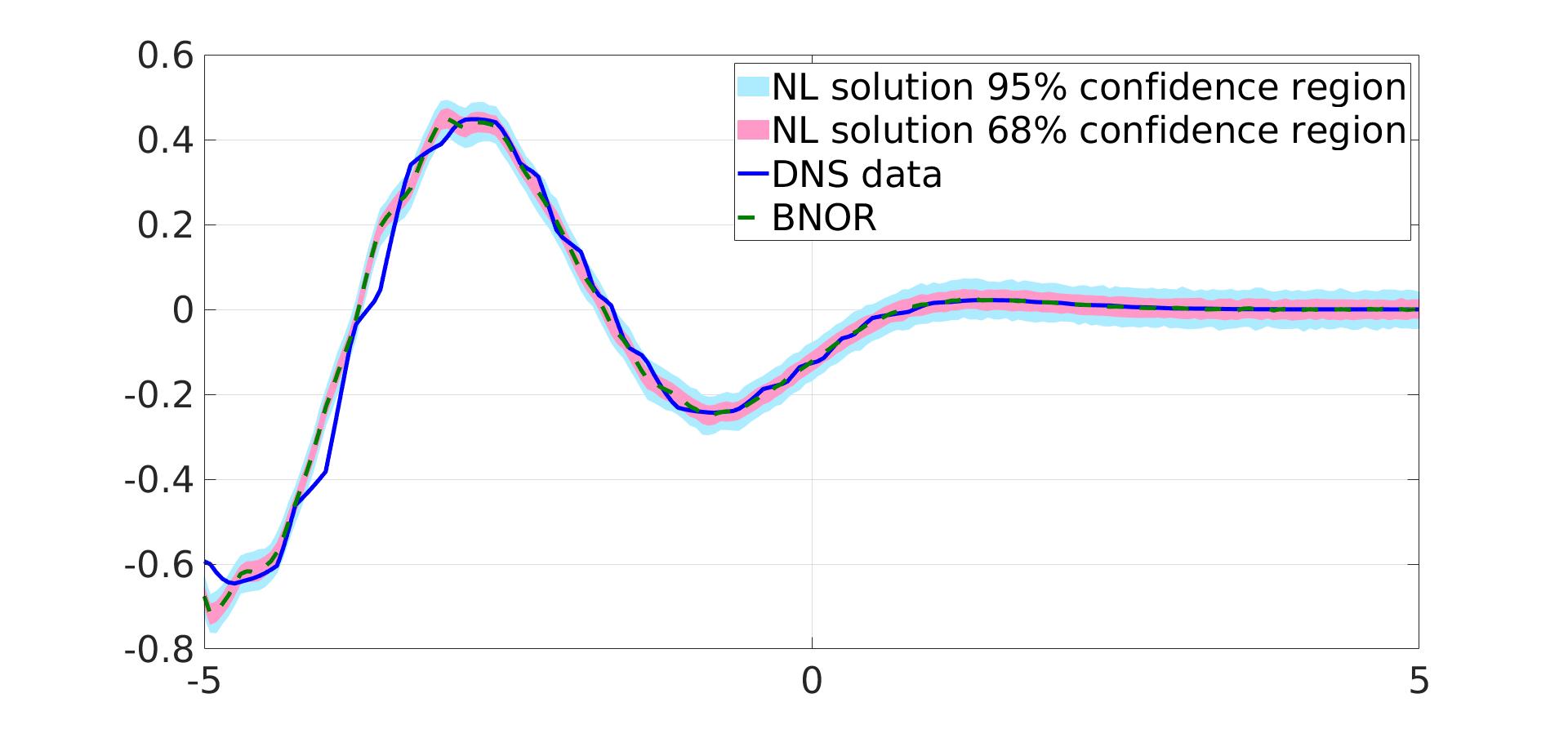}}
\caption{Comparison of posterior uncertainty (PFP and PP) between ENOR and BNOR for the disordered material.  The columns correspond to different samples in training data and the rows correspond to different methods.}
\label{fig:dis_pfppp}
\end{figure}

Further, in Fig.\ref{fig:dis_pfppp} we compare the posterior predictive density from ENOR and the BNOR PFP/PP densities for two samples from two wave types at $t=2$. While ENOR again provides a better prediction for solution uncertainty, the $95\%$ confidence region is significantly larger, especially in Fig.\ref{fig:dis_pfppp}(a). Note that according to Table.\ref{table:crps_tpf} and Fig.\ref{fig:disorder_crps}, $l_{gp}=0.3125$ is best for sample 10 in wave type 1, while $l_{gp}=2.5$ is best for sample 3 in wave type 2. Thus, the correlation length $l_{gp}=40$ employed here is not ideal for both samples. 
%,  where both should be smaller than $l_{gp}$=40. This explains why the confidence region in Fig.\ref{fig:dis_pfppp} (a) and (b) are not visually good in some specific regions. 
This again suggests that different correlation length should ideally be employed for different loading scenarios. When comparing the level of solution uncertainty from periodic (see Fig.\ref{fig:pfppp} (a) and (b)) and disordered materials (Fig.\ref{fig:dis_pfppp} (a) and (b)), a larger confidence region is observed in the later case. In fact, the average standard deviation over the bar in sample 10 of data setting 1 is $0.0037$ in the periodic microstructure case and $0.0122$ in the disordered case. Similarly, the average standard deviations in sample 3 are $0.0232$ and $0.0262$ for the periodic and disordered microstructure cases, respectively. These results highlight the increase in uncertainty when randomness is introduced in the microstructure.

\subsubsection{Parametric Uncertainty versus Model Error}

\begin{figure}[htp]
\centering
\subfigure[Sample 10 from type 1 data at $t=2.0$. Confidence region using 100 ess and 100 realization each.]{\includegraphics[width=.48\columnwidth]{./dis_tpf_s10_lam05.jpg}}
\hspace{0.5em}
\subfigure[Sample 3 from type 2 data at $t=2.0$. Confidence region using 100 effective samples and 100 realization each.]{\includegraphics[width=.48\columnwidth]{./dis_tpf_s23_lam05_jun25.jpg}}
\subfigure[Sample 10 from type 1 data at $t=2.0$. Confidence region using fixed value of kernel parameters $\Cb$ and $\sigma_{gp}$ with 1000 realization of GP.]{\includegraphics[width=.48\columnwidth]{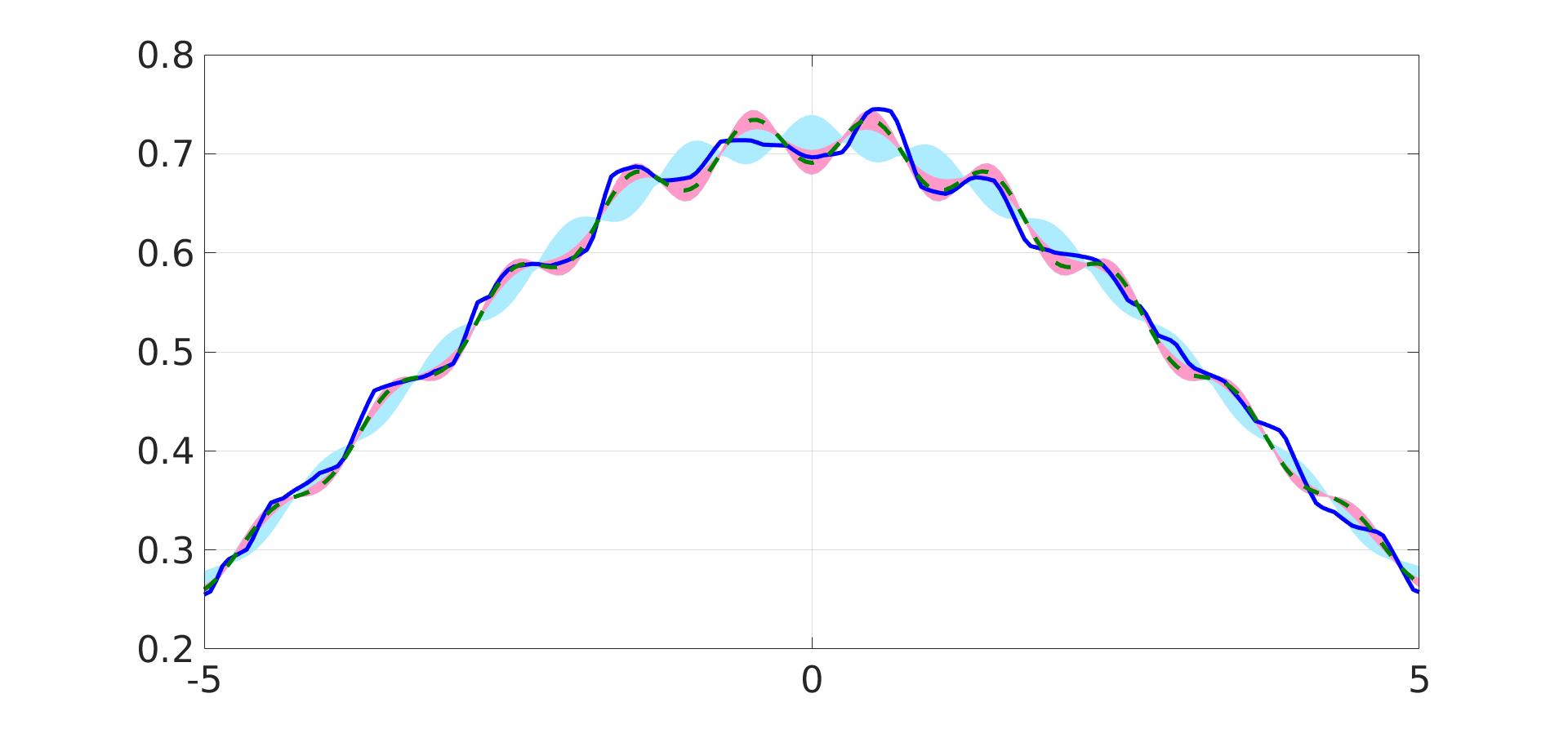}}
\hspace{0.5em}
\subfigure[Sample 3 from type 2 data at $t=2.0$. Confidence region using fixed value of kernel parameters $\Cb$ and $\sigma_{gp}$ with 1000 realization of GP.]{\includegraphics[width=.48\columnwidth]{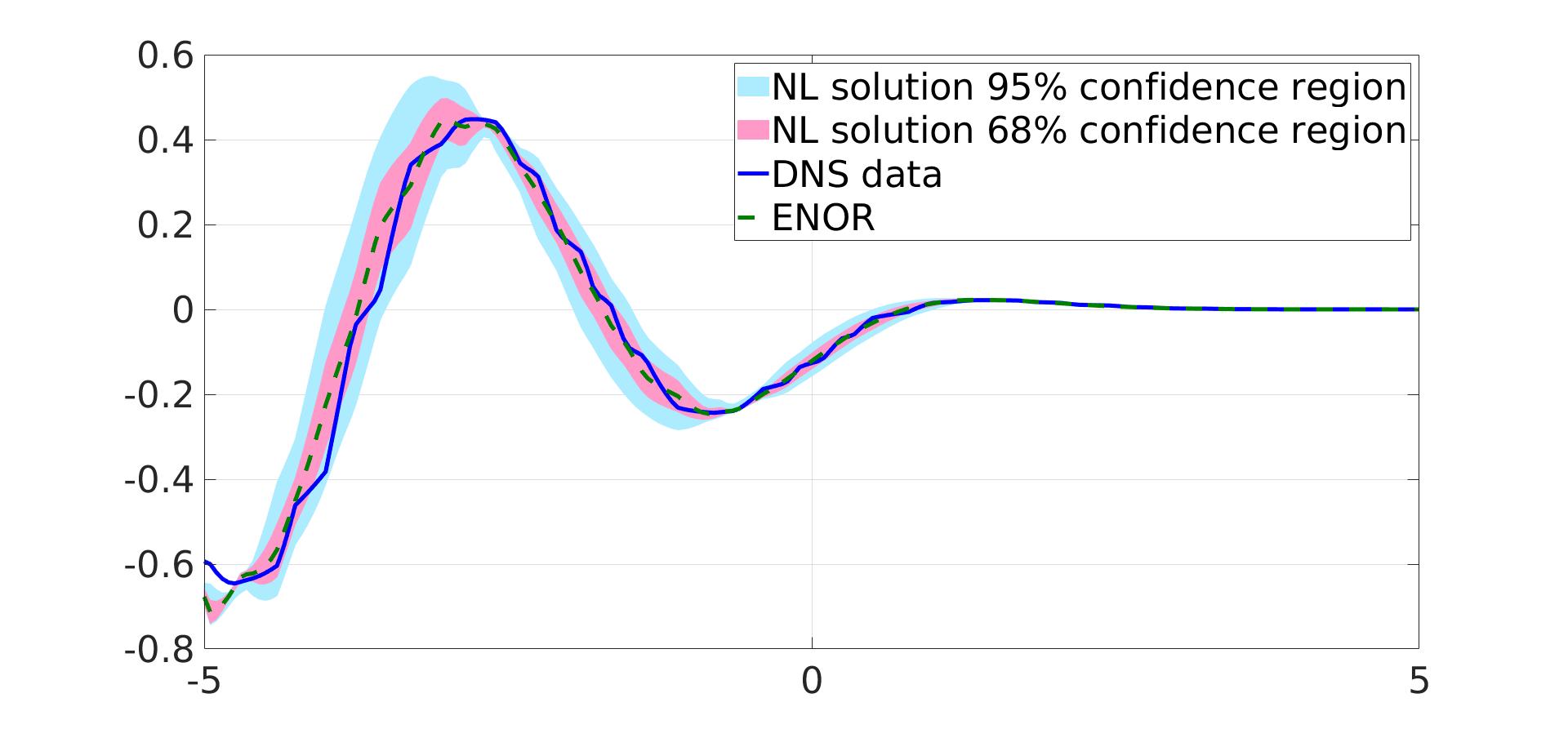}}
\subfigure[Sample 10 from type 1 data at $t=2.0$. Confidence region using 100 effective samples of kernel parameters $\Cb$ and $\sigma_{gp}$ without GP (i.e. set $\xi_m$=0).]{\includegraphics[width=.48\columnwidth]{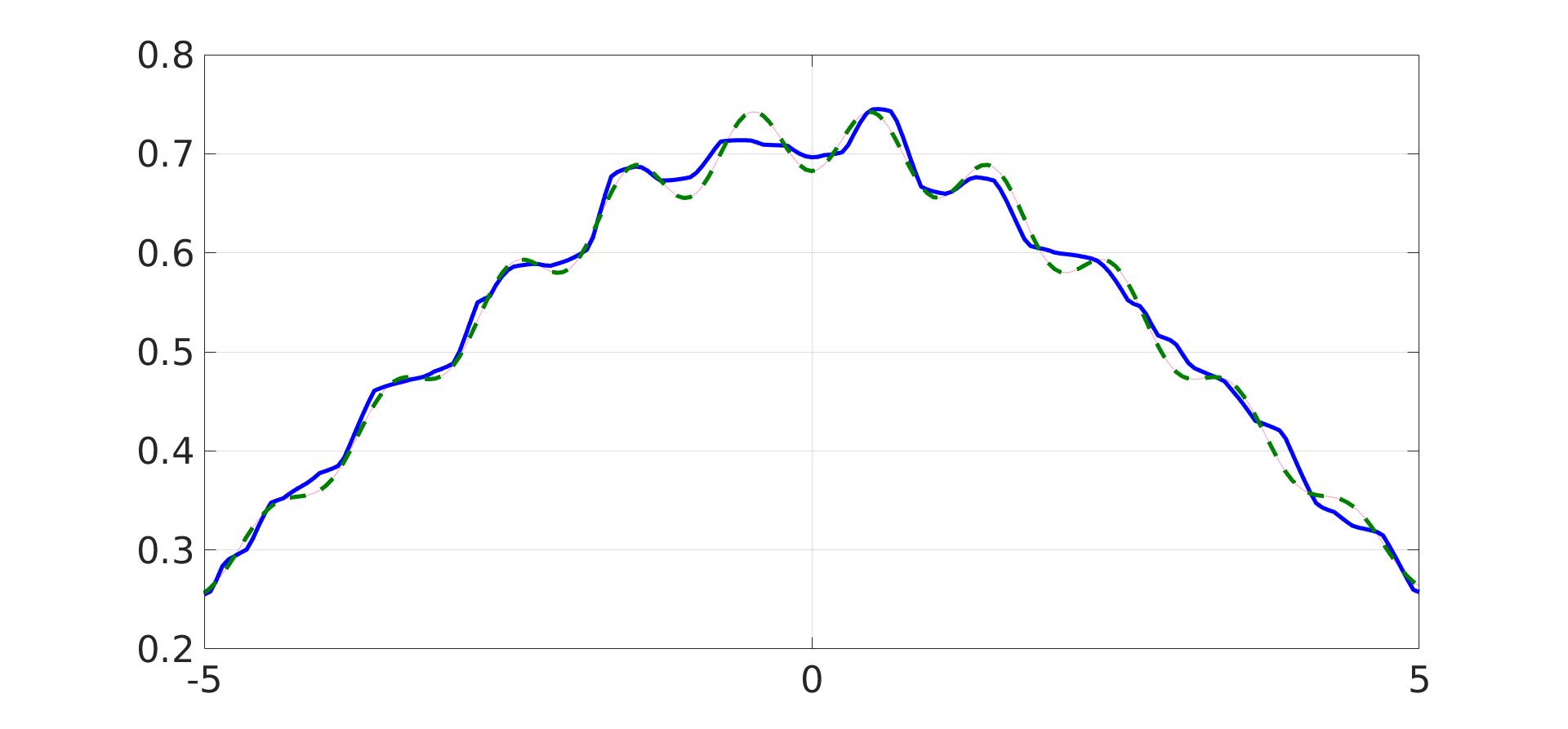}}
\hspace{0.5em}
\subfigure[Sample 3 from type 2 data at $t=2.0$. Confidence region using 100 effective samples of kernel parameters $\Cb$ and $\sigma_{gp}$ without GP (i.e. set $\xi_m$=0).]{\includegraphics[width=.48\columnwidth]{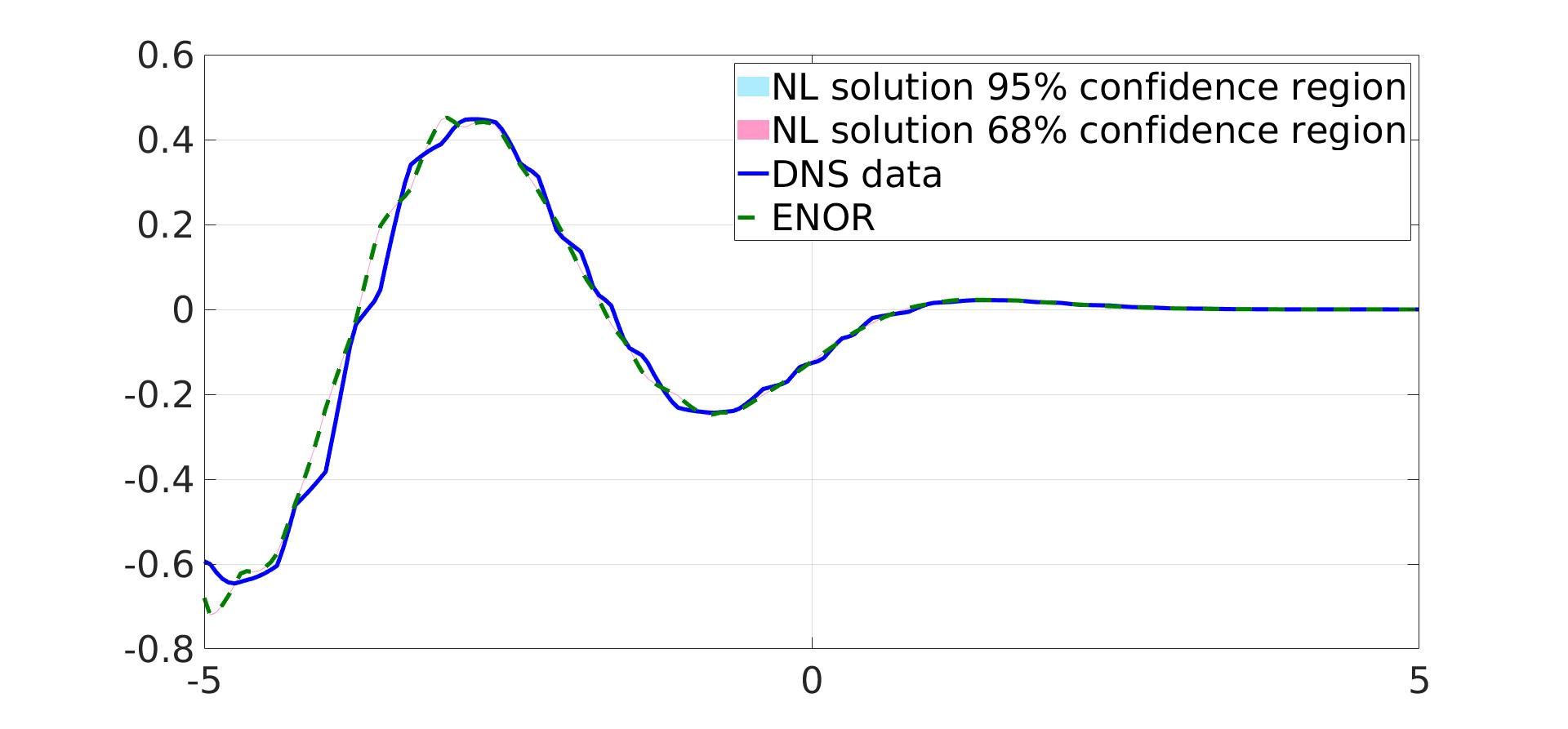}}
\caption{An illustration of the relative impact of parametric uncertainty and model error on resulting predictive uncertainty, in the disordered material. The columns correspond to different samples in training data and the rows correspond to posterior prediction using different sources of uncertainty.}
\label{fig:dis_gsa}
\end{figure}

With the purpose of investigating the impact of different source of uncertainty, we examine the predictive uncertainty resulting from parametric uncertainty versus model error for the disordered material, following the same steps in section \ref{sec:puvme}. As shown in Fig.\ref{fig:dis_gsa}, the embedded model error again dominates the posterior uncertainty, with a negligible role for parametric uncertainty.

\section{Conclusion}\label{sec:conclusion}

In this work, we proposed ENOR, a novel Bayesian embedded model error framework, to learn the optimal nonlocal surrogate for multiscale homogenization while also characterizing the impact of model error on predictive uncertainty. When learning the bottom-up nonlocal surrogate from microscale simulations, we find that the best fit surrogate has unavoidable discrepancy from the microscale model, rendering a model error UQ study necessary. While the prior work \cite{fan2023bayesian} focused on an additive \emph{iid}  error construction, we here introduce, for the first time, an embedded model error representation in nonlocal operator learning, capturing the impact of model error on predictive uncertainty. 
%This paper is an extension of \cite{fan2023bayesian}, which also aims to provide a nonlocal surrogate with uncertainty prediction from high-fidelity data or experimental measurements. Since the NOR method \cite{you2021md} has been successfully developed for unseen prediction tasks and the advantages of Bayesian inference, the NOR model and additive Gaussian noise are combined in \cite{fan2023bayesian} to achieve the goal. However, this error assumption is relatively simple, which renders the model only capture the average model discrepancy spanning the physical domain and gives visually mismatching uncertainty in some specific region.
%In order to capturing the variation of the model discrepancy, we employ embedded model error \cite{sargsyan2019embedded} to model, which is famous for its discrepancy-adaptive prediction, separation of errors’ contribution, non-shrinking uncertainty and etc. 
The algorithm is developed by adding a Gaussian process to the nonlocal kernel, such that the kernel parameters and the Gaussian process parameters can be inferred simultaneously. To solve the Bayesian inference problem, a two phase algorithm, Alg.\ref{algorithm:1}, and a multilevel delayed acceptance Markov chain Monte Carlo (MLDA-MCMC) method are proposed, to provide efficient sampling and fast converging chains. The effectiveness of ENOR is demonstrated on the stress wave propagation problem in heterogeneous bars. Comparing to the prior work \cite{fan2023bayesian}, ENOR improves the accuracy in (1) capturing the correct group velocity; (2) producing high-fidelity training data and predicting the substantially different wave type with accurate confidence region structure; (3) posterior sampling of the model parameters; and (4) selection for the correlation length for different tasks. 

%We focused on the improvement on the confidence region in this work, and it successfully generate the uncertainty that span the discrepancy in the data. However, the best choice of correlation length remains an open question. 
From both visual inspection and quantitative tests, it was observed that the optimal choice of the GP correlation length may differ for different frequencies and wave types. As a natural extension, we plan to investigate frequency-dependent models. We also plan to explore uncertainty quantification for more complex data-driven homogenization models, such as peridynamics models \cite{you2021md,you2023towards} in 2D or 3D, and nonlinear models based on neural networks~\cite{jafarzadeh2024peridynamic,jafarzadeh2024heterogeneous,you2022learning}. Since the number of trainable parameters increases substantially in these models, efficient Bayesian inference techniques and reduced order error models would be desired.
%Another interesting approach is treating the correlation length $l_{gp}$ as a hyperparameter through the training process, which might allow for more information on $l_{gp}$.
%When computational resource permitted, another interesting direction is treating  the correlation length $l_{gp}$ as a parameter that could be inferred simultaneously with the current set of parameters, which allows for more information on $l_{gp}$.

%% The Acknowledgements part is started with the command \acknowledgements;
%% acknowledgements are then done as normal sections before appendix
%% \acknowledgements

\section*{Acknowledgements}

YF and YY acknowledge support by the National Science Foundation under award DMS-2436624, the National Institute of Health under award 1R01GM157589-01, and the AFOSR grant FA9550-22-1-0197. Portions of this research were conducted on Lehigh University's Research Computing infrastructure partially supported by NSF Award 2019035. HNN acknowledges the support of the U.S. Department of Energy, Office of Science, Office of Advanced Scientific Computing Research (ASCR), Scientific Discovery through Advanced Computing (SciDAC) Program through the FASTMath Institute. Sandia National Laboratories is a multi-mission laboratory managed and operated by National Technology \& Engineering Solutions of Sandia, LLC (NTESS), a wholly owned subsidiary of Honeywell International Inc., for the U.S. Department of Energy’s National Nuclear Security Administration (DOE/NNSA) under contract DE-NA0003525. This written work is authored by an employee of NTESS. The employee, not NTESS, owns the right, title and interest in and to the written work and is responsible for its contents. Any subjective views or opinions that might be expressed in the written work do not necessarily represent the views of the U.S. Government. The publisher acknowledges that the U.S. Government retains a non-exclusive, paid-up, irrevocable, world-wide license to publish or reproduce the published form of this written work or allow others to do so, for U.S. Government purposes. The DOE will provide public access to results of federally sponsored research in accordance with the DOE Public Access Plan.

%% appendix sections are then done as normal sections and after Acknowledgements

\section{Appendix}

\subsection{Additional Results}

\begin{figure}[htp]
\centering
\subfigure[$\omega=1$, t=100.0, $l_{gp}=40$.]{\includegraphics[width=.48\columnwidth]{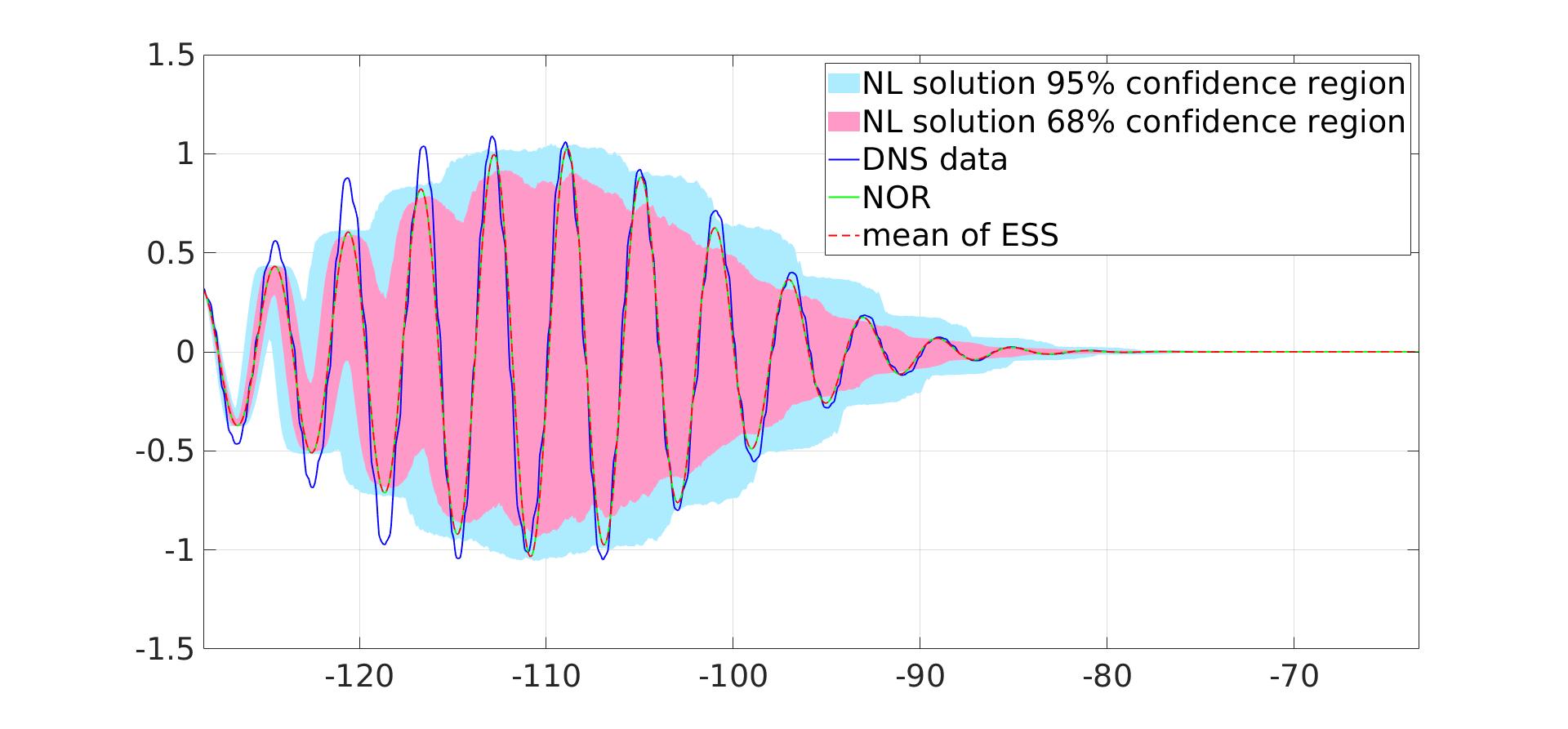}}
\subfigure[$\omega=1$, t=100.0, $l_{gp}=0.15625$.]{\includegraphics[width=.48\columnwidth]{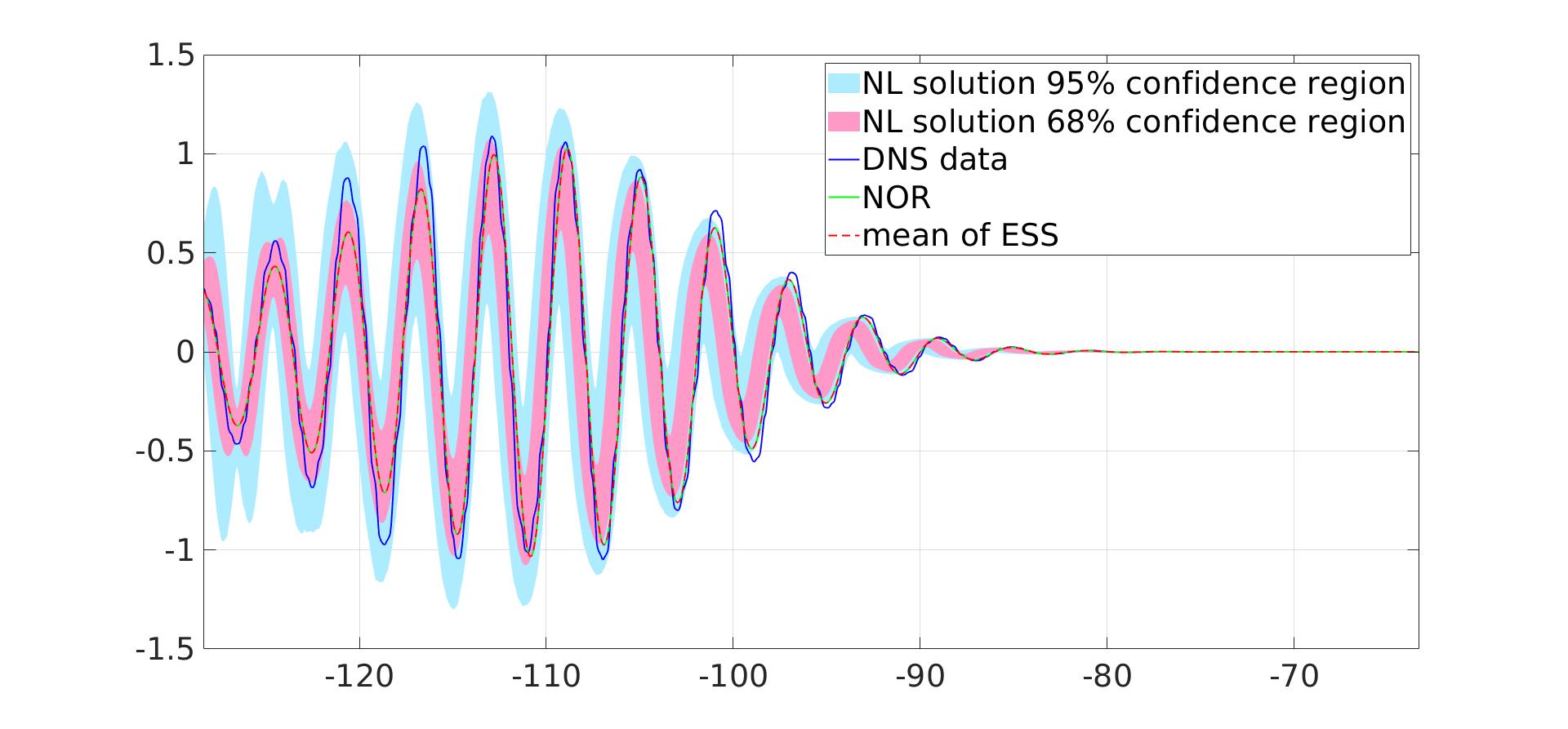}}
\subfigure[$\omega=2$, t=100.0, $l_{gp}=40$.]{\includegraphics[width=.48\columnwidth]{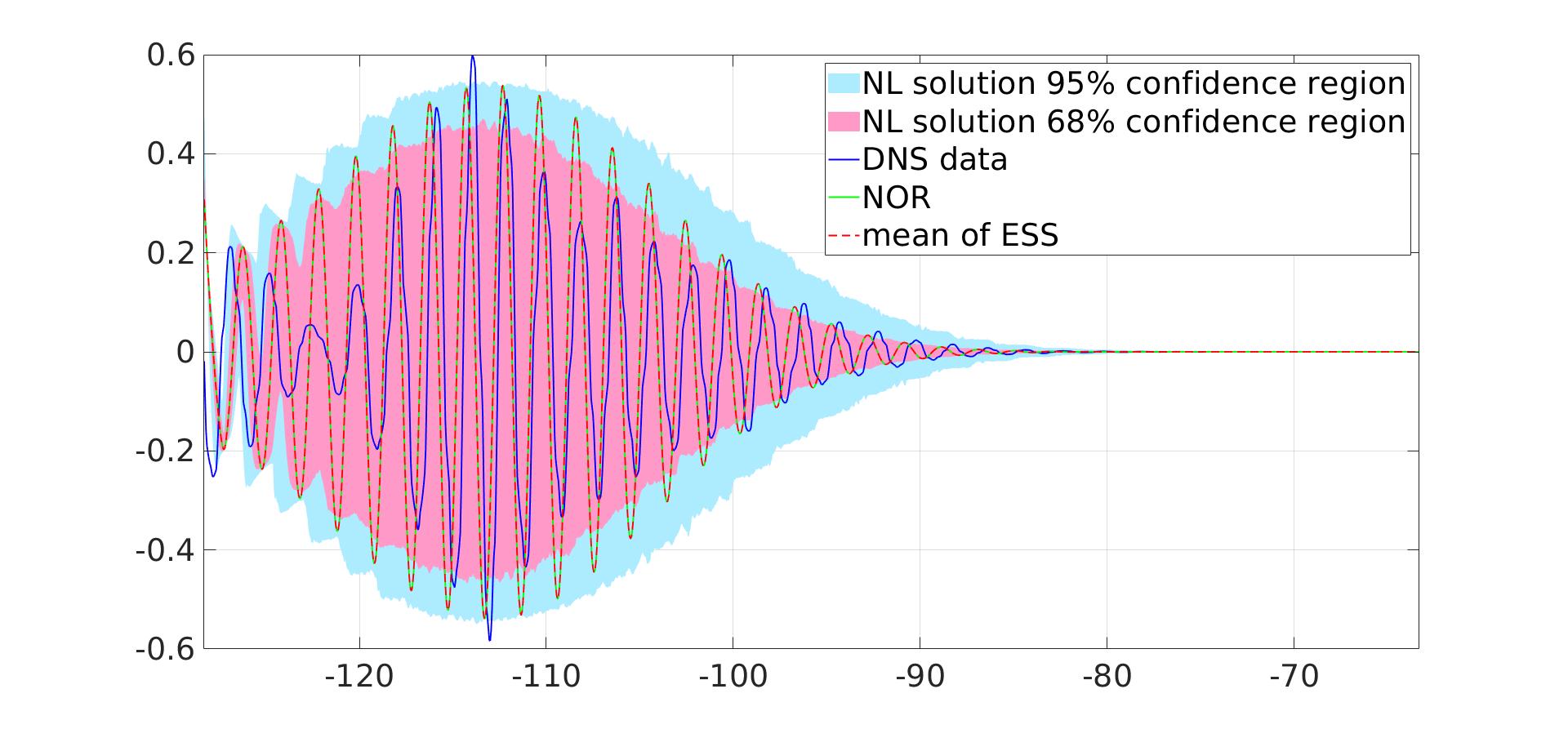}}
\subfigure[$\omega=2$, t=100.0, $l_{gp}=0.15625$.]{\includegraphics[width=.48\columnwidth]{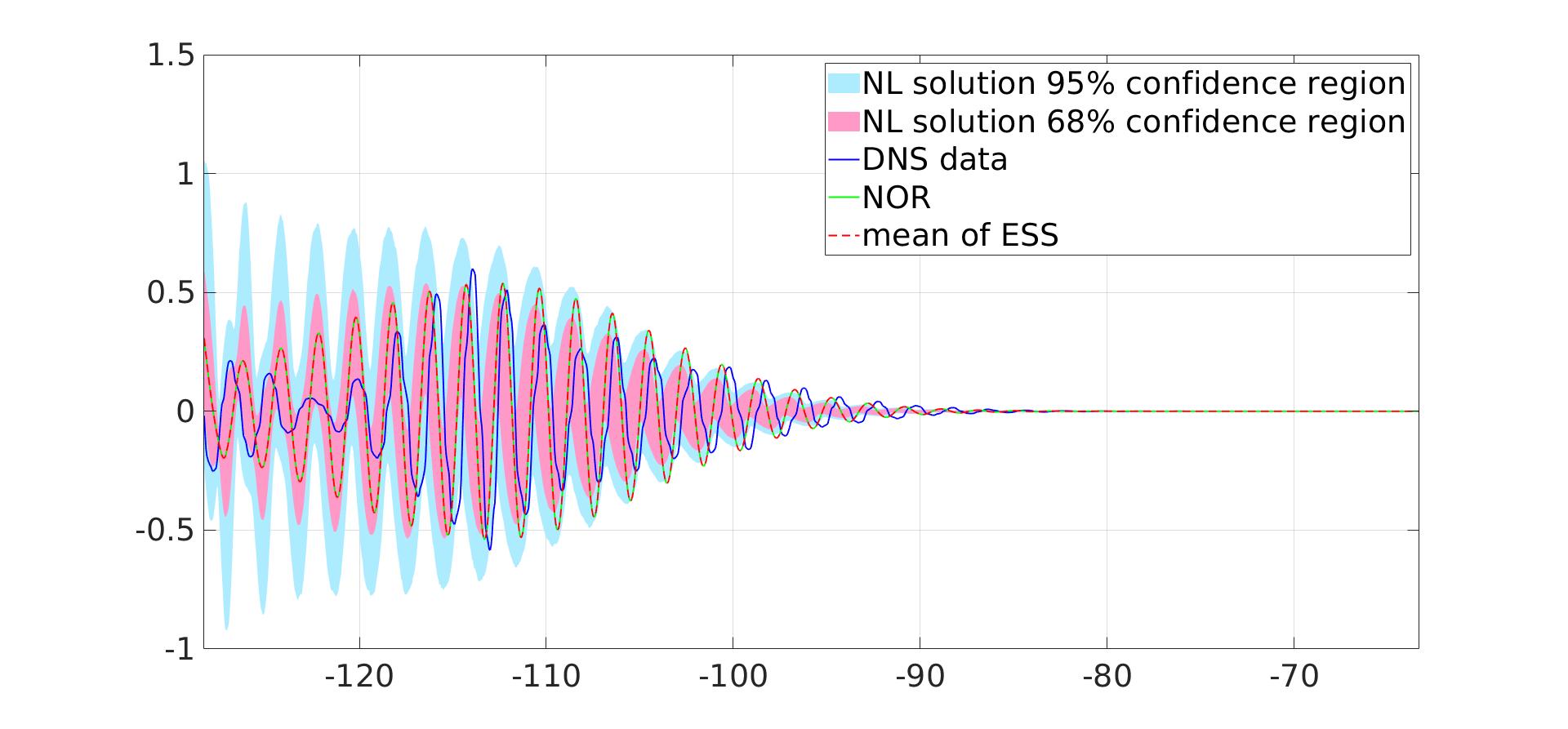}}
\subfigure[$\omega=3$, t=100.0, $l_{gp}=40$.]{\includegraphics[width=.48\columnwidth]{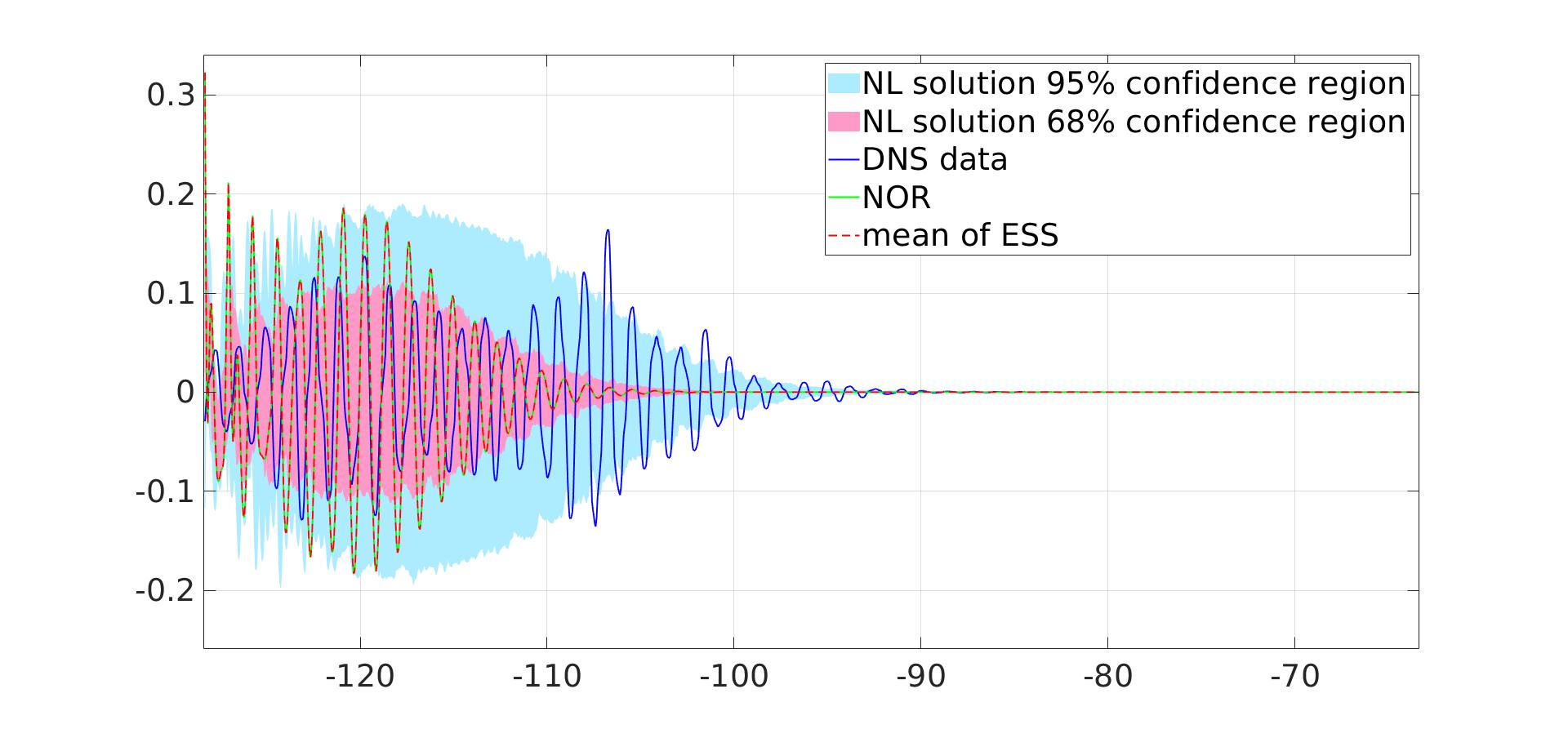}}
\subfigure[$\omega=3$, t=100.0, $l_{gp}=0.15625$.]{\includegraphics[width=.48\columnwidth]{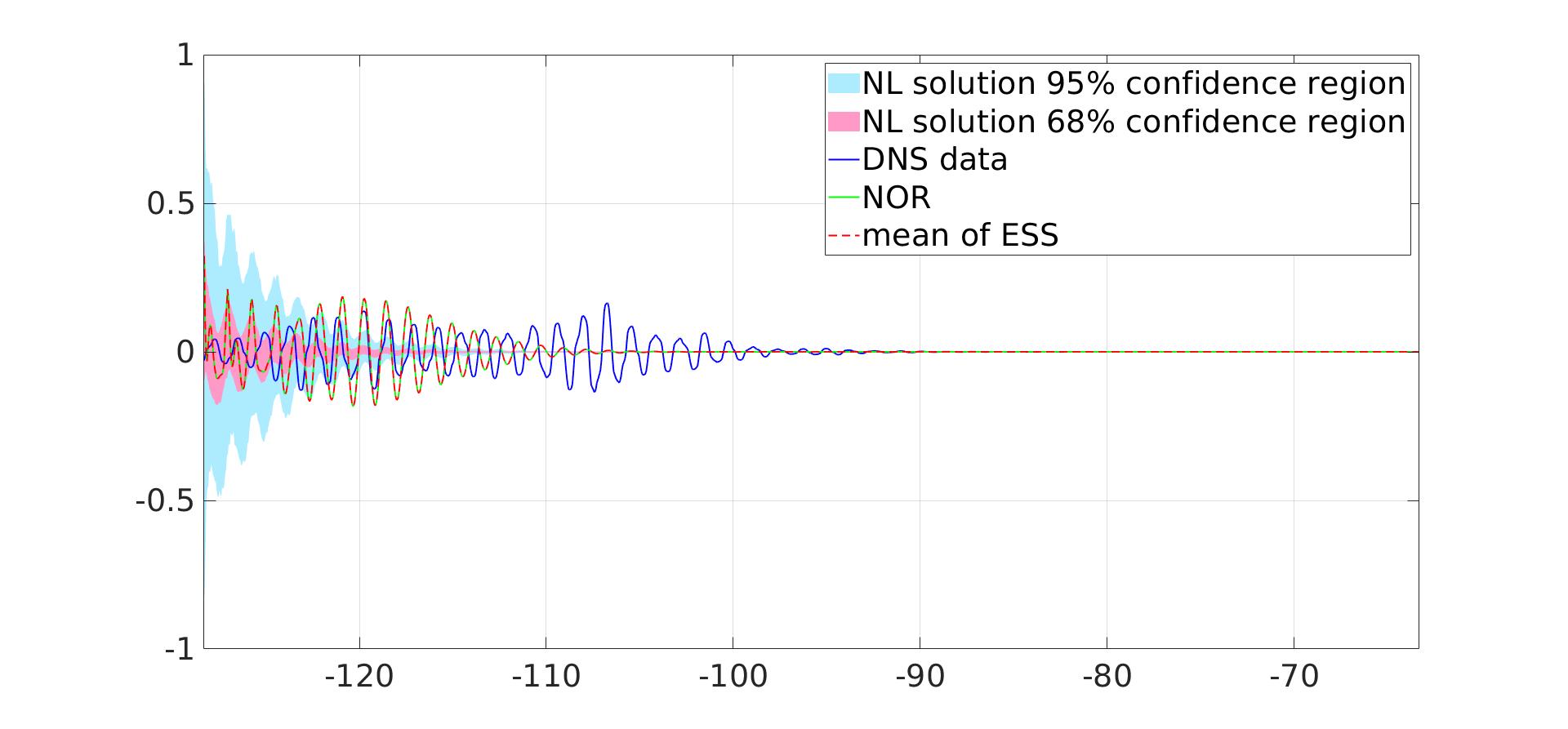}}
\subfigure[$\omega=4$, t=100.0, $l_{gp}=40$.]{\includegraphics[width=.48\columnwidth]{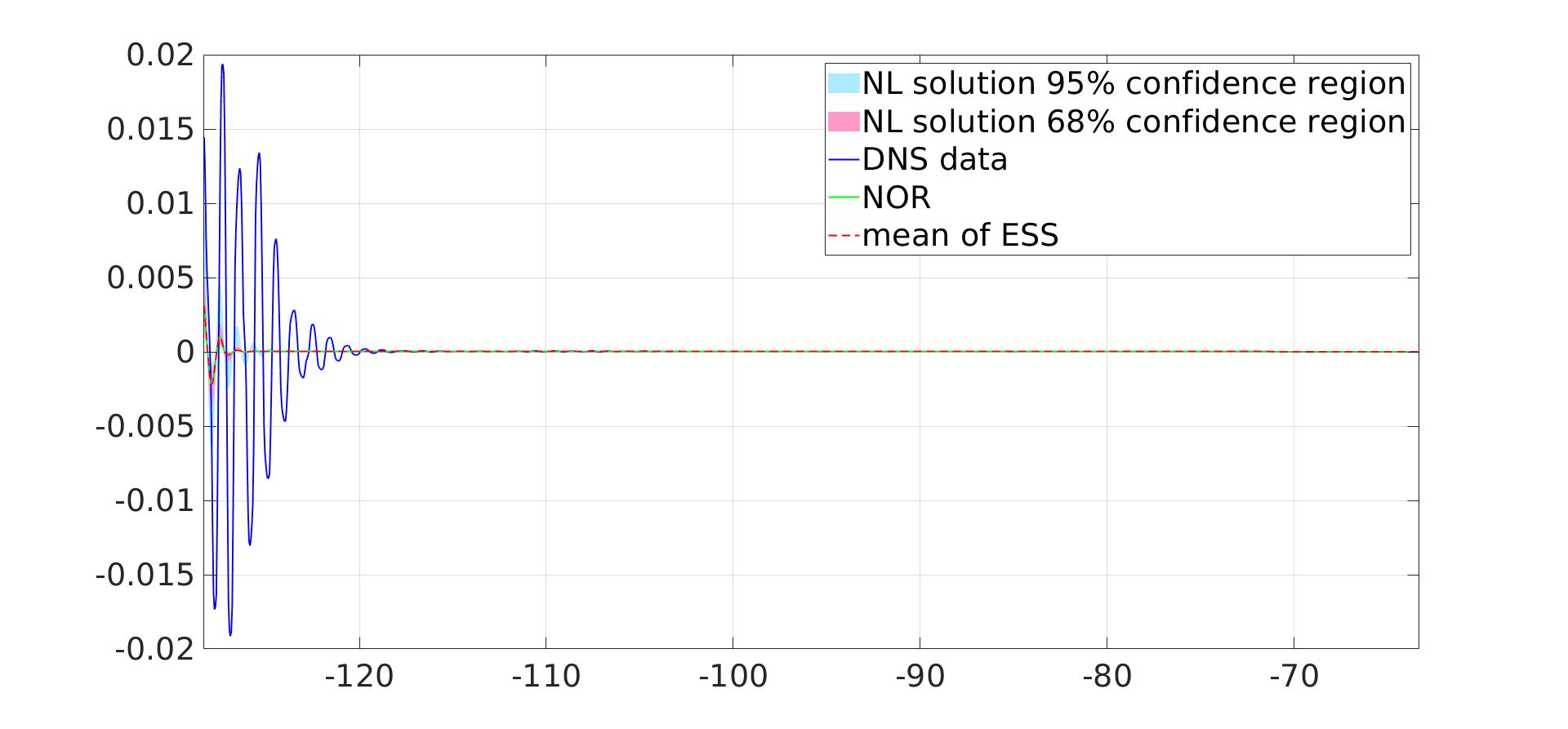}}
\subfigure[$\omega=4$, t=100.0, $l_{gp}=0.15625$.]{\includegraphics[width=.48\columnwidth]{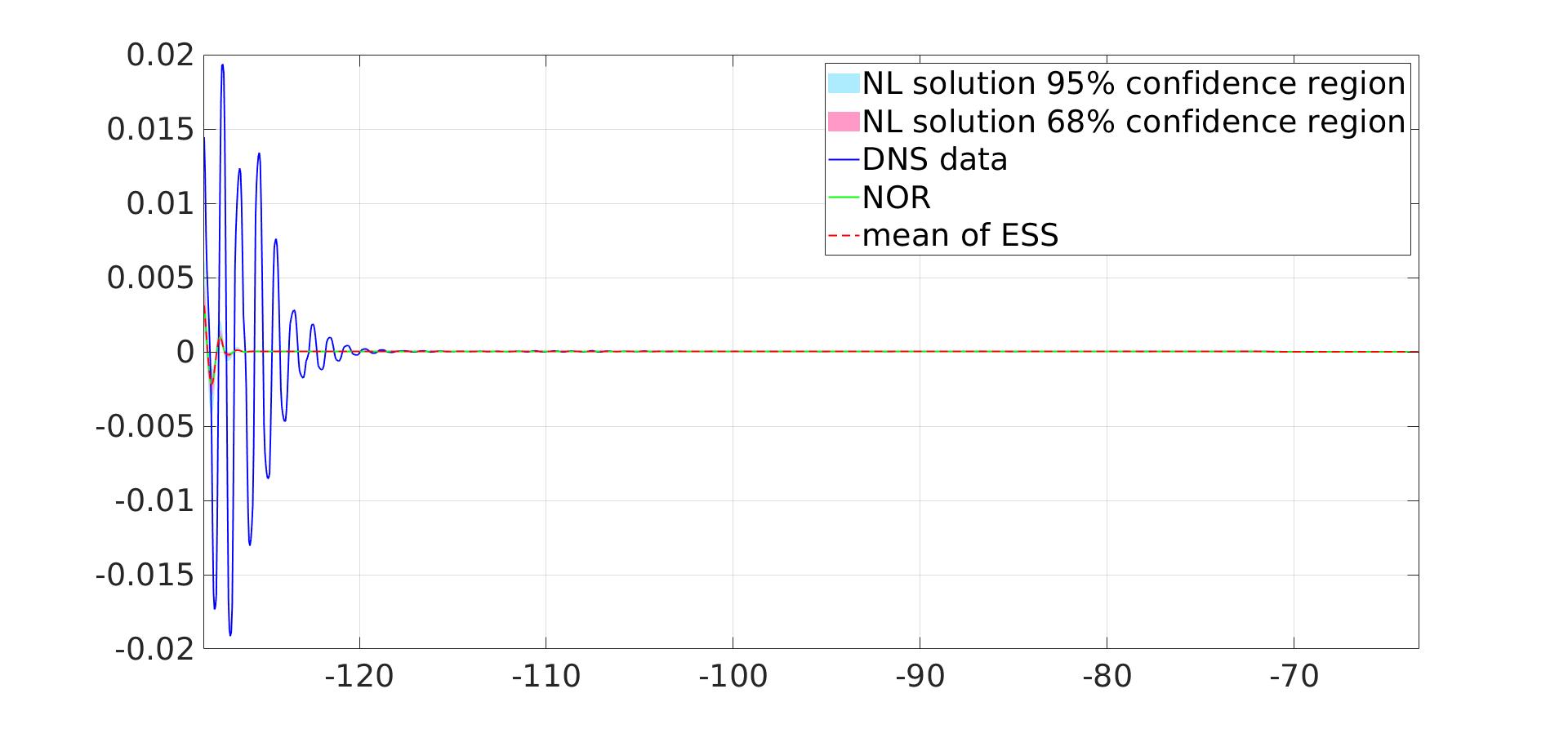}}
\caption{Validation on wave packet for a bar with disordered microstructure at the last time step $t=100.0$. The columns correspond to different correlation length $l_{gp}$ and the rows correspond to different frequencies $\omega$. }
\label{fig:dis_wppf}
\end{figure}

In Figure \ref{fig:dis_wppf}, we provide the validation results using $l_{gp}=40$ and $l_{gp}=0.15625$ on wave packet for the random microstructure material, at the last time step $t=100.0$. At low frequencies ($\omega=1,2$), $l_{gp}=0.15625$ works better in reflecting the local magnitude of the solution discrepancy. Near the band stop ($\omega=3$), $l_{gp}=40$ works better since the confidence region from the small $l_{gp}$ case fails to cover the DNS data. For the large frequency case ($\omega=4$), both ENOR models have successfully predicted that the stress wave should stop propagating. These observations are consistent with the results from the periodic microstructure material.

\begin{thebibliography}{10}

\bibitem{zohdi2017homogenization}
Tarek~I Zohdi.
\newblock Homogenization methods and multiscale modeling.
\newblock {\em Encyclopedia of Computational Mechanics Second Edition}, pages
  1--24, 2017.

\bibitem{bensoussan2011asymptotic}
Alain Bensoussan, Jacques-Louis Lions, and George Papanicolaou.
\newblock {\em Asymptotic analysis for periodic structures}, volume 374.
\newblock American Mathematical Soc., 2011.

\bibitem{weinan2003multiscale}
E~Weinan and Bjorn Engquist.
\newblock Multiscale modeling and computation.
\newblock {\em Notices of the AMS}, 50(9):1062--1070, 2003.

\bibitem{efendiev2013generalized}
Yalchin Efendiev, Juan Galvis, and Thomas~Y Hou.
\newblock Generalized multiscale finite element methods (gmsfem).
\newblock {\em Journal of computational physics}, 251:116--135, 2013.

\bibitem{junghans2008transport}
Christoph Junghans, Matej Praprotnik, and Kurt Kremer.
\newblock Transport properties controlled by a thermostat: An extended
  dissipative particle dynamics thermostat.
\newblock {\em Soft Matter}, 4(1):156--161, 2008.

\bibitem{kubo1966fluctuation}
Rep Kubo.
\newblock The fluctuation-dissipation theorem.
\newblock {\em Reports on progress in physics}, 29(1):255, 1966.

\bibitem{santosa1991dispersive}
Fadil Santosa and William~W Symes.
\newblock A dispersive effective medium for wave propagation in periodic
  composites.
\newblock {\em SIAM Journal on Applied Mathematics}, 51(4):984--1005, 1991.

\bibitem{dobson2010sharp}
Matthew Dobson, Mitchell Luskin, and Christoph Ortner.
\newblock Sharp stability estimates for the force-based quasicontinuum
  approximation of homogeneous tensile deformation.
\newblock {\em Multiscale Modeling \& Simulation}, 8(3):782--802, 2010.

\bibitem{ortiz1987method}
M~Ortiz.
\newblock A method of homogenization of elastic media.
\newblock {\em International journal of engineering science}, 25(7):923--934,
  1987.

\bibitem{moes1999simplified}
Nicolas Mo{\"e}s, J~Tinsley Oden, Kumar Vemaganti, and Jean-Fran{\c{c}}ois
  Remacle.
\newblock Simplified methods and a posteriori error estimation for the
  homogenization of representative volume elements (rve).
\newblock {\em Computer methods in applied mechanics and engineering},
  176(1-4):265--278, 1999.

\bibitem{hughes2004energy}
Thomas~JR Hughes, Garth~N Wells, and Alan~A Wray.
\newblock Energy transfers and spectral eddy viscosity in large-eddy
  simulations of homogeneous isotropic turbulence: Comparison of dynamic
  smagorinsky and multiscale models over a range of discretizations.
\newblock {\em Physics of Fluids}, 16(11):4044--4052, 2004.

\bibitem{milton2022theory}
Graeme~W Milton.
\newblock {\em The theory of composites}.
\newblock SIAM, 2022.

\bibitem{chapman2021homogenization}
S~Jonathan Chapman and Zachary~M Wilmott.
\newblock Homogenization of flow through periodic networks.
\newblock {\em SIAM Journal on Applied Mathematics}, 81(3):1034--1051, 2021.

\bibitem{espanol1995statistical}
Pep Espanol and Patrick Warren.
\newblock Statistical mechanics of dissipative particle dynamics.
\newblock {\em Europhysics letters}, 30(4):191, 1995.

\bibitem{groot1997dissipative}
Robert~D Groot and Patrick~B Warren.
\newblock Dissipative particle dynamics: Bridging the gap between atomistic and
  mesoscopic simulation.
\newblock {\em The Journal of chemical physics}, 107(11):4423--4435, 1997.

\bibitem{hoogerbrugge1992simulating}
PJ~Hoogerbrugge and JMVA Koelman.
\newblock Simulating microscopic hydrodynamic phenomena with dissipative
  particle dynamics.
\newblock {\em Europhysics letters}, 19(3):155, 1992.

\bibitem{beran70}
MJ~Beran and JJ~McCoy.
\newblock Mean field variations in a statistical sample of heterogeneous
  linearly elastic solids.
\newblock {\em International Journal of Solids and Structures},
  6(8):1035--1054, 1970.

\bibitem{cher06}
Kirill Cherednichenko, Valery~P Smyshlyaev, and VV~Zhikov.
\newblock Non-local homogenised limits for composite media with highly
  anisotropic periodic fibres.
\newblock {\em Proceedings of the Royal Society of Edinburgh Section A
  Mathematics}, 136(1):87--114, 2006.

\bibitem{karal64}
Frank~C Karal~Jr and Joseph~B Keller.
\newblock Elastic, electromagnetic, and other waves in a random medium.
\newblock {\em Journal of Mathematical Physics}, 5(4):537--547, 1964.

\bibitem{rahali15}
Y~Rahali, I~Giorgio, JF~Ganghoffer, and Francesco dell'Isola.
\newblock Homogenization {\`a} la piola produces second gradient continuum
  models for linear pantographic lattices.
\newblock {\em International Journal of Engineering Science}, 97:148--172,
  2015.

\bibitem{silling2021propagation}
Stewart~A Silling.
\newblock Propagation of a stress pulse in a heterogeneous elastic bar.
\newblock {\em Journal of Peridynamics and Nonlocal Modeling}, 3(3):255--275,
  2021.

\bibitem{you2021md}
Huaiqian You, Yue Yu, Stewart Silling, and Marta D’Elia.
\newblock A data-driven peridynamic continuum model for upscaling molecular
  dynamics.
\newblock {\em Computer Methods in Applied Mechanics and Engineering},
  389:114400, 2022.

\bibitem{you2023towards}
HQ~You, Xiao Xu, Yue Yu, Stewart Silling, Marta D’Elia, and John Foster.
\newblock Towards a unified nonlocal, peridynamics framework for the
  coarse-graining of molecular dynamics data with fractures.
\newblock {\em Applied Mathematics and Mechanics}, 44(7):1125--1150, 2023.

\bibitem{you2024nonlocal}
Huaiqian You, Yue Yu, Stewart Silling, and Marta D’Elia.
\newblock Nonlocal operator learning for homogenized models: From high-fidelity
  simulations to constitutive laws.
\newblock {\em Journal of Peridynamics and Nonlocal Modeling}, pages 1--16,
  2024.

\bibitem{zhang2022metanor}
Lu~Zhang, Huaiqian You, and Yue Yu.
\newblock {MetaNOR}: A meta-learnt nonlocal operator regression approach for
  metamaterial modeling.
\newblock {\em MRS Communications}, 2022.

\bibitem{tadmor2014critical}
Eitan Tadmor and Changhui Tan.
\newblock Critical thresholds in flocking hydrodynamics with non-local
  alignment.
\newblock {\em Philosophical Transactions of the Royal Society A: Mathematical,
  Physical and Engineering Sciences}, 372(2028):20130401, 2014.

\bibitem{du2020multiscale}
Qiang Du, Bjorn Engquist, and Xiaochuan Tian.
\newblock Multiscale modeling, homogenization and nonlocal effects:
  Mathematical and computational issues.
\newblock {\em Contemporary mathematics}, 754, 2020.

\bibitem{silling2000reformulation}
Stewart~A Silling.
\newblock Reformulation of elasticity theory for discontinuities and long-range
  forces.
\newblock {\em Journal of the Mechanics and Physics of Solids}, 48(1):175--209,
  2000.

\bibitem{gilboa2009nonlocal}
Guy Gilboa and Stanley Osher.
\newblock Nonlocal operators with applications to image processing.
\newblock {\em Multiscale Modeling \& Simulation}, 7(3):1005--1028, 2009.

\bibitem{Scalas2000}
E.~Scalas, R.~Gorenflo, and F.~Mainardi.
\newblock Fractional calculus and continuous time finance.
\newblock {\em Physica A}, 284:376--384, 2000.

\bibitem{DElia2017}
M.~D'Elia, Q.~Du, M.~Gunzburger, and R.~Lehoucq.
\newblock Nonlocal convection-diffusion problems on bounded domains and
  finite-range jump processes.
\newblock {\em Computational Methods in Applied Mathematics}, 29:71--103, 2017.

\bibitem{Meerschaert2012}
M.~Meerschaert and A.~Sikorskii.
\newblock {\em Stochastic models for fractional calculus}.
\newblock Studies in mathematics, Gruyter, 2012.

\bibitem{silling2024peridynamic}
Stewart~A Silling, Siavash Jafarzadeh, and Yue Yu.
\newblock Peridynamic models for random media found by coarse graining.
\newblock {\em Journal of Peridynamics and Nonlocal Modeling}, pages 1--30,
  2024.

\bibitem{jafarzadeh2024heterogeneous}
Siavash Jafarzadeh, Stewart Silling, Lu~Zhang, Colton Ross, Chung-Hao Lee,
  SM~Rahman, Shuodao Wang, and Yue Yu.
\newblock Heterogeneous peridynamic neural operators: Discover biotissue
  constitutive law and microstructure from digital image correlation
  measurements.
\newblock {\em arXiv preprint arXiv:2403.18597}, 2024.

\bibitem{smy00}
Valery~P Smyshlyaev and Kirill~D Cherednichenko.
\newblock On rigorous derivation of strain gradient effects in the overall
  behaviour of periodic heterogeneous media.
\newblock {\em Journal of the Mechanics and Physics of Solids},
  48(6-7):1325--1357, 2000.

\bibitem{willis85}
John~R Willis.
\newblock The nonlocal influence of density variations in a composite.
\newblock {\em International Journal of Solids and Structures}, 21(7):805--817,
  1985.

\bibitem{eringen1972nonlocal}
A.~C. Eringen and D.~G.~B. Edelen.
\newblock On nonlocal elasticity.
\newblock {\em International Journal of Engineering Science}, 10(3):233--248,
  1972.

\bibitem{bobaru2016handbook}
Florin Bobaru, John~T Foster, Philippe~H Geubelle, and Stewart~A Silling.
\newblock {\em Handbook of peridynamic modeling}.
\newblock CRC press, 2016.

\bibitem{You2021}
Huaiqian You, Yue Yu, Stewart Silling, and Marta D'Elia.
\newblock Data-driven learning of nonlocal models: from high-fidelity
  simulations to constitutive laws.
\newblock {\em AAAI Spring Symposium: MLPS}, 2021.

\bibitem{you2020data}
Huaiqian You, Yue Yu, Nathaniel Trask, Mamikon Gulian, and Marta D’Elia.
\newblock Data-driven learning of nonlocal physics from high-fidelity synthetic
  data.
\newblock {\em Computer Methods in Applied Mechanics and Engineering},
  374:113553, 2021.

\bibitem{xu2021machine}
Xiao Xu, Marta D’Elia, and John~T Foster.
\newblock A machine-learning framework for peridynamic material models with
  physical constraints.
\newblock {\em Computer Methods in Applied Mechanics and Engineering},
  386:114062, 2021.

\bibitem{jafarzadeh2024peridynamic}
Siavash Jafarzadeh, Stewart Silling, Ning Liu, Zhongqiang Zhang, and Yue Yu.
\newblock Peridynamic neural operators: A data-driven nonlocal constitutive
  model for complex material responses.
\newblock {\em arXiv preprint arXiv:2401.06070}, 2024.

\bibitem{de2023machine}
Eduardo A~Barros de~Moraes, Marta D’Elia, and Mohsen Zayernouri.
\newblock Machine learning of nonlocal micro-structural defect evolutions in
  crystalline materials.
\newblock {\em Computer Methods in Applied Mechanics and Engineering},
  403:115743, 2023.

\bibitem{xu2022sub}
Xiao Xu, Marta D'Elia, Christian Glusa, and John~T Foster.
\newblock Machine-learning of nonlocal kernels for anomalous subsurface
  transport from breakthrough curves.
\newblock {\em arXiv preprint arXiv:2201.11146}, 2022.

\bibitem{yu2024nonlocal}
Yue Yu, Ning Liu, Fei Lu, Tian Gao, Siavash Jafarzadeh, and Stewart Silling.
\newblock Nonlocal attention operator: Materializing hidden knowledge towards
  interpretable physics discovery.
\newblock {\em arXiv preprint arXiv:2408.07307}, 2024.

\bibitem{lu2022nonparametric}
Fei Lu, Qingci An, and Yue Yu.
\newblock Nonparametric learning of kernels in nonlocal operators.
\newblock {\em arXiv preprint arXiv:2205.11006}, 2022.

\bibitem{fan2023bayesian}
Yiming Fan, Marta D’Elia, Yue Yu, Habib~N Najm, and Stewart Silling.
\newblock Bayesian nonlocal operator regression: A data-driven learning
  framework of nonlocal models with uncertainty quantification.
\newblock {\em Journal of Engineering Mechanics}, 149(8):04023049, 2023.

\bibitem{Laplace:1814}
M.~de Laplace.
\newblock {\em Essai philosophique sur les probabilit\'es}.
\newblock Gauthier-Villars, Paris, 1814.
\newblock English Ed.: "A Philosophical Essay on Probabilities", Dover Pub.,
  6$^{th}$ ed., 1995.

\bibitem{Casella:1990}
George Casella and Roger~L Berger.
\newblock {\em Statistical inference}, volume~70.
\newblock Duxbury Press Belmont, CA, 1990.

\bibitem{Jaynes:2003}
E.T. Jaynes.
\newblock {\em {Probability Theory: The Logic of Science, {\rm G.L. Bretthorst,
  Ed}}}.
\newblock Cambridge University Press, Cambridge, UK, 2003.

\bibitem{Robert:2004}
C.P. Robert and G.~Casella.
\newblock {\em {Monte Carlo Statistical Methods}}.
\newblock {Springer Texts in Statistics}. Springer, 2004.

\bibitem{Sivia:2006}
D.~S. Sivia and J.~Skilling.
\newblock {\em Data Analysis: A Bayesian Tutorial, Second Edition}.
\newblock Oxford University Press, 2006.

\bibitem{Carlin:2011}
B.~P. Carlin and T.~A. Louis.
\newblock {\em Bayesian Methods for Data Analysis}.
\newblock Chapman and Hall/CRC, Boca Raton, FL, 2011.

\bibitem{Kennedy:2001}
M.~C. Kennedy and A.~O'Hagan.
\newblock Bayesian calibration of computer models.
\newblock {\em Journal of the Royal Statistical Society: Series {B}},
  63(3):425--464, 2001.

\bibitem{Higdon:2003}
D.~Higdon, H.~Lee, and C.~Holloman.
\newblock Markov chain {M}onte {C}arlo-based approaches for inference in
  computationally intensive inverse problems.
\newblock {\em Bayesian Statistics}, 7:181--197, 2003.

\bibitem{Oliver:2011}
T.~A. Oliver and R.~D. Moser.
\newblock Bayesian uncertainty quantification applied to {RANS} turbulence
  models.
\newblock {\em J. Phys.: Conf. Ser.}, 318, 2011.

\bibitem{Cui:2016b}
Tiangang Cui, Youssef Marzouk, and Karen Willcox.
\newblock Scalable posterior approximations for large-scale {Bayesian} inverse
  problems via likelihood-informed parameter and state reduction.
\newblock {\em Journal of Computational Physics}, 315:363--387, 2016.

\bibitem{hakim2018probabilistic}
Layal Hakim, Guilhem Lacaze, Mohammad Khalil, Khachik Sargsyan, Habib Najm, and
  Joseph Oefelein.
\newblock Probabilistic parameter estimation in a 2-step chemical kinetics
  model for n-dodecane jet autoignition.
\newblock {\em Combustion Theory and Modelling}, 22(3):446--466, 2018.

\bibitem{Huan:2018b}
Xun Huan, Cosmin Safta, Khachik Sargsyan, Gianluca Geraci, Michael~S. Eldred,
  Zachary~P. Vane, Guilhem Lacaze, Joseph~C. Oefelein, and Habib~N. Najm.
\newblock {Global Sensitivity Analysis and Estimation of Model Error, toward
  Uncertainty Quantification in Scramjet Computations}.
\newblock {\em AIAA Journal}, 56(3):1170--1184, 2018.

\bibitem{bayarri2007framework}
Maria~J Bayarri, James~O Berger, Rui Paulo, Jerry Sacks, John~A Cafeo, James
  Cavendish, Chin-Hsu Lin, and Jian Tu.
\newblock A framework for validation of computer models.
\newblock {\em Technometrics}, 49(2):138--154, 2007.

\bibitem{kleijnen2009kriging}
Jack~PC Kleijnen.
\newblock Kriging metamodeling in simulation: A review.
\newblock {\em European journal of operational research}, 192(3):707--716,
  2009.

\bibitem{sargsyan2015statistical}
K~Sargsyan, HN~Najm, and R~Ghanem.
\newblock On the statistical calibration of physical models.
\newblock {\em International Journal of Chemical Kinetics}, 47(4):246--276,
  2015.

\bibitem{sargsyan2019embedded}
Khachik Sargsyan, Xun Huan, and Habib~N Najm.
\newblock Embedded model error representation for bayesian model calibration.
\newblock {\em International Journal for Uncertainty Quantification}, 9(4),
  2019.

\bibitem{lykkegaard2020adaptivemlda}
Mikkel~B. Lykkegaard, Grigorios Mingas, Robert Scheichl, Colin Fox, and Tim~J.
  Dodwell.
\newblock Multilevel delayed acceptance mcmc with an adaptive error model in
  pymc3, 2020.
\newblock Workshop on machine learning for engineering modeling, simulation and
  design, NeurIPS 2020.

\bibitem{lykkegaard2023multilevel}
Mikkel~B Lykkegaard, Tim~J Dodwell, Colin Fox, Grigorios Mingas, and Robert
  Scheichl.
\newblock Multilevel delayed acceptance mcmc.
\newblock {\em SIAM/ASA Journal on Uncertainty Quantification}, 11(1):1--30,
  2023.

\bibitem{liu2023ino}
Ning Liu, Yue Yu, Huaiqian You, and Neeraj Tatikola.
\newblock Ino: Invariant neural operators for learning complex physical systems
  with momentum conservation.
\newblock In {\em International Conference on Artificial Intelligence and
  Statistics}, pages 6822--6838. PMLR, 2023.

\bibitem{liu2024harnessing}
Ning Liu, Yiming Fan, Xianyi Zeng, Milan Kl{\"o}wer, LU~ZHANG, and Yue Yu.
\newblock Harnessing the power of neural operators with automatically encoded
  conservation laws.
\newblock In {\em Forty-first International Conference on Machine Learning},
  2024.

\bibitem{du2017peridynamic}
Qiang Du, Yunzhe Tao, and Xiaochuan Tian.
\newblock A peridynamic model of fracture mechanics with bond-breaking.
\newblock {\em Journal of Elasticity}, pages 1--22, 2017.

\bibitem{Kenn:OHag:2001}
M.~C. Kennedy and A.~O'Hagan.
\newblock Bayesian calibration of computer models (with discussion).
\newblock {\em Journal of the Royal Statistical Society B}, 63:425--464, 2001.

\bibitem{Kenn:OHag:Higg:2002}
M.~C. Kennedy, A.~O'Hagan, and N.~Higgins.
\newblock Bayesian analysis of computer code outputs.
\newblock In {\em Quantitative Methods for Current Environmental Issues.
  \mbox{C. W. Anderson, V. Barnett, P. C. Chatwin, and A. H. El-Shaarawi
  (eds.)}}, pages 227--243. Springer-Verlag: London, 2002.

\bibitem{Higd:2004}
D.~Higdon, M.~C. Kennedy, J.~Cavendish, J.~Cafeo, and R.~D. Ryne.
\newblock Combining field data and computer simulations for calibration and
  prediction.
\newblock {\em SIAM Journal on Scientific Computing}, 26:448--466, 2004.

\bibitem{Bayarri:2009a}
M.~J. Bayarri, J.~O. Berger, M.~C. Kennedy, A.~Kottas, R.~Paulo, J.~Sacks,
  J.~A. Cafeo, C.~H. Lin, and J.~Tu.
\newblock Predicting vehicle crashworthiness: validation of computer models for
  functional and hierarchical data.
\newblock {\em Journal of the American Statistical Association}, 104:929--942,
  2009.

\bibitem{Hakim:2018}
Layal Hakim, Guilhem Lacaze, Mohammad Khalil, Khachik Sargsyan, Habib Najm, and
  Joseph Oefelein.
\newblock Probabilistic parameter estimation in a 2-step chemical kinetics
  model for n-dodecane jet autoignition.
\newblock {\em Combustion Theory and Modelling}, 22(3):446--466, 2018.

\bibitem{Huan:2018}
Xun Huan, Cosmin Safta, Khachik Sargsyan, Gianluca Geraci, Michael~S. Eldred,
  Zachary~P. Vane, Guilhem Lacaze, Joseph~C. Oefelein, and Habib~N. Najm.
\newblock Global sensitivity analysis and estimation of model error, toward
  uncertainty quantification in scramjet computations.
\newblock {\em AIAA Journal}, 56(3):1170--1184, 2018.

\bibitem{Hegde:2024}
Arun Hegde, Elan Weiss, Wolfgang Windl, Habib~N. Najm, and Cosmin Safta.
\newblock A bayesian calibration framework with embedded model error for model
  diagnostics.
\newblock {\em International Journal for Uncertainty Quantification},
  14(6):37--70, 2024.

\bibitem{Beaumont:2002}
M.~Beaumont, W.~Zhang, and D.~J. Balding.
\newblock Approximate {B}ayesian {C}omputation in population genetics.
\newblock {\em Genetics}, 162(4):2025--2035, 2002.

\bibitem{Marjoram:2003}
P.~Marjoram, J.~Molitor, V.~Plagnol, and S.~Tavar\'{e}.
\newblock Markov chain {M}onte {C}arlo without likelihoods.
\newblock {\em Proc Natl Acad Sci USA}, 100(26):15324--15328, 2003.

\bibitem{Sisson:2011}
S.~A. Sisson and Y.~Fan.
\newblock {\em Handbook of Markov Chain Monte Carlo}, chapter Likelihood-free
  Markov chain Monte Carlo, pages 313--338.
\newblock Chapman and Hall/CRC Press, 2011.

\bibitem{Karhunen:1946}
K.~Karhunen.
\newblock Zur spektraltheorie stochastischer prozesse.
\newblock {\em Ann. Acad. Sci. Fennicae}, 37, 1946.

\bibitem{spanos2007karhunen}
Pol~D Spanos, Michael Beer, and John Red-Horse.
\newblock Karhunen--lo{\`e}ve expansion of stochastic processes with a modified
  exponential covariance kernel.
\newblock {\em Journal of Engineering Mechanics}, 133(7):773--779, 2007.

\bibitem{lucor2004generalized}
Didier Lucor, C-H Su, and George~Em Karniadakis.
\newblock Generalized polynomial chaos and random oscillators.
\newblock {\em International Journal for Numerical Methods in Engineering},
  60(3):571--596, 2004.

\bibitem{cui2011bayesian}
Tiangang Cui, Colin Fox, and MJ~O'sullivan.
\newblock Bayesian calibration of a large-scale geothermal reservoir model by a
  new adaptive delayed acceptance metropolis hastings algorithm.
\newblock {\em Water Resources Research}, 47(10), 2011.

\bibitem{abrate2003wave}
S~Abrate.
\newblock Wave propagation in lightweight composite armor.
\newblock In {\em Journal de Physique IV (Proceedings)}, volume 110, pages
  657--662. EDP sciences, 2003.

\bibitem{abril2023pymc}
Oriol Abril-Pla, Virgile Andreani, Colin Carroll, Larry Dong, Christopher~J
  Fonnesbeck, Maxim Kochurov, Ravin Kumar, Junpeng Lao, Christian~C Luhmann,
  Osvaldo~A Martin, et~al.
\newblock Pymc: a modern, and comprehensive probabilistic programming framework
  in python.
\newblock {\em PeerJ Computer Science}, 9:e1516, 2023.

\bibitem{ter2008differential}
Cajo~JF Ter~Braak and Jasper~A Vrugt.
\newblock Differential evolution markov chain with snooker updater and fewer
  chains.
\newblock {\em Statistics and Computing}, 18:435--446, 2008.

\bibitem{vehtari2021rank}
Aki Vehtari, Andrew Gelman, Daniel Simpson, Bob Carpenter, and Paul-Christian
  B{\"u}rkner.
\newblock Rank-normalization, folding, and localization: An improved $\hat{R}$
  ̂for assessing convergence of mcmc (with discussion).
\newblock {\em Bayesian analysis}, 16(2):667--718, 2021.

\bibitem{vats2019multivariate}
Dootika Vats, James~M Flegal, and Galin~L Jones.
\newblock Multivariate output analysis for markov chain monte carlo.
\newblock {\em Biometrika}, 106(2):321--337, 2019.

\bibitem{Gelman:1995}
A.~Gelman, J.B. Carlin, H.S. Stern, and D.B. Rubin.
\newblock {\em {Bayesian Data Analysis}}.
\newblock Chapman and Hall/CRC, London, 1995.

\bibitem{Gelman:1996}
Andrew Gelman, Xiao-Li Meng, and Hal Stern.
\newblock Posterior predictive assessment of model fitness via realized
  discrepancies.
\newblock {\em Statistica Sinica}, 6(4):733--760, 1996.

\bibitem{zamo2018estimation}
Micha{\"e}l Zamo and Philippe Naveau.
\newblock Estimation of the continuous ranked probability score with limited
  information and applications to ensemble weather forecasts.
\newblock {\em Mathematical Geosciences}, 50(2):209--234, 2018.

\bibitem{you2022learning}
Huaiqian You, Quinn Zhang, Colton~J Ross, Chung-Hao Lee, and Yue Yu.
\newblock Learning deep implicit fourier neural operators (ifnos) with
  applications to heterogeneous material modeling.
\newblock {\em arXiv preprint arXiv:2203.08205}, 2022.

\end{thebibliography}
\end{document}